\documentclass{article}

\usepackage{microtype}
\usepackage{graphicx}
\usepackage{subcaption}
\usepackage{booktabs} % for professional tables
\usepackage{enumitem}
\newcommand{\bitem}[1]{\par\noindent\hangindent=1em\hangafter=1\makebox[1em][l]{$\bullet$}\textbf{#1}~}
\usepackage{hyperref}

\usepackage[preprint]{icml2027}
\usepackage{amsmath}
\usepackage{amssymb}
\usepackage{mathtools}
\usepackage{amsthm}

\usepackage[capitalize,noabbrev]{cleveref}

\theoremstyle{plain}
\newtheorem{theorem}{Theorem}[section]
\newtheorem{proposition}[theorem]{Proposition}
\newtheorem{lemma}[theorem]{Lemma}
\newtheorem{corollary}[theorem]{Corollary}
\theoremstyle{definition}

\newtheorem{assumption}[theorem]{Assumption}
\theoremstyle{remark}
\newtheorem{remark}[theorem]{Remark}

\newcommand{\argmin}{\mathop{\rm arg~min}\limits}

\usepackage{algorithm}
\usepackage{algorithmic}

\providecommand{\R}{\mathbb{R}}
\providecommand{\E}{\mathbb{E}}
\providecommand{\cP}{\mathcal{P}}

\providecommand{\Ltwo}{\mathcal{L}^2(\nu)}
\providecommand{\ip}[2]{\langle #1,#2\rangle_\nu}

\providecommand{\tF}{\Tilde{F}}
\providecommand{\tH}{\Tilde{H}}
\providecommand{\tG}{\Tilde{G}}
\providecommand{\kap}{\varkappa}
\providecommand{\dom}{\mathrm{dom}\,}
\providecommand{\prox}{\mathrm{prox}}
\providecommand{\Id}{\mathrm{Id}}
\providecommand{\Delt}{\Delta_K}
\providecommand{\push}{_{\#}}

\usepackage[textsize=tiny]{todonotes}

\icmltitlerunning{Accelerated Multiobjective Proximal Gradient over Probability Measures}

\begin{document}

\twocolumn[
  \icmltitle{Accelerated Proximal Gradient Methods for Multiobjective Optimization over Probability Measures via Hilbert-Space Lifting}

  % It is OKAY to include author information, even for blind submissions: the
  % style file will automatically remove it for you unless you've provided
  % the [accepted] option to the icml2027 package.

  % List of affiliations: The first argument should be a (short) identifier you
  % will use later to specify author affiliations Academic affiliations
  % should list Department, University, City, Region, Country Industry
  % affiliations should list Company, City, Region, Country

  % You can specify symbols, otherwise they are numbered in order. Ideally, you
  % should not use this facility. Affiliations will be numbered in order of
  % appearance and this is the preferred way.
  \icmlsetsymbol{equal}{*}

  \begin{icmlauthorlist}
    \icmlauthor{Dai Hai Nguyen}{xxx}
    %\icmlauthor{Firstname2 Lastname2}{equal,yyy,comp}
    %\icmlauthor{Firstname3 Lastname3}{comp}
    %\icmlauthor{Firstname4 Lastname4}{sch}
    %\icmlauthor{Firstname5 Lastname5}{yyy}
   
  \end{icmlauthorlist}

  \icmlaffiliation{xxx}{Graduate School of Information Science and Technology, Hokkaido University, Japan}
  %\icmlaffiliation{comp}{Company Name, Location, Country}
  %\icmlaffiliation{sch}{School of ZZZ, Institute of WWW, Location, Country}

  \icmlcorrespondingauthor{Dai Hai Nguyen}{hai@ist.hokudai.ac.jp or haidnguyen0909@gmail.com}
  %\icmlcorrespondingauthor{Firstname2 Lastname2}{first2.last2@www.uk}

  % You may provide any keywords that you find helpful for describing your
  % paper; these are used to populate the "keywords" metadata in the PDF but
  % will not be shown in the document
  \icmlkeywords{Multiobjective optimization, optimal transport map, proximal gradient methods, accelerated methods}

  \vskip 0.3in
]

% this must go after the closing bracket ] following \twocolumn[ ...

% This command actually creates the footnote in the first column listing the
% affiliations and the copyright notice. The command takes one argument, which
% is text to display at the start of the footnote. The \icmlEqualContribution
% command is standard text for equal contribution. Remove it (just {}) if you
% do not need this facility.

% Use ONE of the following lines. DO NOT remove the command.
% If you have no special notice, KEEP empty braces:
\printAffiliationsAndNotice{}  % no special notice (required even if empty)
% Or, if applicable, use the standard equal contribution text:
% \printAffiliationsAndNotice{\icmlEqualContribution}
\begin{abstract}
Many sampling and inference tasks require a single distribution that balances several targets.
We formulate this task as multiobjective optimization of Kullback--Leibler (KL) divergences over probability measures, in the Pareto sense.
The particle method MWGraD~\cite{nguyen2025multiple} uses explicit Wasserstein gradient steps,
and its accelerated version A-MWGraD~\cite{nguyen2026accelerated} lacks discrete-time guarantees.
We propose the multiobjective Wasserstein proximal gradient (MWPG), which uses
implicit Wasserstein gradient steps, and its accelerated version, multiobjective Wasserstein accelerated proximal gradient (MWAPG), with rigorous discrete-time guarantees.
Our key idea is a Hilbert-space lifting: each measure is represented as the pushforward of a
fixed base measure by a map in a Hilbert space.
The problem then becomes a constrained multiobjective composite problem over transport maps, to which accelerated
proximal gradient methods apply.
For convex and smooth potentials, we show that MWAPG decreases a Pareto merit function at the rate
$\mathcal{O}(1/k^2)$; for $m$-strongly convex and $L$-smooth potentials, the rate is
$\mathcal{O}\big((1-\sqrt{m/L})^k\big)$.
We also give closed-form Gaussian updates and particle implementations based on input-convex
neural networks, which require no score estimation.
Experiments on synthetic and real-world dataset show that MWAPG needs far fewer iterations than MWPG
and matches or outperforms A-MWGraD in Bayesian multi-task learning.
\end{abstract}

\section{Introduction}
Many problems in machine learning and statistics can be cast as minimizing a functional over
probability measures. For example, sampling from a posterior $\pi\propto e^{-V}$
amounts to minimizing $\mathrm{KL}(\cdot\|\pi)$, and variational inference solves the same problem
over a family of distributions \cite{jordan1999introduction,fox2012tutorial,pmlr-v258-nguyen25d,nguyen2023mirror}.
Multi-task Bayesian learning, federated learning and sampling under competing models involve
several targets $\pi_1,\dots,\pi_K$ \cite{phan2022stochastic,nguyen2025multiple}.
Since a single distribution rarely minimizes all divergences at once, we consider the
multiobjective problem.
Given functionals $G_{1},\dots,G_{K}:\mathcal{P}_{2}(\mathbb{R}^d)\to\mathbb{R}$, $K\ge2$,
on the space $\mathcal{P}_{2}(\mathbb{R}^d)$ of probability measures on $\mathbb{R}^{d}$ with finite
second moment, we seek a distribution that solves
\begin{equation}
\label{definition:MODO}
\min_{\mu\in\mathcal{P}_{2}(\mathbb{R}^d)}\big(G_{1}(\mu),\dots,G_{K}(\mu)\big),  G_i(\mu)=\mathrm{KL}(\mu\|\pi_i),
\end{equation}
in the Pareto sense: $\mu^*$ is Pareto optimal if no $\mu$ improves one objective without
worsening another.
For KL objectives, the Pareto set is explicit: it consists of the normalized geometric means
$\pi_{\boldsymbol{\lambda}}\propto\prod_i\pi_i^{\lambda_i}$, $\boldsymbol{\lambda}\in\Delta_K$
(Proposition~\ref{proposition:meritforklobjective}).
A method for \eqref{definition:MODO} must therefore find the weights $\boldsymbol{\lambda}$ and sample
$\pi_{\boldsymbol{\lambda}}$ simultaneously.

Accelerating \eqref{definition:MODO} in discrete time faces two obstacles absent from the
Euclidean setting of \cite{tanabe2023accelerated}.
First, the Wasserstein space is curved, so momentum, a linear extrapolation, has no intrinsic meaning.
Second, the entropy part of $\mathrm{KL}$ is nonsmooth and convex only along generalized geodesics
\cite{ambrosio2005gradient}.
We resolve both with one idea, \emph{Hilbert-space lifting}: we work in the Hilbert space
$\mathcal{L}^2(\nu)$ of transport maps from a base measure $\nu$.
In this flat space, the potential energies are convex and smooth, momentum is a linear combination
of maps, and the entropy is convex on the closed convex cone $\mathcal{K}_{\nu}$ of optimal
transport (Brenier) maps.
The curved, nonsmooth problem thus becomes a constrained composite problem in a Hilbert space
(Table~\ref{tab:hilbert-lifting}).
The proximal step is restricted to the cone; extrapolated points may leave it, but the entropy is
never evaluated there, and every convexity inequality is applied at points in the cone.
\textbf{Our contributions in this work are as follows.}

\noindent \textbf{Hilbert-space lifting.} We show that lifting to $\mathcal{L}^2(\nu)$ allows the
use of accelerated multiobjective proximal gradient methods and gives momentum a precise meaning.
For convex potentials $V_i$ ($\pi_i\propto e^{-V_i}$), all objectives are convex on the closed
convex cone $\mathcal{K}_\nu$ (Lemma~\ref{lemma:liftedproblem}). The subproblem is well posed in
infinite dimensions. The Pareto set and the merit function have closed forms
(Proposition~\ref{proposition:meritforklobjective}).

\noindent \textbf{Algorithms.} We propose MWPG, a \textbf{m}ultiobjective \textbf{W}asserstein
\textbf{p}roximal \textbf{g}radient method with implicit steps, and two accelerated versions:
MWAPG-C for convex potentials and MWAPG-SC for strongly convex potentials
(Algorithm~\ref{alg:mwapg}). Each step solves the anchored multiobjective subproblem
of \cite{tanabe2023accelerated}, lifted to $\mathcal{L}^2(\nu)$ and restricted to $\mathcal{K}_\nu$.
We prove that this subproblem has a unique solution. The solution is a single entropy proximal step,
with weights from a $K$-dimensional concave dual problem. The gradient of the dual is given
explicitly by Danskin's theorem \cite{danskin2012theory} (Lemma~\ref{lem:sub}).

 \noindent\textbf{Discrete-time guarantees.} For convex, $L$-smooth potentials, MWAPG-C satisfies $u_0(\mu_k)\leq 2LR/(k+1)^2$ for the merit function $u_0$ (Theorem~\ref{thm:convex}). For $m$-strongly convex potentials, MWAPG-SC satisfies $u_0(\mu_k)\leq 2(1-\sqrt{m/L})^k \max_i \mathrm{KL}(\mu_0\|\pi_i)$ (Theorem~\ref{thm:sc}, Corollary~\ref{cor:sc-explicit}). The rates still hold when the errors in the proximal steps and weights decay fast enough (Theorem~\ref{thm:inexact}). To our knowledge, these are the first discrete-time accelerated rates for multiobjective optimization over probability measures.

 \noindent\textbf{Exact treatment of the entropy.} We handle the entropy by its proximal map on $\mathcal{K}_\nu$, so we need no score estimation and no kernel bandwidth. For Gaussian families, the proximal map has a closed form. We prove that this closed form is the exact solution over all of $\mathcal{K}_\nu$ (Proposition~\ref{prop:gauss}). For general targets, we compute the proximal map with an input-convex neural network (ICNN) \cite{amos2017input}. The gradient of an ICNN is a Brenier map by construction.

 \noindent\textbf{Experiments.} On ill-conditioned Gaussian targets, our accelerated methods need up to $58\times$ fewer iterations than MWPG and MWGraD, and their iteration counts grow like $\sqrt{\varkappa}$ or slower. On a problem with a non-Gaussian Pareto set, the particle versions reach the Pareto set, while every Gaussian approximation fails, and acceleration helps in every method family. On multi-client Bayesian logistic regression with two real datasets, our accelerated methods reach Pareto stationarity $10^{-8}$ in a few hundred iterations, while the Gaussian MWGraD and A-MWGraD stall at $5\cdot10^{-6}$--$4\cdot10^{-5}$. Our solutions match the exact Pareto posterior. In Bayesian multi-task learning with neural networks, which is beyond our theory, MWAPG matches or exceeds A-MWGraD on three benchmarks and has the lowest test error in the early iterations.
%\end{itemize}

\section{Related work}

\paragraph{Multiobjective first-order methods.}
Steepest descent methods and MGDA~\cite{desideri2012multiple} combine gradients through a
min-norm problem, and they are widely used in multi-task learning~\cite{sener2018multi}.
\citet{tanabe2019proximal,tanabe2023accelerated} developed proximal and accelerated proximal
methods with rates for the merit function $u_0$.
Their accelerated method uses a subproblem anchored at the current iterate, which we lift to the
space of maps.
Accelerated and inexact multiobjective methods in Euclidean space are studied
in~\cite{sonntag2024fast,luo2025accelerated}.
For strongly convex problems, \citet{chen2024scaled} obtained the linear rate
$1-\sqrt{\min_i m_i/L_i}$, and \citet{sampaio2025complexity} gave lower complexity bounds.

\paragraph{Optimization over measures.}
The JKO scheme~\cite{jordan1998variational} and Wasserstein proximal gradient
methods~\cite{salim2020wasserstein,di2025inexact} treat the entropy implicitly.
\citet{ambrosio2005gradient} study generalized geodesics and the convexity of the entropy along them.
Acceleration over measures via transport maps was studied for a single
objective~\cite{tanaka2023accelerated,tang2026nesterov}, and for Gaussian families in the
Bures--Wasserstein space~\cite{nylof2026accelerated}.
Inexact accelerated proximal methods in Hilbert spaces are analysed
in~\cite{schmidt2011convergence,salzo2012inexact}; our inexact result follows their error model.
ICNNs have been used to compute JKO steps~\cite{alvarez2021optimizing} and transport
maps~\cite{makkuva2020optimal}.

\paragraph{Multi-target sampling.}
MOO-SVGD~\cite{liu2021profiling} applies MGDA to each particle within SVGD, spreading the
particles over the Pareto front.
MWGraD~\cite{nguyen2025multiple} applies the min-norm combination to the Wasserstein gradients
of the objectives, moving all particles to a single Pareto optimal distribution.
A-MWGraD~\cite{nguyen2026accelerated} accelerates this flow, with continuous-time guarantees.
Our methods are discrete-time, treat the entropy by a proximal step, and come with
non-asymptotic rates.

\begin{table}[t]
\centering
\caption{Hilbert-space lifting: objects over measures and their lifted counterparts in
$\mathcal{L}^2(\nu)$.}
\label{tab:hilbert-lifting}
\scriptsize
\setlength{\tabcolsep}{3pt}
\renewcommand{\arraystretch}{1.1}
\begin{tabular}{p{0.45\columnwidth} p{0.48\columnwidth}}
\toprule
\textbf{Over measures $\mathcal{P}_2(\mathbb{R}^d)$} & \textbf{Lifted to $\mathcal{L}^2(\nu)$} \\
\midrule
measure $\mu$ & map $X$ with $X_{\#}\nu=\mu$ \\
optimal transport map from $\nu$ & element of the cone $\mathcal{K}_\nu$ \\
$\mathcal{W}_2(\nu,\rho)$ & $\|\mathrm{Id}-Z_\rho\|_\nu$ \\
generalized geodesic (base $\nu$) & segment $(1-s)Z_0+sZ_1\subset\mathcal{K}_\nu$ \\
$F_i(\mu)=\int V_i\,d\mu$ & $\tilde F_i$: convex, $L$-smooth on all of $\mathcal{L}^2(\nu)$ \\
$H(\mu)=\int\mu\log\mu$ & $\tilde H$: convex on $\mathcal{K}_\nu$ (not on $\mathcal{L}^2(\nu)$) \\
momentum: no intrinsic meaning & $Y=X^k+\gamma_k(X^k-X^{k-1})$ \\
JKO / entropic proximal step & $\operatorname{prox}^{\mathcal{K}_\nu}_{\tilde H/L}$ (Lemma~\ref{lem:sub}) \\
Pareto set $\{\pi_{\boldsymbol{\lambda}}\}$ & $\{Z_{\pi_{\boldsymbol{\lambda}}}\}\subset\mathcal{K}_\nu$ \\
\bottomrule
\end{tabular}
\end{table}

\section{Setting}
\label{sec:setting}

\paragraph{Notation.}
$\mathcal{P}_2(\mathbb{R}^d)$ is the set of Borel probability measures with finite second moment,
$\mathcal{P}^\mathrm{ac}_2$ its subset of absolutely continuous measures, and $\mathcal{W}_2$ the
2-Wasserstein distance. For $\nu\in\mathcal{P}^\mathrm{ac}_2$, we write
$\mathcal{L}^2(\nu):=\mathcal{L}^2(\nu;\mathbb{R}^d)$, with inner product
$\langle X,Y\rangle_\nu=\int\langle X(x),Y(x)\rangle\,d\nu(x)$ and norm $\lVert\cdot\rVert_\nu$.
$\Delta_K$ is the unit simplex in $\mathbb{R}^K$. Appendix~\ref{app:notation} lists the notation.

\paragraph{Objectives.}
Let $\pi_i=e^{-V_i}/Z_i$. Then $G_i=F_i+H+\log Z_i$, where $F_i(\mu)=\int V_i\,d\mu$ is the
potential energy and $H(\mu)=\int\mu\log\mu$ is the negative entropy.
Additive constants play no role, so we drop $\log Z_i$ whenever only differences of $G_i$ appear.

\begin{assumption}\label{ass:1}
\leavevmode
\begin{enumerate}[label=\textbf{(A\arabic*)}, ref=(A\arabic*), leftmargin=*]
\item\label{ass:A1} Each $V_i:\mathbb{R}^d\to\mathbb{R}$ is convex and $L$-smooth, i.e.,
$\nabla V_i$ is $L$-Lipschitz, and $e^{-V_i}$ is integrable. In the strongly convex case, each
$V_i$ is $m$-strongly convex, and $\varkappa=L/m$ is the condition number.
\item\label{ass:A2} The base measure is the initial distribution, i.e.,
$\nu=\mu_0\in\mathcal{P}_2^{\mathrm{ac}}(\mathbb{R}^d)$, and $H(\mu_0)<\infty$.
\item\label{ass:A3} Every $\rho\in\mathcal{P}_2$ with $G_i(\rho)\le G_i(\mu_0)$ for all
$i=1,\dots,K$ is weakly dominated by a weakly Pareto optimal $\rho^*\in\mathcal{P}_2$, i.e.,
$G_i(\rho^*)\le G_i(\rho)$ for all $i$; and the Pareto radius $R$ in~\eqref{eq:R} is finite.
\end{enumerate}
\end{assumption}

Assumption~\ref{ass:A3} is standard in the convex case~\cite{tanabe2019proximal,nguyen2025multiple}.
In the strongly convex case it holds automatically, with an explicit bound on $R$ (Lemma~\ref{lem:A3sc}).
The choice $\nu=\mu_0$ in~\ref{ass:A2} is only for convenience: any $\nu\in\mathcal{P}_2^{\mathrm{ac}}$
works, with any $X^0\in\mathcal{K}_\nu$ such that $X^0_{\#}\nu=\mu_0$.

\paragraph{Hilbert-space lifting.}
For $X\in\mathcal{L}^2(\nu)$, let $\mu_X=X_{\#}\nu$, and define $\tilde F_i(X)=\int V_i(X)\,d\nu$,
$\tilde H(X)=H(\mu_X)$ and $\tilde G_i=\tilde F_i+\tilde H$.
The cone of optimal transport maps from $\nu$ is
\begin{align*}
  \mathcal{K}_\nu=\Big\{Z\in\mathcal{L}^2(\nu):\ & Z\in\partial\varphi\ \nu\text{-a.e.\ for a proper}\\
  & \text{lower semicontinuous (lsc) convex }\varphi\Big\}.
\end{align*}
The following facts are classical; Appendix~\ref{proof:liftedproblem} proves them or gives references.
\begin{lemma}[Structure of the lifted problem]
\label{lemma:liftedproblem}
Under~\ref{ass:A1}--\ref{ass:A2}:
\begin{enumerate}[label=(\alph*), leftmargin=*]
\item $\mathcal{K}_\nu$ is a closed convex cone in $\mathcal{L}^2(\nu)$ containing the identity
$\mathrm{Id}$. For every $\rho\in\mathcal{P}_2$, there is a unique $Z_\rho\in\mathcal{K}_\nu$ with
$(Z_\rho)_{\#}\nu=\rho$, and $\lVert\mathrm{Id}-Z_\rho\rVert_\nu=\mathcal{W}_2(\nu,\rho)$~\cite{mccann1995existence}.
\item $\tilde F_i$ is finite, convex and $L$-smooth on $\mathcal{L}^2(\nu)$, with
$\nabla\tilde F_i(X)=\nabla V_i\circ X$. If $V_i$ is $m$-strongly convex, so is $\tilde F_i$.
\item $\tilde H$ is convex and lsc (in the norm topology) on $\mathcal{K}_\nu$, but not convex on
$\mathcal{L}^2(\nu)$.
\item $\tilde H(X)\ge-\frac d2\log\big(\frac{2\pi e}{d}\lVert X\rVert^2_\nu\big)$ for every
$X\in\mathcal{L}^2(\nu)$.
\end{enumerate}
\end{lemma}
The convexity of $\tilde H$ on $\mathcal{K}_\nu$ is the convexity of the entropy along generalized
geodesics with base $\nu$ \citep[Sec.~9.2]{ambrosio2005gradient}: for $Z_0,Z_1\in\mathcal{K}_\nu$
and $s\in[0,1]$, the curve $\big((1-s)Z_0+sZ_1\big)_{\#}\nu$ is such a geodesic.
\paragraph{Pareto optimality and merit function.}
A measure $\mu$ is weakly Pareto optimal if no $\rho$ satisfies $G_i(\rho)<G_i(\mu)$ for all
$i=1,\dots,K$. Following~\cite{nguyen2025multiple,nguyen2026accelerated}, we measure the distance
of $\mu$ to the weakly Pareto set $\mathcal{P}^*$ by the merit function
\begin{equation}
\label{eqn:defofmeritfunction}
  u_0(\mu)=\sup_{\rho\in\mathcal{P}_2}\min_i\big[G_i(\mu)-G_i(\rho)\big].
\end{equation}
Taking $\rho=\mu$ gives $u_0(\mu)\ge0$, with equality if and only if $\mu$ is weakly Pareto optimal.
\begin{proposition}[Merit function for KL objectives]
\label{proposition:meritforklobjective}
Under~\ref{ass:A1}, for every $\mu\in\mathcal{P}_2$ with $G_i(\mu)<\infty$ for all $i=1,\dots,K$,
\begin{equation}
\label{eqn:meritfunction}
  u_0(\mu)=\min_{\boldsymbol{\lambda}\in\Delta_K}\mathrm{KL}(\mu\|\pi_{\boldsymbol{\lambda}}),
  \qquad\pi_{\boldsymbol{\lambda}}=\frac{e^{-\sum_i\lambda_iV_i}}{Z_{\boldsymbol{\lambda}}}.
\end{equation}
Consequently, $\mathcal{P}^*=\{\pi_{\boldsymbol{\lambda}}:\boldsymbol{\lambda}\in\Delta_K\}$, and every
weakly Pareto point is log-concave. For Gaussian targets $\pi_i=\mathcal{N}(m_i,\Sigma_i)$,
\begin{align*}
  \pi_{\boldsymbol{\lambda}}&=\mathcal{N}\big(m_{\boldsymbol{\lambda}},P^{-1}_{\boldsymbol{\lambda}}\big),\\
  P_{\boldsymbol{\lambda}}&=\textstyle\sum_i\lambda_i\Sigma_i^{-1},\quad
  m_{\boldsymbol{\lambda}}=P_{\boldsymbol{\lambda}}^{-1}\sum_i\lambda_i\Sigma^{-1}_im_i.
\end{align*}
\end{proposition}
The proof is in Appendix~\ref{appendix:proofmeritfunction}. By~\eqref{eqn:meritfunction}, $u_0$ is
the KL divergence to the nearest Pareto point; for Gaussian iterates, we evaluate it exactly.

\paragraph{Pareto radius.}
Let $\mathcal{S}_0=\{\rho\in\mathcal{P}_2:G_i(\rho)\le G_i(\mu_0)\ \forall i\}$ be the initial level
set. Following~\cite{nguyen2025multiple}, we define
\begin{equation}
\label{eq:R}
  R=\sup_{\rho\in\mathcal{P}^*\cap\mathcal{S}_0}\mathcal{W}^2_2(\mu_0,\rho).
\end{equation}
For KL objectives, the supremum runs over the compromises $\pi_{\boldsymbol{\lambda}}$ that are at
least as good as $\mu_0$ for every target, so $R\le\max_{\boldsymbol{\lambda}\in\Delta_K}\mathcal{W}^2_2(\mu_0,\pi_{\boldsymbol{\lambda}})$.

\paragraph{Enlarged level set.}
For the inexact methods (Theorem~\ref{thm:inexact}), the iterates may leave $\mathcal{S}_0$ slightly.
For $S\ge0$, let $\mathcal{S}_S=\{\rho\in\mathcal{P}_2:G_i(\rho)\le G_i(\mu_0)+S\ \forall i\}$ and
$R_S=\sup_{\rho\in\mathcal{P}^*\cap\mathcal{S}_S}\mathcal{W}^2_2(\mu_0,\rho)$.
We write \textbf{(A3$_S$)} for Assumption~\ref{ass:A3} with $\mathcal{S}_0$ and $R$ replaced by
$\mathcal{S}_S$ and $R_S$; thus~\ref{ass:A3} is \textbf{(A3$_0$)}, and $R_0=R$.

\renewcommand{\algorithmiccomment}[1]{\hfill$\triangleright$~#1}
\begin{algorithm}[t]
\caption{MWPG / MWAPG-C / MWAPG-SC (plain or monotone)}\label{alg:mwapg}
\begin{algorithmic}[1]
\REQUIRE $X^0=X^{-1}=Z^0\in\mathcal{K}_\nu$ (e.g.\ $\mathrm{Id}$ for $\nu=\mu_0$); $t_0=0$
\FOR{$k=0,1,2,\dots$}
  \STATE $t_{k+1}=\frac12\big(1+\sqrt{1+4t_k^2}\big)$ \COMMENT{$t_1=1$}
  \STATE \textbf{Plain:} $Y^k\gets X^k+\gamma_k(X^k-X^{k-1})$, with
  \item[] \hspace{1.5em}$\gamma_k=0$ (MWPG),
  \item[] \hspace{1.5em}$\gamma_k=(t_k-1)_+/t_{k+1}$ (MWAPG-C),
  \item[] \hspace{1.5em}$\gamma_k=\frac{1-\theta}{1+\theta}$, $\theta=\sqrt{m/L}$ (MWAPG-SC)
  \STATE \textbf{Monotone C:} $Y^k\gets X^k+\frac{t_k}{t_{k+1}}(Z^k-X^k)$
  \item[] \hspace{6.2em}$+\frac{(t_k-1)_+}{t_{k+1}}(X^k-X^{k-1})$
  \STATE \textbf{Monotone SC:} $Y^k\gets X^k+\frac{\theta}{1+\theta}(U^k-X^k)$,
  \item[] \hspace{6.2em}$U^k=X^{k-1}+\theta^{-1}(Z^k-X^{k-1})$
  \STATE $g_i\gets\nabla V_i\circ Y^k$, $i=1,\dots,K$
  \STATE $\boldsymbol{\lambda}^k\gets\arg\max_{\boldsymbol{\lambda}\in\Delta_K}q(\boldsymbol{\lambda})$ \COMMENT{$K$-dim.\ dual}
  \STATE $Z^{k+1}\gets Z(\boldsymbol{\lambda}^k)$ from~\eqref{eqn:dualstep} \COMMENT{prox on $\mathcal{K}_\nu$}
  \STATE \textbf{Plain:} $X^{k+1}\gets Z^{k+1}$
  \STATE \textbf{Monotone:} $X^{k+1}\gets Z^{k+1}$ if no $G_i$ increases,
  \item[] \hspace{5.2em}else $X^{k+1}\gets X^k$ (optional restart)
\ENDFOR
\STATE \textbf{return} $\mu_k=X^k_{\#}\nu$
\end{algorithmic}
\end{algorithm}

\section{Multiobjective Wasserstein gradient methods through Hilbert-space lifting}
\label{sec:methods}
We now describe our algorithms for~\eqref{definition:MODO} as updates of transport maps.
We extend the multiobjective proximal methods of~\cite{tanabe2023accelerated} from Euclidean
space to the convex cone $\mathcal{K}_\nu$ of transport maps.

\paragraph{Proximal step for multiple objectives.}
Let $X^k\in\mathcal{K}_\nu$ be the current iterate and $Y^k\in\mathcal{L}^2(\nu)$ an extrapolated
point, which need not lie in $\mathcal{K}_\nu$. Let $g_i=\nabla\tilde F_i(Y^k)=\nabla V_i\circ Y^k$.
Following~\cite{tanabe2023accelerated}, we define
\begin{equation}
\label{eq:psi}
  \psi_i(Z)=\langle g_i,Z-Y^k\rangle_\nu+\tilde F_i(Y^k)+\tilde H(Z)-\tilde G_i(X^k),
\end{equation}
and the next iterate
\begin{equation}
\label{eqn:proximalstep}
\begin{split}
  X^{k+1}&=\argmin_{Z\in\mathcal{K}_\nu}\Phi(Z),\\
  \Phi(Z)&=\max_i\psi_i(Z)+\frac{L}{2}\lVert Z-Y^k\rVert^2_\nu.
\end{split}
\end{equation}
Here $\psi_i$ linearizes $\tilde F_i$ at $Y$ but keeps the exact entropy. The anchor
$\tilde F_i(Y)-\tilde G_i(X^k)$ makes $\psi_i$ a model of the change $\tilde G_i(Z)-\tilde G_i(X^k)$
relative to the current iterate. The entropy is evaluated only on $\mathcal{K}_\nu$.

\paragraph{Dual update for multiple objectives.}
We write $\max_i\psi_i=\max_{\boldsymbol{\lambda}\in\Delta_K}\sum_i\lambda_i\psi_i$ and exchange
$\min$ and $\max$ (justified by Lemma~\ref{lem:sub}(b)). Then~\eqref{eqn:proximalstep} becomes
$\max_{\boldsymbol{\lambda}\in\Delta_K}q(\boldsymbol{\lambda})$ with
$q(\boldsymbol{\lambda})=\min_{Z\in\mathcal{K}_\nu}\sum_i\lambda_i\psi_i(Z)+\frac L2\lVert Z-Y^k\rVert^2_\nu$.
Since all $\psi_i$ share the same entropy term, the inner minimizer is a single entropy proximal step,
\begin{equation}
\label{eqn:dualstep}
\begin{split}
  Z(\boldsymbol{\lambda})&=\mathrm{prox}_{\tilde H/L}^{\mathcal{K}_\nu}\Big(Y^k-\frac{1}{L}\sum_i\lambda_ig_i\Big),\\
  \mathrm{prox}_{\tilde H/L}^{\mathcal{K}_\nu}(T)&=\argmin_{Z\in\mathcal{K}_\nu}\tilde H(Z)+\frac{L}{2}\lVert Z-T\rVert^2_\nu.
\end{split}
\end{equation}
The next lemma collects the main properties of~\eqref{eqn:proximalstep} and~\eqref{eqn:dualstep};
the proof is in Appendix~\ref{app:sub}.
\begin{lemma}
\label{lem:sub}
Under~\ref{ass:A1}--\ref{ass:A2}, for every $X^k\in\mathcal{K}_\nu$ with $\tilde H(X^k)<\infty$
and every $Y\in\mathcal{L}^2(\nu)$:
\begin{enumerate}[label=(\alph*), leftmargin=*]
\item \eqref{eqn:proximalstep} has a unique solution $X^{k+1}$, and $Z(\boldsymbol{\lambda})$
in~\eqref{eqn:dualstep} is well defined for every $\boldsymbol{\lambda}\in\Delta_K$;
\item strong duality holds, $\min_{Z\in\mathcal{K}_\nu}\Phi(Z)=\max_{\boldsymbol{\lambda}\in\Delta_K}q(\boldsymbol{\lambda})$,
the dual has a maximizer $\boldsymbol{\lambda}^k$, and $X^{k+1}=Z(\boldsymbol{\lambda}^k)$;
\item complementary slackness holds: $\sum_i\lambda^k_i\psi_i(X^{k+1})=\max_i\psi_i(X^{k+1})$;
\item $q$ is concave and differentiable on $\Delta_K$ with
$\nabla q(\boldsymbol{\lambda})=\big(\psi_i(Z(\boldsymbol{\lambda}))\big)_{i=1}^K$, and $\nabla q$
is Lipschitz with constant $\lambda_{\max}(Q)/L$, where $Q_{ij}=\langle g_i,g_j\rangle_\nu$.
\end{enumerate}
\end{lemma}

\paragraph{Proposed algorithms.}
Algorithm~\ref{alg:mwapg} applies~\eqref{eqn:proximalstep} at $Y^k=X^k+\gamma_k(X^k-X^{k-1})$;
the three methods differ only in $\gamma_k$.
%\begin{itemize}
\bitem{MWPG} ($\gamma_k=0$, so $Y^k=X^k$) uses no momentum. It is the multiobjective
proximal gradient method of~\cite{tanabe2019proximal} lifted to $\mathcal{K}_\nu$; it decreases
every objective and needs only $L$.
For convex potentials, $u_0(\mu_k)\le LR/(2k)$, so $u_0(\mu_k)\le\epsilon$ after $O(LR/\epsilon)$ iterations.
For $m$-strongly convex potentials, $u_0(\mu_k)=O((1-1/\kap)^k)$, so $u_0(\mu_k)\le\epsilon$ after
$O(\kap\log\frac1\epsilon)$ iterations (Proposition~\ref{prop:mwpg}).
\bitem{MWAPG-C} ($\gamma_k=(t_k-1)_+/t_{k+1}$, with $t_k\approx k/2$) uses FISTA momentum
and needs only $L$. For convex potentials, $u_0(\mu_k)\le2LR/(k+1)^2$, so $u_0(\mu_k)\le\epsilon$
after $O(\sqrt{LR/\epsilon})$ iterations (Theorem~\ref{thm:convex}).
\bitem{MWAPG-SC} ($\gamma_k=\frac{1-\theta}{1+\theta}$, with $\theta=\sqrt{m/L}$) uses constant
momentum and needs both $L$ and $m$. For $m$-strongly convex potentials,
$u_0(\mu_k)\le(1-\theta)^k\big[u_0(\mu_0)+\frac m2R\big]$, so $u_0(\mu_k)\le\epsilon$ after
$O(\sqrt\kap\log\frac1\epsilon)$ iterations (Theorem~\ref{thm:sc}).
%\end{itemize}

\paragraph{Monotone versions.}
The plain momentum methods may increase an objective. The \emph{monotone} versions, multiobjective
analogues of MFISTA~\cite{beck2009fast}, accept $Z^{k+1}$ only if no objective increases.
Otherwise, they set $X^{k+1}=X^k$ and may also restart the momentum.
They extrapolate from the proximal points $Z^k$ (lines~4--5 of Algorithm~\ref{alg:mwapg}).
They keep the rates (Theorem~\ref{thm:mono}) at the cost of one evaluation of the $K$ objectives
per iteration. For $K=1$, they reduce to Wasserstein (M)FISTA with an exact entropy proximal step.
For Dirac initializations and no entropy, they formally reduce to the method of~\cite{tanabe2023accelerated}.

\begin{remark}
\textbf{Iterates stay well defined.} The extrapolated point $Y^k$ combines two Brenier maps with a
negative coefficient. In general it is not a Brenier map, and $\tilde H(Y^k)$ has no useful meaning.
However, the method never evaluates $\tilde H$ at $Y^k$; it uses only $\tilde F_i(Y^k)$ and
$\nabla\tilde F_i(Y^k)$, which are well defined on $\mathcal{L}^2(\nu)$.
The output of the proximal step lies in $\mathcal{K}_\nu$ by construction.
\textbf{Cost per iteration.} One iteration needs the $K$ gradient fields $\nabla V_i\circ Y^k$, the
Gram matrix $Q$ and a few proximal maps for the dual solver. It needs no kernel and no pairwise
particle interactions.
\end{remark}

%=====================================================================================================================
\section{Convergence analysis}
\label{sec:theory}

Our main tool is a one-step inequality that holds for every comparison map $W$ in the cone.
It is the analogue of the basic inequality of proximal gradient methods, with the
multiobjective weights $\boldsymbol{\lambda}^k$ and the restriction $W\in\mathcal{K}_\nu$.

\begin{lemma}[One-step inequality]\label{lem:onestep}
Under~\ref{ass:A1}--\ref{ass:A2}, let $X^{k+1}$ solve~\eqref{eqn:proximalstep} with dual solution
$\boldsymbol{\lambda}^k$, and let $m\ge0$ be a strong convexity constant of all $V_i$ ($m=0$ in the
convex case). Then for every $W\in\mathcal{K}_\nu$,
\begin{equation}\label{eq:onestep}
\begin{aligned}
  &\max_i\big[\tG_i(X^{k+1})-\tG_i(X^k)\big]\le\textstyle\sum_i\lambda^k_i\big[\tG_i(W)-\tG_i(X^k)\big]\\
  &\qquad+\tfrac{L-m}2\|W-Y^k\|_\nu^2-\tfrac L2\|W-X^{k+1}\|_\nu^2.
\end{aligned}
\end{equation}
\end{lemma}

Accelerated iterates may increase an objective, so monotonicity is unavailable. The next lemma,
which follows from Lemma~\ref{lem:onestep} with $W=X^k$, replaces it: all iterates stay in the
initial level set $\mathcal{S}_0$.

\begin{lemma}[Level set]\label{lem:level}
Under~\ref{ass:A1}--\ref{ass:A2}, the iterates of all three methods satisfy $X^k\in\mathcal{K}_\nu$
and $G_i(\mu_k)\le G_i(\mu_0)-\frac L2\|X^k-X^{k-1}\|_\nu^2$ for all $i$ and $k\ge1$; in particular,
$\mu_k\in\mathcal{S}_0$.
\end{lemma}

We now state our main results. For convex potentials, MWAPG-C uses only the smoothness constant
$L$ and decreases the merit function at the accelerated rate $O(1/k^2)$.

\begin{theorem}[Convex case]\label{thm:convex}
Under Assumption~\ref{ass:1} (convex case), the iterates of MWAPG-C satisfy
\[
  u_0(\mu_k)\le\frac{L\,R}{2t_k^2}\le\frac{2LR}{(k+1)^2}\qquad(k\ge1).
\]
\end{theorem}

If the potentials are also strongly convex, MWAPG-SC converges linearly, at a rate that depends
on $\sqrt{\kap}$ rather than $\kap$.

\begin{theorem}[Strongly convex case]\label{thm:sc}
If, in addition, each $V_i$ is $m$-strongly convex, the iterates of MWAPG-SC satisfy, with
$\theta=\sqrt{m/L}=\kap^{-1/2}$,
\[
  u_0(\mu_k)\le(1-\theta)^k\Big[u_0(\mu_0)+\frac m2R\Big]\qquad(k\ge0).
\]
\end{theorem}

The bound in Theorem~\ref{thm:sc} contains the Pareto radius $R$. In the strongly convex case,
$R$ is bounded by the initial KL divergences (Lemma~\ref{lem:A3sc}), which makes the bound fully explicit.

\begin{corollary}\label{cor:sc-explicit}
In the strongly convex case, $R\le\frac2m\max_i\mathrm{KL}(\mu_0\|\pi_i)$; hence
\[u_0(\mu_k)\le2(1-\theta)^k\max_i\mathrm{KL}(\mu_0\|\pi_i),\]
and $u_0(\mu_k)\le\epsilon$ after
$k\ge\sqrt\kap\,\log\big(2\max_i\mathrm{KL}(\mu_0\|\pi_i)/\epsilon\big)$ iterations.
\end{corollary}

For comparison, we also give the rates of MWPG. MWPG decreases every objective but is slower,
needing $O(\kap)$ rather than $O(\sqrt{\kap})$ iterations per digit of accuracy.

\begin{proposition}[No momentum]\label{prop:mwpg}
Under Assumption~\ref{ass:1}, MWPG decreases every objective,
$G_i(\mu_{k+1})\le G_i(\mu_k)-\frac L2\|X^{k+1}-X^k\|_\nu^2$, and $u_0(\mu_k)\le LR/(2k)$.
In the $m$-strongly convex case, with $q=1-1/\kap$, $u_0(\mu_k)\le\frac m2R\,q^k/(1-q^k)$, which
requires $O(\kap\log(1/\epsilon))$ iterations.
\end{proposition}

The plain accelerated methods may increase an objective. The next theorem shows that the monotone
versions never do, while keeping the same rates.

\begin{theorem}[Monotone variants]\label{thm:mono}
Under the assumptions of Theorems~\ref{thm:convex} and~\ref{thm:sc}, the monotone versions of
MWAPG-C and MWAPG-SC (without restart) satisfy $G_i(\mu_{k+1})\le G_i(\mu_k)$ for all $i$ and $k$,
and the bounds of Theorems~\ref{thm:convex} and~\ref{thm:sc} hold verbatim.
\end{theorem}

\paragraph{Inexact proximal steps.}
In the particle implementation, the proximal map is computed approximately, and the dual is
solved to a tolerance. We call $(\hat{\boldsymbol{\lambda}}^k,X^{k+1})\in\Delta_K\times\mathcal{K}_\nu$
an $(\varepsilon_k,\delta_k)$-solution of~\eqref{eqn:proximalstep} if, with
$\Phi_k=\sum_i\hat\lambda^k_i\psi_i+\frac L2\|\cdot-Y^k\|_\nu^2$ and for all $W\in\mathcal{K}_\nu$,
\begin{equation}\label{eq:inexact}
\begin{aligned}
  &\Phi_k(X^{k+1})+\tfrac L2\|W-X^{k+1}\|_\nu^2\le\Phi_k(W)+\varepsilon_k,\\
  &\delta_k=\max_i\psi_i(X^{k+1})-\textstyle\sum_i\hat\lambda^k_i\psi_i(X^{k+1}).
\end{aligned}
\end{equation}
With the exact proximal map at an approximate weight vector, $\varepsilon_k=0$ and $\delta_k$ is the
duality gap at $\hat{\boldsymbol{\lambda}}^k$, which is computable and which we monitor.
The next theorem shows that the accelerated rates persist when the errors decay fast enough.

\begin{theorem}[Inexact version]\label{thm:inexact}
Under~\ref{ass:A1}--\ref{ass:A2}, let $\eta_k=\varepsilon_k+\delta_k$ and $S=\sum_k\eta_k<\infty$,
and assume~\textbf{(A3$_S$)}. Then MWAPG-C satisfies
\[
  u_0(\mu_k)\le\frac{2LR_S+4\sum_{j<k}t_{j+1}^2\eta_j}{(k+1)^2},
\]
which is $O(1/k^2)$ if $\sum_jj^2\eta_j<\infty$. MWAPG-SC satisfies
$u_0(\mu_k)\le(1-\theta)^k[u_0(\mu_0)+\frac m2R_S]+\sum_{j<k}(1-\theta)^{k-1-j}\eta_j$, which is
linear if $\eta_j$ decays geometrically.
\end{theorem}

\begin{remark}
MWAPG-SC reaches $\epsilon$-accuracy after $O(\sqrt{\kap}\log(1/\epsilon))$ iterations, and MWPG
after $O(\kap\log(1/\epsilon))$. The $\sqrt{\kap}$ dependence cannot be improved for first-order
multiobjective methods in Euclidean space~\cite{sampaio2025complexity}.
Our rates use the global constants $m=\min_im_i$ and $L=\max_iL_i$; scaling each objective as
in~\cite{chen2024scaled} could improve them.
\end{remark}

\section{Implementation}
\label{sec:impl}

\paragraph{Gaussian closed form.}
If $\mu_0$ and all $\pi_i$ are Gaussian, then all iterates are Gaussian. We take
$\nu=\mathcal N(0,I)$ (Section~\ref{sec:setting} allows any $\nu$) and represent the iterates by
affine maps $X(\xi)=A\xi+b$ with $A$ symmetric positive definite, which lie in $\mathcal{K}_\nu$.
Then $\nabla\tF_i(X)(\xi)=P_iA\xi+P_i(b-m_i)$ with $P_i=\Sigma_i^{-1}$,
$\ip XY=\mathrm{tr}(A_X^\top A_Y)+b_X^\top b_Y$, and $\tH(X)=H(\nu)-\log\det A$.

\begin{proposition}[Gaussian proximal map]\label{prop:gauss}
Let $T(\xi)=T_A\xi+T_b$ with an arbitrary $T_A\in\R^{d\times d}$, and let
$\frac12(T_A+T_A^\top)=U\mathrm{diag}(s)U^\top$.
The unique minimizer of $\tH(Z)+\frac L2\|Z-T\|_\nu^2$ over all of $\mathcal{K}_\nu$ is the affine map
\[
  Z^\star(\xi)=U\,\mathrm{diag}(a)\,U^\top\xi+T_b,
  a_j=\tfrac12\big(s_j+\sqrt{s_j^2+4/L}\big).
\]
\end{proposition}
Each $a_j$ solves $a^2-s_ja-1/L=0$, so it is positive even when $s_j<0$, and the iterate stays in
the cone. The key point is that the minimizer over the whole infinite-dimensional cone is affine.
The Gaussian algorithm therefore \emph{is} Algorithm~\ref{alg:mwapg}, and our theorems apply to it
without approximation (proof in Appendix~\ref{app:gauss}).
We solve the dual by accelerated projected ascent with step $L/\lambda_{\max}(Q)$ until
$\delta_k<10^{-11}$, and we evaluate $u_0$ exactly by~\eqref{eqn:meritfunction}.

\paragraph{Particles and ICNN proximal steps.}
In general, we store the iterate as particles $X^k(\xi_j)$ at fixed base samples $\xi_j\sim\nu$,
$j=1,\dots,n$. Momentum acts on the particle arrays, and $\nabla V_i\circ Y^k$ is evaluated at the
extrapolated particles. We compute the proximal map~\eqref{eqn:dualstep} as
$Z=\nabla\varphi_\vartheta$, where $\varphi_\vartheta$ is an ICNN with a quadratic skip connection,
\begin{align}
\begin{split}
\label{eq:icnn}
  \varphi_\vartheta(x)&=\tfrac12x^\top Ax + b^\top x\\
  &+\textstyle\sum_{h=1}^{n_h}a_h\,\mathrm{softplus}(v_h^\top x+c_h),
\end{split}
\end{align}
with $A=\Lambda\Lambda^\top+\epsilon_0I$ and $a_h\ge0$. Then $\varphi_\vartheta$ is strictly convex,
and $\nabla\varphi_\vartheta$ is a Brenier map.
By the change of variables formula, $\tH(\nabla\varphi_\vartheta)=H(\nu)-\E_\nu\log\det\nabla^2\varphi_\vartheta$.
The proximal step at $T=Y-\frac1L\sum_i\lambda_ig_i$ minimizes
\begin{align}\label{eq:icnnloss}
\begin{split}
  -\tfrac1{n}\textstyle\sum_{j}&\log\det\nabla^2\varphi_\vartheta(\xi'_j)\\
  &+\tfrac L{2n}\sum_j\big\|\nabla\varphi_\vartheta(\xi_j)-T(\xi_j)\big\|_2^2,
\end{split}
\end{align}
where the entropy term uses fresh base samples $\xi'_j$ at every iteration.
We minimize~\eqref{eq:icnnloss} by L-BFGS, warm-started from the previous network.
We solve the $K$-dimensional dual by a few Newton steps, with the Hessian estimated by finite
differences of $\nabla q=\psi(Z(\cdot))$; each evaluation is a network fit.
On the simplex only the differences $\psi_i-\psi_j$ matter, and the shared term $\tH(Z)$ cancels in
them, so the dual iterations never need the entropy value.
Algorithm~\ref{alg:icnn} (Appendix~\ref{app:icnn}) gives the procedure and all hyperparameters.
It produces $(\varepsilon_k,\delta_k)$-solutions in the sense of~\eqref{eq:inexact}.
We monitor $\delta_k$; the error $\varepsilon_k$ contains the optimization and approximation errors
of the network, and we do not certify it.

%=====================================================================================================================

\section{Experiments}
\label{sec:exp}

\paragraph{Baselines.}
We compare MWPG, MWAPG-C and MWAPG-SC with MWGraD~\cite{nguyen2025multiple} and
A-MWGraD~\cite{nguyen2026accelerated}, including its strongly convex variant A-MWGraD-SC. All methods start from the same
distribution, and the particle methods use the same particles. Both baselines move particles along the combined field
$\bar\varphi=\sum_i\lambda_i\varphi_i$, $\varphi_i=-\nabla V_i-\nabla\log\mu_k$, where $\boldsymbol\lambda$ are the min-norm weights in
$\mathcal{L}^2(\mu_k)$. We solve this weight problem exactly at every iteration, as in the theory of~\cite{nguyen2025multiple}; their
practical algorithm takes one projected gradient step on the weights instead. The score $\nabla\log\mu_k$ is exact for Gaussian
iterates (Experiments~1 and~3, Appendix~\ref{app:amwgrad}) and is estimated with SVGD or blob kernels for particles
(Experiments~2, 3 and~4). MWGraD upudates the particles as $\textbf{x}^j_{k+1}=\textbf{x}^j_{k}+\eta\bar\varphi(\textbf{x}^j_{k})$. A-MWGraD gives each particle a velocity and update the particles as $\textbf{x}^j_{k+1}=\textbf{x}^j_{k}+a_k(\textbf{x}^j_{k}-\textbf{x}^j_{k-1})+\eta\bar\varphi(\textbf{x}^j_{k})$, in which the velocity is updated before the particles move.
Algorithm~2 of~\cite{nguyen2026accelerated} moves the particles with the previous velocity first; this order diverged in our tests
at step $1/L$, and Appendix~\ref{app:amwgrad} shows that it is stable only if $\eta\lambda_{\max}<1-a_k$. The heavy-ball update still
oscillates over several orders of magnitude, so we add a gradient-norm restart: the counter $j$ resets to $0$ whenever
$\|\bar\varphi_k\|_{L^2(\mu_k)}$ increases, and $a_k=(j-1)/(j+2)$; A-MWGraD-SC uses $a_k=\min(\beta',(j-1)/(j+2))$ with
$\beta'=(1-\sqrt{m\eta})/(1+\sqrt{m\eta})$. These changes make A-MWGraD stable, so they favour it. Our methods use the untuned
step $1/L$ in all experiments. The baselines use $1/L$ in Experiments~1 and~2, and a step tuned on the validation NLL in Experiment~3,
which favours them.

\begin{table}[t]
\centering
\caption{Experiment 1: iterations to reach $u_0<10^{-6}$ with step $1/L$.
MWAPG-C/SC: monotone version with restart (other versions in Table~\ref{tab:exp1-variants}).
Baselines: Gaussian versions; both A-MWGraD variants use our stabilization, which favours them
(Appendix~\ref{app:amwgrad}).}
\label{tab:exp1}
\small\setlength{\tabcolsep}{5pt}
\begin{tabular}{rrrrrrr}
\toprule
 & & \multicolumn{2}{c}{A-MWGraD} & & \multicolumn{2}{c}{MWAPG} \\
\cmidrule(lr){3-4}\cmidrule(lr){6-7}
$\kap$ & MWGraD & base & SC & MWPG & C & SC \\
\midrule
  8 &   27 & 13 & 16 &   28 & \textbf{7} & \textbf{7} \\
 16 &   56 & 16 & 16 &   55 & 10 & \textbf{7} \\
 40 &  132 & 18 & 18 &  135 & 15 & \textbf{7} \\
 80 &  264 & 24 & 25 &  268 & 21 & \textbf{7} \\
200 &  663 & 36 & 38 &  666 & 49 & \textbf{19} \\
400 & 1306 & 55 & 58 & 1328 & 47 & \textbf{23} \\
\bottomrule
\end{tabular}
\end{table}

\subsection{Experiment 1: ill-conditioned Gaussian targets}
\label{sec:exp1}
We use four Gaussian targets in $\R^2$, $\pi_i=\mathcal N(m_i,\Sigma_i)$, with means $m_i=c+1.5(\pm1,\pm1)$ around $c=(0,2)$
and $\Sigma_i=\mathrm{Rot}(\phi_i)\,\mathrm{diag}(4,s)\,\mathrm{Rot}(\phi_i)^\top$, $\phi_i\in\{45^\circ,135^\circ,225^\circ,315^\circ\}$. Each target
is a long, thin ellipse pointing from $c$ to its mean, so the targets disagree strongly, and the Pareto set is a genuine
three-parameter family. Here $L=1/s$ and $m=1/4$, so $\kap=4/s$ ranges from $8$ to $400$. We run all methods in closed form from
$\mu_0=\mathcal N((1,-2),\,0.25I)$, far from all targets, and evaluate $u_0$ exactly at every iteration. MWGraD and
A-MWGraD use the exact Gaussian score $\nabla\log\mu_k$, their most favourable setting. Their fields are then affine, so
their iterates stay Gaussian, and we update the mean and covariance in closed form (Appendix~\ref{app:amwgrad}).

\paragraph{Results.} Table~\ref{tab:exp1} shows the iteration counts, and Figure~\ref{fig:exp1-scaling}
(Appendix~\ref{app:exp1}) plots them against $\kap$.
\bitem{No momentum.} MWPG and MWGraD scale linearly in $\kap$, as the theory predicts, and are almost identical.
\bitem{MWAPG-C} roughly follows the $\sqrt{\kap}$ slope (7--49 iterations), although it does not use $m$. \bitem{MWAPG-SC} is almost
insensitive to the conditioning (7--23 iterations). It is $58\times$ faster than MWPG at $\kap=400$ and well within the
bound of Theorem~\ref{thm:sc}.
\bitem{Monotone safeguard.} With the plain momentum methods, $u_0$ rises by up to $2.6$ orders of magnitude. With the
safeguard of Algorithm~\ref{alg:mwapg}, every objective, and hence $u_0$, decreases, at similar iteration counts
(Table~\ref{tab:exp1-variants}).
\bitem{A-MWGraD with restart} is 2--24 times faster than MWGraD and comparable to MWAPG-C. However, even with our stabilization,
it is 1.9--3.4 times slower than MWAPG-SC. A-MWGraD-SC is not faster than A-MWGraD.

Appendix~\ref{app:exp1} shows the trajectories for $\kap=200$ (Figure~\ref{fig:exp1-curves}) and
for $\kap=8$ and $40$ (Figure~\ref{fig:app-exp1}), and the final Gaussians
(Figures~\ref{fig:app-shapes}, \ref{fig:app-shapes40} and~\ref{fig:app-shapes200}).
Appendix~\ref{app:exp1_stepsize} analyses the sensitivity to the step size, and
Appendix~\ref{app:exp1_high_dim} considers random targets in $\R^{10}$ and $\R^{50}$.
\begin{table}[t]
\centering
\caption{Experiment 3 (mean over 5 splits; Gaussian family): iterations $k_w$ until Pareto stationarity $w<10^{-4}$ (median),
final $w$, and test accuracy. Ours: 300 iterations; baselines: 3000 iterations, tuned step; A-MWGraD(-SC) with our stabilization. Particle methods and more metrics in
Table~\ref{tab:exp3}.}
\label{tab:exp3-main}
\footnotesize\setlength{\tabcolsep}{3pt}
\resizebox{\linewidth}{!}{\begin{tabular}{l rrc rrc}
\toprule
 & \multicolumn{3}{c}{Breast cancer ($\kap=587$)} & \multicolumn{3}{c}{Digits ($\kap=1169$)}\\
\cmidrule(lr){2-4}\cmidrule(lr){5-7}
Method & $k_w$ & final $w$ & Acc & $k_w$ & final $w$ & Acc\\
\midrule
MWPG & 699 & $7\cdot10^{-10}$ & 0.972 & 1207 & $2\cdot10^{-7}$ & 0.850 \\
\textbf{MWAPG-C} & 82 & $1\cdot10^{-10}$ & 0.972 & 178 & $6\cdot10^{-6}$ & 0.851 \\
\textbf{MWAPG-SC} & \textbf{72} & $<10^{-12}$ & 0.973 & \textbf{137} & $9\cdot10^{-10}$ & 0.851 \\
\midrule
MWGraD & 337 & $6\cdot10^{-6}$ & 0.972 & 699 & $3\cdot10^{-5}$ & 0.852 \\
A-MWGraD & 113 & $5\cdot10^{-6}$ & 0.971 & 163 & $3\cdot10^{-5}$ & 0.850 \\
A-MWGraD-SC & 113 & $5\cdot10^{-6}$ & 0.972 & 234 & $4\cdot10^{-5}$ & 0.831 \\
\bottomrule
\end{tabular}}
\end{table}
\subsection{Experiment 2: non-Gaussian Pareto set}
\label{sec:exp2}

We need a problem that requires particles but still satisfies Assumption~\ref{ass:1}. We use
\[
\begin{aligned}
  V_i(x)&=\tfrac12(x-m_i)^\top\Sigma_i^{-1}(x-m_i)+g(x),\\
  g(x)&=\textstyle\sum_{j=1}^3\mathrm{sp}\big(b(\mathbf n_j^\top(x-c)-1)\big),
\end{aligned}
\]
with $\mathrm{sp}=\mathrm{softplus}$, $m_i$ and $c$ as in Experiment~1, and $\Sigma_i=\mathrm{Rot}(\phi_i)\mathrm{diag}(4,1)\mathrm{Rot}(\phi_i)^\top$.
Here $g$ is a smooth triangular barrier, with unit normals $\mathbf n_j$ at $90^\circ,210^\circ,330^\circ$. Its Hessian is
bounded by $\frac{b^2}4\sum_j\mathbf n_j\mathbf n_j^\top=\frac{3b^2}8I$. Hence $m=1/4$ and $L=1+\frac38b^2$, and $b\in\{4,8,16\}$ gives
$\kap\in\{28,100,388\}$.

Since $\sum_i\lambda_i=1$, the barrier enters every Pareto point $\pi_{\boldsymbol{\lambda}}\propto e^{-\sum_i\lambda_iV_i}$ with full
weight, so all Pareto points are triangle-shaped (Figure~\ref{fig:exp2-snap}). For the three values of
$\kap$, the best Gaussian approximation of the Pareto set has $\min_{G,{\boldsymbol{\lambda}}}\mathrm{KL}(G\|\pi_{\boldsymbol{\lambda}})=0.049$, $0.147$
and $0.293$. No Gaussian method can beat this.

All methods use the same 500 particles, $X^0(\xi_j)=\frac12\xi_j+(1,-2)$. Our methods use ICNNs
of width 32 and run 60, 100 or 150 iterations. The kernel methods use SVGD fields with the median
bandwidth, and blob fields~\cite{carrillo2019blob} with the full two-term formula, a Mahalanobis
kernel and $0.3\times$ the median bandwidth. They run 2000 iterations. With the default
bandwidth, the blob particles collapsed onto the edges of the triangle; our choice prevents this.

\paragraph{Metric.}
Kernel particles are not independent, and nearest-neighbour entropy estimators~\cite{kozachenko1987sample} returned negative KL
values for them. Therefore, we report the energy distance~\cite{szekely2013energy} to the nearest Pareto point,
$\min_{\boldsymbol{\lambda}}\mathrm{ED}(\mu_k,\pi_{\boldsymbol{\lambda}})$, with ${\boldsymbol{\lambda}}$ on a simplex grid of step $1/8$ and each
$\pi_{\boldsymbol{\lambda}}$ sampled exactly on a fine spatial grid. 500 exact i.i.d.\ samples give
$(1.6\text{--}2.3)\cdot10^{-3}$. A run has converged when $\mathrm{ED}<3\cdot10^{-3}$.

\paragraph{Results.}
Figure~\ref{fig:exp2-curves} (Appendix~\ref{app:exp2}) compares each plain method with its accelerated versions.
Figure~\ref{fig:exp2-speedup} and Table~\ref{tab:exp2} (Appendix~\ref{app:exp2}) give the iteration counts and speed-ups.
\bitem{Accuracy.} All accelerated methods reach the level of exact i.i.d.\ samples.
\bitem{Acceleration helps in every family, and it helps more for larger $\kap$.} The speed-up of MWAPG-C over MWPG grows
from $\times1.9$ ($\kap=28$) to more than $\times6.8$ ($\kap=388$). For blob, it grows from $\times2.0$ to $\times7.3$, and for
SVGD it is $\times3$--$15$.
\bitem{Iteration counts.} Our accelerated methods need 12--22 iterations, the fewest of all, except MWAPG-SC at $\kap=388$ (35).
The accelerated blob methods need 20--30 iterations, but only with the reduced bandwidth.
\bitem{Monotone safeguard.} Without it, the energy distance of MWAPG-SC rises by a factor of 3.3--3.8 at $\kap=100$ and
$388$. With it, the curves decrease monotonically (Table~\ref{tab:exp2}). Here MWAPG-SC is not always faster than MWAPG-C. Its
momentum is tuned to the global $(m,L)$, but the curvature of the barrier varies in space.
The snapshots (Figure~\ref{fig:exp2-snap}) show that the accelerated particles cover the Pareto density after 25 iterations,
while those of MWPG are still entering the triangle.

\bitem{Cost.}
In wall-clock time, the kernel methods are faster on this small problem: $0.015$\,s per step, against $1.1$--$2.0$\,s for an
ICNN proximal step (Appendix~\ref{app:exp2}). However, our step needs no bandwidth, and the bandwidth was the critical parameter
for the kernel methods.

% requires \usepackage{booktabs}
\begin{table}[t]
\caption{Experiment 4: average test accuracy (\%), mean $\pm$ s.d. over the last 10 evaluations. Best result in each row in bold.}
\label{tab:exp4-full}
\centering
\small
\setlength{\tabcolsep}{3pt}
\begin{tabular}{lcccc}
\toprule
Kernel & MWGraD & MWPG & A-MWGraD & MWAPG \\
\midrule
\multicolumn{5}{l}{\textit{Multi-MNIST}} \\
\quad SVGD & 86.50{\scriptsize$\pm$0.23} & 86.50{\scriptsize$\pm$0.23} & \textbf{94.39}{\scriptsize$\pm$0.02} & 94.33{\scriptsize$\pm$0.03} \\
\quad Blob & 90.80{\scriptsize$\pm$0.24} & 90.89{\scriptsize$\pm$0.26} & 93.58{\scriptsize$\pm$0.06} & \textbf{94.03}{\scriptsize$\pm$0.09} \\
\midrule
\multicolumn{5}{l}{\textit{Multi-Fashion}} \\
\quad SVGD & 76.19{\scriptsize$\pm$0.34} & 76.19{\scriptsize$\pm$0.33} & 86.30{\scriptsize$\pm$0.09} & \textbf{86.42}{\scriptsize$\pm$0.10} \\
\quad Blob & 81.29{\scriptsize$\pm$0.38} & 81.31{\scriptsize$\pm$0.34} & 84.97{\scriptsize$\pm$0.09} & \textbf{86.24}{\scriptsize$\pm$0.04} \\
\midrule
\multicolumn{5}{l}{\textit{Multi-F+M}} \\
\quad SVGD & 81.58{\scriptsize$\pm$0.26} & 81.60{\scriptsize$\pm$0.25} & 91.20{\scriptsize$\pm$0.18} & \textbf{91.31}{\scriptsize$\pm$0.05} \\
\quad Blob & 86.77{\scriptsize$\pm$0.44} & 86.87{\scriptsize$\pm$0.20} & 87.47{\scriptsize$\pm$0.19} & \textbf{89.90}{\scriptsize$\pm$0.22} \\
\bottomrule
\end{tabular}
\end{table}

\subsection{Experiment 3: multi-client Bayesian logistic regression on real data}
\label{sec:exp3}

\paragraph{Setting.}
We use Bayesian logistic regression with two real datasets: breast cancer ($K=4$ clients) and digits ($K=5$ clients). Each
client has its own posterior $\pi_i$. Each $V_i$ is strongly convex and smooth, so Assumption~\ref{ass:A1} holds. We run our methods in the
Gaussian family, so our theorems apply without change. We compare them with MWGraD, A-MWGraD and A-MWGraD-SC in the same
family, and with particle methods based on SVGD and blob kernels.

\paragraph{Results.}
Table~\ref{tab:exp3-main} summarizes the results. MWAPG-C and MWAPG-SC reach Pareto stationarity
$w<10^{-4}$ $7$--$10\times$ faster than MWPG. A-MWGraD with restart needs 113 and 163 iterations,
but all Gaussian baselines then stall at $w\approx5\cdot10^{-6}$ (breast cancer) and $3$--$4\cdot10^{-5}$
(digits). MWAPG-SC goes further and reaches $w<10^{-8}$ in 178 and 263 iterations. Our solutions closely
match the exact Pareto posterior, while the blob particles collapse. Appendix~\ref{app:exp3main} gives
the details and the full results (Table~\ref{tab:exp3}).

\subsection{Experiment 4: Bayesian multi-task learning with neural networks}
\label{sec:exp4}
\paragraph{Setting.}
We use the Bayesian multi-task benchmark of~\citet{phan2022stochastic} on Multi-MNIST, Multi-Fashion and Multi-Fashion+MNIST, with two classification tasks per image. The posterior is represented by $n=5$ networks with a shared encoder and task-specific heads; the multiobjective methods update the encoders, with one target per task. Our methods use the same kernel approximation of the entropy (SVGD or Blob) as the baselines: MWGraD and A-MWGraD, so this experiment is outside our theory. MWAPG is the plain version (no-monotone), because the safeguard needs objective values, which are not available. 
%See %Appendix~\ref{app:exp4} for details.

\paragraph{Results.} Table~\ref{tab:exp4-full} summarizes the results. MWPG and MWGraD are
indistinguishable (at most $0.10$\,pp apart).
MWAPG is the most accurate method in five of six settings and is within $0.06$\,pp of
A-MWGraD in the sixth. It is $3.1$--$10.2$\,pp above MWGraD and has the lowest test error
over the first $2000$ iterations in all six settings. Its advantage over A-MWGraD depends
on the kernel: the two are on par with SVGD (within $0.12$\,pp), whereas with the Blob kernel
MWAPG is ahead by $0.45$--$2.43$\,pp, with the gap growing with task difficulty. See Appendix~\ref{app:exp4} for details.

%=====================================================================================================================
\section{Conclusion}
\label{sec:conclusion}

We proposed discrete-time accelerated proximal gradient methods for multiobjective
optimization over probability measures. They achieve rates $O(1/k^2)$ and
$O((1-\sqrt{m/L})^k)$ for the distance to the Pareto set.
The key idea is lifting measures to transport maps in a Hilbert space.
It gives momentum a meaning and makes all objectives convex on a closed convex cone.
%Combining the anchored subproblem of~\cite{tanabe2023accelerated} with a proximal entropy step
%removes the need for score estimation.
Experiments confirm the predicted dependence on the condition number, and the method
remains competitive in multi-task learning with neural networks.
However, our theory requires log-concave targets and does not certify the network's
optimization error. We will address these limitations in future work. %In Experiment~4, the entropy is approximated by kernels, so our guarantees do not apply there. With Blob, the momentum methods need a smaller learning rate. In low dimension, particle steps cost more than kernel steps.

%\paragraph{Limitations.}
%Our theory needs log-concave targets, and it does not certify the optimization error of a neural network. In Experiment~4, the entropy is approximated by kernels, so our guarantees do not apply there. With Blob, the momentum methods need a smaller learning rate. In low dimension, particle steps cost more than kernel steps.

\section*{AI Use Statement}
We used a large language model (LLM) to revise and condense the text
of several sections. The research ideas, methods, theorems, proofs and experiments are the authors' own, and the
authors take full responsibility for the content of the paper.

\section*{Impact Statement}

This paper presents work whose goal is to advance the field of Machine
Learning. There are many potential societal consequences of our work, none
which we feel must be specifically highlighted here.

% In the unusual situation where you want a paper to appear in the
% references without citing it in the main text, use \nocite

\bibliography{main}
\bibliographystyle{icml2027}

%%%%%%%%%%%%%%%%%%%%%%%%%%%%%%%%%%%%%%%%%%%%%%%%%%%%%%%%%%%%%%%%%%%%%%%%%%%%%%%
%%%%%%%%%%%%%%%%%%%%%%%%%%%%%%%%%%%%%%%%%%%%%%%%%%%%%%%%%%%%%%%%%%%%%%%%%%%%%%%
% APPENDIX
%%%%%%%%%%%%%%%%%%%%%%%%%%%%%%%%%%%%%%%%%%%%%%%%%%%%%%%%%%%%%%%%%%%%%%%%%%%%%%%
%%%%%%%%%%%%%%%%%%%%%%%%%%%%%%%%%%%%%%%%%%%%%%%%%%%%%%%%%%%%%%%%%%%%%%%%%%%%%%%
\newpage
\appendix
\onecolumn

%---------------------------------------------------------------------------------------------------------------------
\section{Notation}
\label{app:notation}

\begin{center}
\small
\begin{tabular}{ll}
\toprule
Symbol & Meaning\\
\midrule
$\mathcal{P}_2$, $\mathcal{P}_2^{\mathrm{ac}}$, $\mathcal{W}_2$ & measures with finite second moment, absolutely continuous ones, Wasserstein-2 distance\\
$\pi_i=e^{-V_i}/Z_i$ & targets, $i=1,\dots,K$, with potentials $V_i$\\
$G_i(\mu)=\mathrm{KL}(\mu\|\pi_i)=F_i(\mu)+H(\mu)+\log Z_i$ & objectives; $F_i(\mu)=\int V_i\,d\mu$, $H(\mu)=\int\mu\log\mu$\\
$\nu$, $\Ltwo$ & base measure ($=\mu_0$ in the analysis), Hilbert space of maps $\mathcal{L}^2(\nu;\R^d)$\\
$\ip XY=\int\langle X(x),Y(x)\rangle\,d\nu(x)$, $\|X\|_\nu=\ip XX^{1/2}$ & inner product and norm of $\Ltwo$\\
$\langle a,b\rangle$, $\|a\|_2$ & Euclidean inner product and norm on $\R^d$ or $\R^K$\\
$\mu_X=X\push\nu$ & measure represented by the map $X$\\
$\tF_i,\tH,\tG_i$ & lifted functionals, $\tF_i(X)=F_i(\mu_X)$ etc.\\
$\mathcal{K}_\nu$, $\iota_\mathcal{K}$ & cone of optimal transport maps from $\nu$, its indicator\\
$h=\tH+\iota_\mathcal{K}$ & convex nonsmooth part on $\Ltwo$\\
$Z_\rho$ & the unique map in $\mathcal{K}_\nu$ with $(Z_\rho)\push\nu=\rho$\\
$\Delt$ & unit simplex of $\R^K$; $\mathbf 1$ the all-ones vector\\
$\pi_{\boldsymbol{\lambda}}=e^{-V_{\boldsymbol{\lambda}}}/Z_{\boldsymbol{\lambda}}$, $V_{\boldsymbol{\lambda}}=\sum_i\lambda_iV_i$ & geometric mean of the targets\\
$u_0$ & merit function~\eqref{eqn:defofmeritfunction}\\
$\mathcal{P}^*$, $\mathcal{S}_0$, $R$ & weakly Pareto set, initial level set, Pareto radius~\eqref{eq:R}\\
$\mathcal{S}_S$, $R_S$ & enlarged level set $\{\rho:G_i(\rho)\le G_i(\mu_0)+S\ \forall i\}$ and its radius; \textbf{(A3$_S$)}: \textbf{(A3)} on $\mathcal{S}_S$\\
$L$, $m$, $\kap=L/m$, $\theta=\sqrt{m/L}$, $\beta=\frac{1-\theta}{1+\theta}$ & smoothness, strong convexity, condition number, momentum\\
$t_k$, $\gamma_k$ & FISTA sequence ($t_0=0$, $t_{k+1}^2-t_{k+1}=t_k^2$), momentum coefficient\\
$Y^k$, $g_i=\nabla V_i\circ Y^k$, $c_i$ & extrapolated point, gradients at it, offsets in~\eqref{eq:psi}\\
$Q=(\ip{g_i}{g_j})_{ij}$, $\mathcal G\boldsymbol{\lambda}=\sum_i\lambda_ig_i$ & Gram matrix, $\mathcal G^*\mathcal G=Q$\\
$\psi_i$, $\Phi$, $q$, $Z({\boldsymbol{\lambda}})$ & subproblem models~\eqref{eq:psi}, primal~\eqref{eqn:proximalstep}, dual, proximal map~\eqref{eqn:dualstep}\\
$\Phi_{\boldsymbol{\lambda}}(Z)=\sum_i\lambda_i\psi_i(Z)+\frac L2\|Z-Y^k\|_\nu^2$ & Lagrangian; $\Phi=\max_{\boldsymbol{\lambda}\in\Delt}\Phi_{\boldsymbol{\lambda}}$, $q(\boldsymbol{\lambda})=\min_{Z\in\mathcal{K}_\nu}\Phi_{\boldsymbol{\lambda}}(Z)$\\
$\sigma_k=\min_i[\tG_i(X^k)-\tG_i(Z)]$ & gap to a fixed comparison map $Z\in \mathcal{K}_\nu$\\
$a_k=\|X^k-X^{k-1}\|_\nu^2$ & squared step length\\
$U^k$ & far point (defined separately in the convex and strongly convex proofs)\\
\bottomrule
\end{tabular}
\end{center}

\section{Proof of Lemma \ref{lemma:liftedproblem}}
\label{proof:liftedproblem}
Throughout, $\nu\in\mathcal{P}_2^{\mathrm{ac}}(\mathbb{R}^d)$. For $X,X^\prime\in \mathcal{L}^2(\nu)$, the plan $(X, X^\prime)_{\#}\nu$ couples $\mu_{X}$ and $\mu_{X^\prime}$, so
\begin{equation}
\label{eqn:12}
    \mathcal{W}_2(\mu_X, \mu_{X^\prime})\leq \lVert X-X^\prime\rVert_\nu, \text{ and }\int\lVert x\rVert^2_2d\mu_X=\lVert X \rVert^2_\nu.
\end{equation}

\noindent
\textbf{(a) The cone $\mathcal{K}_{\nu}$.} \textit{convex cone.} if $Z_0\in \partial\varphi_0$ and $Z_1\in \partial\varphi_1$ $\nu$-a.e. and $\alpha,\beta\geq 0$, then $\alpha Z_0 + \beta Z_1\in \alpha\partial\varphi_0+\beta\partial\varphi_1\subset \partial(\alpha \varphi_0 + \beta \varphi_1)$ $\nu$-a.e., and $\alpha \varphi_0 +\beta \varphi_1$ is proper (both are finite $\nu$-a.e., hence on a common point), lsc and convex. Hence $\mathcal{K}_{\nu}$ is a convex cone. 

\textit{Closed.} Let $Z_n\in \mathcal{K}_\nu, Z_n \rightarrow Z$ in $\mathcal{L}^2(\nu)$. Passing to subsequence, $Z_n \rightarrow Z$ pointwise on a set $\Omega$ with $\nu(\Omega)=1$, and we may assume that $Z_n(x)\in \partial\varphi_n(x)$ for all $x\in \Omega$ and all $n$. The graph $\left\{(x, Z_n(x)):x\in \Omega\right\}$ is cyclically monotone: for every finite cycle $x_1, \dots, x_{N+1}=x_1\in \Omega$, $\sum_{j=1}^{N}\langle Z_n(x_j), x_{j+1}-x_j \rangle\leq \sum_{j=1}^{N} \left[\varphi_n(x_{j+1})-\varphi_n(x_{j}) \right]=0$. Letting $n\rightarrow \infty$, the graph of $Z$ on $\Omega$ is cyclically monotone. By Rockafellar's theorem \cite{rockafellar1997convex}, there is a proper lsc convex $\varphi$ with $Z(x)\in \partial\varphi(x)$ for all $x\in \Omega$. Hence $Z\in \mathcal{K}_\nu$.

\noindent
\textbf{(b) The potential energies.} By \textbf{(A1)}, for all $x,y\in \mathbb{R}^d$,
\begin{equation}
\label{eqn:13}
    V_i(y) + \langle\nabla V_i(y), x-y \rangle + \frac{m}{2}\lVert x-y\rVert^2_2\leq V_i(x)\leq V_i(y) + \langle\nabla V_i(y), x-y \rangle + \frac{L}{2}\lVert x-y\rVert^2_2,
\end{equation}
with $m=0$ for the convex case. Taking $y=0$ shows $|V_i(x)|\leq C(1+\lVert x\rVert^2_2)$, so $\Tilde{F}_i$ is finite on $\mathcal{L}^2(\nu)$, and $\nabla V_i\circ X\in \mathcal{L}^2(\nu)$ because $\lVert \nabla V_i(x) \rVert_2\leq \lVert \nabla V_i(0) \rVert_2 + L\lVert x \rVert_2$. Substituting $x=X(\zeta), y=Y(\zeta)$, where $X, Y\in\mathcal{L}^2(\nu)$ in \eqref{eqn:13} and integrating against $\nu$ gives
\begin{equation}
    \Tilde{F}_i(Y) + \langle\nabla V_i\circ Y, X-Y \rangle_\nu + \frac{m}{2}\lVert X-Y \rVert^2_\nu \leq \Tilde{F}_i(X) \leq \Tilde{F}_i(Y) + \langle\nabla V_i\circ Y, X-Y \rangle_\nu + \frac{L}{2}\lVert X-Y \rVert^2_\nu.
\end{equation}
The upper bound shows Frechet differentiability with $\nabla \Tilde{F}_i(Y)= \nabla V_i\circ Y$; the lower bound is $m$-strong convexity; and $\lVert \nabla V_i\circ X-\nabla V_i\circ Y\rVert_\nu\leq L\lVert X- Y\rVert_\nu$ is $L$-smoothness.

\noindent
\textbf{(c) The entropy.} \textit{Convexity on $\mathcal{K}_\nu$}. Let $Z_0, Z_1\in \mathcal{K}_\nu$, and $Z_s = (1-s) Z_0 + s Z_1$. Since $Z_0$ and $Z_1$ are optimal maps from $\nu$, the curve $s\mapsto (Z_s)_{\#}\nu$ is a generalized geodesic with base $\nu$ joining $\mu_{Z_0}$ and $\mu_{Z_1}$ \cite{ambrosio2005gradient}. The negative entropy is convex along generalized geodesic. Hence $\Tilde{H}(Z_s) \leq (1-s)\Tilde{H}(Z_0) + s \Tilde{H}(Z_1)$.

\textit{Non-convexity on $\mathcal{L}^2(\nu)$}. We take an example to show this. On $\mathbb{R}$ with $\nu =\mathcal{N}(0,1)$, the maps $X_{+}(\zeta)=\zeta$ and $X_{-}(\zeta)=-\zeta$ push $\nu$ to $\nu$, so $\Tilde{H}(X_+)=\Tilde{H}(X_{-})=H(\nu)$, but $\frac{1}{2}(X_+ + X_{-})=0$ pushes $\nu$ to $\delta_0$ and $\Tilde{H}(0)=+\infty$.

\textit{Lower semicontinuous.} Let $Z_n\rightarrow Z$ in $\mathcal{K}_\nu$. By \eqref{eqn:12}, $\mu_{Z_n}\rightarrow \mu_Z$ in $\mathcal{W}_2$, hence narrowly with uniformly bounded second moment. The negative entropy is lsc for narrow convergence on sets with bounded second moment \cite{ambrosio2005gradient}, so $\Tilde{H}(Z)\leq \lim \inf_n \Tilde{H}(Z_n)$.

\noindent
\textbf{(d) Lower bound.} Let $X\in \mathcal{L}^2(\nu)$ with $\mu=\mu_X$ absolutely continuous, $M=\lVert X \rVert_\nu^2$, and $\gamma_\sigma=\mathcal{N}(0, \sigma^2 I)$. Then $0\leq \mathrm{KL}(\mu\| \gamma_\sigma)=H(\mu)+\frac{d}{2}\log (2\pi \sigma^2)+\frac{M}{2\sigma^2}$. Choosing $\sigma^2 = M/d$ gives $H(\mu)\geq -\frac{d}{2}\log (2\pi M/d)$.

\section{The merit function and Assumption (A3)}
\subsection{Proof of Proposition \ref{proposition:meritforklobjective}}
\label{appendix:proofmeritfunction}
We divide the proof of Proposition \ref{proposition:meritforklobjective} into five steps.

\noindent
\textbf{Step 1: the geometric means.} Let $\boldsymbol{\lambda}\in \Delta_K$ and $V_{\boldsymbol{\lambda}}=\sum_i \lambda_iV_i$, which is convex and $L$-smooth. By Holder's inequality, $Z_{\boldsymbol{\lambda}}=\int\Pi_i e^{\lambda_i V_i(x)}dx\leq \Pi_i Z_i^{\lambda_i}< \infty$, so $\pi_{\boldsymbol{\lambda}}=e^{V_{\boldsymbol{\lambda}}}/Z_{\boldsymbol{\lambda}}$ is a log-concave probability density. Log-concave probability measure have exponential tails \cite{saumard2014log}, so $\pi_{\boldsymbol{\lambda}}\in \mathcal{P}_2^{\mathrm{ac}}$ and $H(\pi_{\boldsymbol{\lambda}})$ is finite. By the weighted AM-GM inequality, $e^{-V_{\boldsymbol{\lambda}}}\leq \sum_i \lambda_i e^{-V_i}\leq \sum_i e^{-V_i}$, so by dominated convergence $\boldsymbol{\lambda}\mapsto \log Z_{\boldsymbol{\lambda}}$ is continuous on $\Delta_K$; it is convex by Holder's inequality.

\noindent
\textbf{Step 2: scalarization.} For $\rho\in \mathcal{P}_2$, the integrals $F_i(\rho)$ are finite (see Appendix \ref{proof:liftedproblem} (b)), so either $H(\rho)=+\infty$ and $G_i(\rho)=+\infty$ for all $i$, or all $G_i(\rho)$ are finite. In the latter case,
\begin{equation}
\label{eqn:scalarization}
    \sum_i \lambda_iG_i(\rho)= H(\rho) + \int V_{\boldsymbol{\lambda}}d\rho + \sum_i \lambda_i \log Z_i = \mathrm{KL}(\rho\| \pi_{\boldsymbol{\lambda}}) + c_{\boldsymbol{\lambda}}, c_{\boldsymbol{\lambda}}=\sum_i \lambda_i \log Z_i - \log Z_{\boldsymbol{\lambda}}\geq 0.
\end{equation}
In particular, $\inf_{\rho\in\mathcal{P}_2}\sum_i \lambda_i G_i=c_{\boldsymbol{\lambda}}$, attained only at $\rho=\pi_{\boldsymbol{\lambda}}$.

\noindent
\textbf{Step 3: minimax.} We fix $\mu$ with all $G_i(\mu)$ finite and let $D=\left\{\rho\in \mathcal{P}_2: H(\rho)<\infty \right\}$, which is convex under mixtures. Measures outside $D$ give the value $-\infty$ in \eqref{eqn:defofmeritfunction}, so the supremum can be restricted to $D$. We consider $f(\boldsymbol{\lambda}, \rho)=\sum_i \lambda_i \left[G_i(\mu)-G_i(\rho) \right]$ over $\Delta_K \times D$. It is linear and continuous in $\boldsymbol{\lambda}$, and concave in $\rho$ under mixtures, because $H$ is convex and $F_i$ is linear under mixtures. Since $\Delta_K$ is compact and convex, the minimax theorem \cite{fan1953minimax} gives
\begin{equation*}
    u_0(\mu) = \sup_{\rho\in D}\min_{\boldsymbol{\lambda}\in\Delta_K} f(\boldsymbol{\lambda}, \rho)=\min_{\boldsymbol{\lambda}\in\Delta_K}\sup_{\rho\in D} f(\boldsymbol{\lambda}, \rho),
\end{equation*}
where we used $\min_i f_i = \min_{\boldsymbol{\lambda}\in\Delta_K}\sum_i \lambda_i f_i$. By \eqref{eqn:defofmeritfunction}, $\sup_{\rho\in D} f(\boldsymbol{\lambda}, \rho)=\mathrm{KL}(\mu\|\pi_{\boldsymbol{\lambda}})$. The map $\boldsymbol{\lambda}\mapsto \mathrm{KL}(\mu\|\pi_{\boldsymbol{\lambda}})= H(\mu) + \sum_i \lambda_i F_i(\mu) + \log Z_{\boldsymbol{\lambda}}$ is continuous on $\Delta_K$ by \textbf{Step 1}, so the minimum is attained. This proves \eqref{eqn:meritfunction}.

\noindent
\textbf{Step 4: the Pareto set.} If all $G_i(\mu)$ are finite, $u_0(\mu)=0$ if and only if $\mathrm{KL}(\mu\|\pi_{\boldsymbol{\lambda}})=0$ for some $\boldsymbol{\lambda}\in \Delta_K$, i.e., $\mu=\pi_{\boldsymbol{\lambda}}$. If $G_i(\mu)=+\infty$ for all $i$, then $\mu$ is strictly dominated by $\pi_{\boldsymbol{\lambda}}$ for any $\boldsymbol{\lambda}$. Hence $\mathcal{P}^*= \left\{\pi_{\boldsymbol{\lambda}}: \boldsymbol{\lambda}\in \Delta_K \right\}$.

\noindent
\textbf{Step 5: Gaussian targets.} If $V_i(x)= \frac{1}{2}(x-m_i)^\top P_i (x-m_i)$, then $V_{\boldsymbol{\lambda}}(x)=\frac{1}{2}x^\top P_{\boldsymbol{\lambda}}x-x^\top \sum_i \lambda_i P_im_i + \mathrm{const}$, which is the potential of $\mathcal{N}(m_{\boldsymbol{\lambda}}, P_{\boldsymbol{\lambda}}^{-1})$. For a Gaussian $\mu=\mathcal{N}(b, \Sigma)$,
\begin{equation}
    u_0(\mu) = \min_{\boldsymbol{\lambda}\in \Delta_K} \frac{1}{2}\left[\mathrm{tr}(P_{\boldsymbol{\lambda}}\Sigma)-d-\log \det (P_{\boldsymbol{\lambda}}\Sigma) +(b-m_{\boldsymbol{\lambda}})^\top P_{\boldsymbol{\lambda}}(b-m_{\boldsymbol{\lambda}})\right],
\end{equation}
a convex problem in $\boldsymbol{\lambda}$ which can be solved numerically to evaluate $u_0$ exactly.

\subsection{Assumption (A3)}
\label{app:A3}

\begin{lemma}[Assumption (A3) in the strongly convex case]\label{lem:A3sc}
If every $V_i$ is $m$-strongly convex, then Assumption (A3$_S$) holds for every $S\ge0$, and $R_S\le\frac2m\max_i\mathrm{KL}(\mu_0\|\pi_i)$. In particular, (A3) holds and $R\le\frac2m\max_i\mathrm{KL}(\mu_0\|\pi_i)$.
\end{lemma}

\begin{proof}
\emph{Radius.} Every $\rho\in\mathcal{P}^*$ is some $\pi_{\boldsymbol{\lambda}}$ (Proposition~\ref{proposition:meritforklobjective}), and $\pi_{\boldsymbol{\lambda}}$ is $m$-strongly log-concave since $V_{\boldsymbol{\lambda}}$ is
$m$-strongly convex. Talagrand's inequality~\cite{otto2000generalization} gives
$\mathcal{W}_2^2(\mu_0,\pi_{\boldsymbol{\lambda}})\le\frac2m\mathrm{KL}(\mu_0\|\pi_{\boldsymbol{\lambda}})$, and by~\eqref{eqn:scalarization},

$\mathrm{KL}(\mu_0\|\pi_{\boldsymbol{\lambda}})=\sum_i\lambda_iG_i(\mu_0)-c_{\boldsymbol{\lambda}}\le\max_iG_i(\mu_0)=\max_i\mathrm{KL}(\mu_0\|\pi_i)$.
This bound holds for every Pareto point, so it bounds $R_S$ for every $S\ge0$.

\emph{Dominating Pareto points.} Let $\rho\in\mathcal{P}_2$ with finite $G_i(\rho)$ for all $i$, and let $\mathcal{D}_\rho=\{\rho':G_i(\rho')\le G_i(\rho)\ \forall i\}$.
Sublevel sets of $\mathrm{KL}(\cdot\|\pi_1)$ are narrowly compact~\cite{dupuis2011weak}, and every $G_i$ is narrowly lsc, so $\mathcal{D}_\rho$ is a nonempty narrowly compact set, and the narrowly lsc function $\frac1K\sum_iG_i$ attains its minimum
on $\mathcal{D}_\rho$ at some $\rho^*$. If some $\rho''$ satisfied $G_i(\rho'')<G_i(\rho^*)$ for all $i$, then $\rho''\in \mathcal{D}_\rho$ with a smaller
value of $\sum_iG_i$, a contradiction. 
Hence $\rho^*$ is weakly Pareto optimal and weakly dominates $\rho$. This argument works for every $\rho$ with finite $G_i(\rho)$, so it works on $\mathcal{S}_S$ for every $S\ge0$. Finally, $\mathcal{W}_2^2(\rho^*,\pi_1)\le\frac2m
\mathrm{KL}(\rho^*\|\pi_1)<\infty$ gives $\rho^*\in\mathcal{P}_2$.
\end{proof}

In the merely convex case, $R\le\max_{\boldsymbol{\lambda}\in\Delta_K}\mathcal{W}_2^2(\mu_0,\pi_{\boldsymbol{\lambda}})$, and this maximum is finite whenever
$\boldsymbol{\lambda}\mapsto\pi_{\boldsymbol{\lambda}}$ is $\mathcal{W}_2$-continuous, e.g.\ when the $V_i$ grow at least quadratically outside a ball (then
$e^{-V_{\boldsymbol{\lambda}}(x)}\le Ce^{-c\|x\|_2^2}$ uniformly in $\boldsymbol{\lambda}$ and dominated convergence applies). We do not pursue minimal conditions.

\section{The subproblem: proof of Lemma~\ref{lem:sub}}
\label{app:sub}

Fix $X^k\in \mathcal{K}_\nu$ with $\tH(X^k)<\infty$ and $Y\in\Ltwo$, and let $c_i=\tF_i(Y)-\tG_i(X^k)$, a finite constant. Let
$\iota_\mathcal{K}$ be the indicator of $\mathcal{K}_\nu$ and $h=\tH+\iota_\mathcal{K}$ on $\Ltwo$. By Lemma~\ref{lemma:liftedproblem}, $h$ is convex (since $\mathcal{K}_\nu$ is
convex and $\tH$ is convex on it), lsc (since $\mathcal{K}_\nu$ is closed and $\tH$ is lsc on it), proper
($h(\Id)=H(\nu)<\infty$, and $h>-\infty$ by Lemma~\ref{lemma:liftedproblem}(d)).

For ${\boldsymbol{\lambda}}\in\Delt$ define the Lagrangian $\Phi_{\boldsymbol{\lambda}}$,
\[
  \Phi_{\boldsymbol{\lambda}}(Z)=\sum_i\lambda_i\psi_i(Z)+\tfrac L2\|Z-Y\|_\nu^2=h(Z)+\Big\langle\sum_i\lambda_ig_i,\,Z-Y\Big\rangle_\nu+\sum_i\lambda_ic_i
  +\tfrac L2\|Z-Y\|_\nu^2 ,
\]
and note that, with $T_{\boldsymbol{\lambda}}=Y-\frac1L\sum_i\lambda_ig_i$, completing the square gives
\begin{equation}\label{eq:square}
  \Phi_{\boldsymbol{\lambda}}(Z)=h(Z)+\tfrac L2\|Z-T_{\boldsymbol{\lambda}}\|_\nu^2+\sum_i\lambda_ic_i-\tfrac1{2L}\Big\|\sum_i\lambda_ig_i\Big\|_\nu^2 .
\end{equation}

\paragraph{(a) Existence and uniqueness.}
$\Phi+\iota_\mathcal{K}=h+\max_i(\ip{g_i}{\cdot-Y}+c_i)+\frac L2\|\cdot-Y\|_\nu^2$ is the sum of a proper convex lsc function, a finite
convex continuous function, and an $L$-strongly convex continuous function. It is therefore proper, lsc and $L$-strongly
convex on the Hilbert space $\Ltwo$, and has a unique minimizer~\cite{bauschke2020correction}. The same argument
applied to~\eqref{eq:square} shows that $Z({\boldsymbol{\lambda}})=\prox_{h/L}(T_{\boldsymbol{\lambda}})$ is well defined, i.e.\ \eqref{eqn:dualstep}. Moreover, the
proximal map of a proper convex lsc function is firmly nonexpansive~\cite{bauschke2020correction}, so
\begin{equation}\label{eq:nonexp}
  \|Z({\boldsymbol{\lambda}})-Z({\boldsymbol{\lambda}}')\|_\nu\le\|T_{\boldsymbol{\lambda}}-T_{{\boldsymbol{\lambda}}'}\|_\nu=\tfrac1L\Big\|\sum_i(\lambda_i-\lambda'_i)g_i\Big\|_\nu.
\end{equation}

\paragraph{(b) Strong duality.}
Let $D_h=\{Z\in \mathcal{K}_\nu:\tH(Z)<\infty\}$, a convex set on which every $\Phi_{\boldsymbol{\lambda}}$ is real-valued. For each ${\boldsymbol{\lambda}}$, $\Phi_{\boldsymbol{\lambda}}$ is
convex and lsc on $D_h$; for each $Z$, $\boldsymbol{\lambda}\mapsto\Phi_{\boldsymbol{\lambda}}(Z)$ is linear on the compact convex set $\Delt$. By Sion's minimax
theorem~\cite{sion1958general},
\[
  \min_{Z\in \mathcal{K}_\nu}\Phi(Z)=\inf_{Z\in D_h}\max_{{\boldsymbol{\lambda}}\in\Delt}\Phi_{\boldsymbol{\lambda}}(Z)=\max_{{\boldsymbol{\lambda}}\in\Delt}\inf_{Z\in D_h}\Phi_{\boldsymbol{\lambda}}(Z)
  =\max_{{\boldsymbol{\lambda}}\in\Delt}q({\boldsymbol{\lambda}}),
\]
using $\max_i\psi_i=\max_{{\boldsymbol{\lambda}}\in\Delt}\sum_i\lambda_i\psi_i$. The dual function $q({\boldsymbol{\lambda}})=\Phi_{\boldsymbol{\lambda}}(Z({\boldsymbol{\lambda}}))$ is an infimum of
affine functions of ${\boldsymbol{\lambda}}$, hence concave; it is continuous by part (d) below, so it attains its maximum on $\Delt$ at
some ${\boldsymbol{\lambda}}^k$. Let $X^{k+1}$ be the primal minimizer. Then
\[
  \Phi(X^{k+1})=q({\boldsymbol{\lambda}}^k)=\Phi_{\boldsymbol{\lambda}^k}(Z({\boldsymbol{\lambda}}^k))\le\Phi_{\boldsymbol{\lambda}^k}(X^{k+1})\le\max_{{\boldsymbol{\lambda}}\in\Delt}\Phi_{\boldsymbol{\lambda}}(X^{k+1})=\Phi(X^{k+1}),
\]
so all inequalities are equalities. The first equality $\Phi_{\boldsymbol{\lambda}^k}(Z({\boldsymbol{\lambda}}^k))=\Phi_{\boldsymbol{\lambda}^k}(X^{k+1})$ and uniqueness of
the minimizer of $\Phi_{\boldsymbol{\lambda}^k}$ give $X^{k+1}=Z({\boldsymbol{\lambda}}^k)$.

\paragraph{(c) Complementary slackness.}
The second equality $\Phi_{\boldsymbol{\lambda}^k}(X^{k+1})=\Phi(X^{k+1})$ reads $\sum_i\lambda^k_i\psi_i(X^{k+1})=\max_i\psi_i(X^{k+1})$, i.e.\
$\lambda^k_i>0$ only for indices attaining the maximum.

\paragraph{(d) Danskin's formula.}
Let ${\boldsymbol{\lambda}},{\boldsymbol{\lambda}}'\in\Delt$ and write $\psi(Z)=(\psi_i(Z))_i$. Since $\boldsymbol{\lambda}\mapsto\Phi_{\boldsymbol{\lambda}}(Z)$ is affine with gradient $\psi(Z)$,
\[
  q({\boldsymbol{\lambda}}')\le\Phi_{\boldsymbol{\lambda}'}(Z({\boldsymbol{\lambda}}))=q({\boldsymbol{\lambda}})+\langle \psi(Z({\boldsymbol{\lambda}})),{\boldsymbol{\lambda}}'-{\boldsymbol{\lambda}}\rangle,\qquad
  q({\boldsymbol{\lambda}})\le q({\boldsymbol{\lambda}}')+\langle \psi(Z({\boldsymbol{\lambda}}')),{\boldsymbol{\lambda}}-{\boldsymbol{\lambda}}'\rangle .
\]
Hence
\begin{equation}\label{eq:danskin}
  0\le q({\boldsymbol{\lambda}})+\langle \psi(Z({\boldsymbol{\lambda}})),{\boldsymbol{\lambda}}'-{\boldsymbol{\lambda}}\rangle-q({\boldsymbol{\lambda}}')\le\langle \psi(Z({\boldsymbol{\lambda}}'))-\psi(Z({\boldsymbol{\lambda}})),{\boldsymbol{\lambda}}'-{\boldsymbol{\lambda}}\rangle.
\end{equation}
Now $\psi_i(Z)-\psi_i(Z')=\ip{g_i}{Z-Z'}+[\tH(Z)-\tH(Z')]$, and the bracket is the same for all $i$. Since
$\langle \mathbf 1,{\boldsymbol{\lambda}}'-{\boldsymbol{\lambda}}\rangle=0$ on the simplex, the common term drops out. With the linear map $\mathcal G:\R^K\to\Ltwo$,
$\mathcal G{\boldsymbol{\lambda}}=\sum_i\lambda_ig_i$, whose adjoint is $\mathcal G^*Z=(\ip{g_i}Z)_i$ and for which $\mathcal G^*\mathcal G=Q$,
\eqref{eq:nonexp} gives
\[
  \langle \psi(Z({\boldsymbol{\lambda}}'))-\psi(Z({\boldsymbol{\lambda}})),{\boldsymbol{\lambda}}'-{\boldsymbol{\lambda}}\rangle=\langle \mathcal G^*(Z({\boldsymbol{\lambda}}')-Z({\boldsymbol{\lambda}})),{\boldsymbol{\lambda}}'-{\boldsymbol{\lambda}}\rangle
  \le\|\mathcal G\|\,\|Z({\boldsymbol{\lambda}}')-Z({\boldsymbol{\lambda}})\|_\nu\,|{\boldsymbol{\lambda}}'-{\boldsymbol{\lambda}}|\le\frac{\lambda_{\max}(Q)}L|{\boldsymbol{\lambda}}'-{\boldsymbol{\lambda}}|^2 .
\]
Thus~\eqref{eq:danskin} shows that $q$ is differentiable along the affine hull of $\Delt$ with gradient $\psi(Z({\boldsymbol{\lambda}}))$
(defined modulo $\mathbf 1$), that $q$ is continuous, and that ${\boldsymbol{\lambda}}\mapsto\mathcal G^*Z({\boldsymbol{\lambda}})$, which equals $\nabla q$ up to a
multiple of $\mathbf 1$, is Lipschitz with constant $\lambda_{\max}(Q)/L$. Since the Euclidean projection onto $\Delt$ satisfies
$P_{\Delt}(v+c\mathbf 1)=P_{\Delt}(v)$, projected gradient ascent with step $L/\lambda_{\max}(Q)$ (and its accelerated version) can use
$\tilde\psi(Z({\boldsymbol{\lambda}}))=(\ip{g_i}{Z({\boldsymbol{\lambda}})-Y}+c_i)_i$ without the entropy term, and converges at the usual rates for smooth
concave maximization~\cite{beck2009fast}. \qed

\section{Proof of the one-step inequality (Lemma~\ref{lem:onestep})}
\label{app:onestep}

Write $Y=Y^k$, ${\boldsymbol{\lambda}}={\boldsymbol{\lambda}}^k$, $X^+=X^{k+1}$, and $\Psi=\sum_i\lambda_i\psi_i$. We prove the slightly more general statement
used later for inexact steps.

\begin{lemma}\label{lem:onestep-general}
Let ${\boldsymbol{\lambda}}\in\Delt$ and $X^+\in \mathcal{K}_\nu$ satisfy, for some $\varepsilon,\delta\ge0$,
\begin{align}
  \Psi(X^+)+\tfrac L2\|X^+-Y\|_\nu^2+\tfrac L2\|W-X^+\|_\nu^2&\le\Psi(W)+\tfrac L2\|W-Y\|_\nu^2+\varepsilon\qquad\forall W\in \mathcal{K}_\nu,\label{eq:three-point}\\
  \max_i\psi_i(X^+)&\le\Psi(X^+)+\delta .\label{eq:cs-gap}
\end{align}
Then for every $W\in \mathcal{K}_\nu$,
\[
  \max_i\big[\tG_i(X^+)-\tG_i(X^k)\big]\le\sum_i\lambda_i\big[\tG_i(W)-\tG_i(X^k)\big]+\tfrac{L-m}2\|W-Y\|_\nu^2-\tfrac L2\|W-X^+\|_\nu^2+\varepsilon+\delta .
\]
\end{lemma}

\begin{proof}
\emph{Step 1 (upper model).} By the upper bound of the lifted inequality in Appendix~\ref{proof:liftedproblem}(b) with $X=X^+$,
\[
  \tG_i(X^+)-\tG_i(X^k)=\tF_i(X^+)+\tH(X^+)-\tG_i(X^k)\le\underbrace{\tF_i(Y)+\ip{g_i}{X^+-Y}+\tH(X^+)-\tG_i(X^k)}_{=\psi_i(X^+)}
  +\tfrac L2\|X^+-Y\|_\nu^2 .
\]
Taking the maximum over $i$ and using~\eqref{eq:cs-gap},
\begin{equation}\label{eq:os1}
  \max_i\big[\tG_i(X^+)-\tG_i(X^k)\big]\le\Psi(X^+)+\tfrac L2\|X^+-Y\|_\nu^2+\delta .
\end{equation}
\emph{Step 2 (three-point inequality).} By~\eqref{eq:three-point},
\begin{equation}\label{eq:os2}
  \Psi(X^+)+\tfrac L2\|X^+-Y\|_\nu^2\le\Psi(W)+\tfrac L2\|W-Y\|_\nu^2-\tfrac L2\|W-X^+\|_\nu^2+\varepsilon .
\end{equation}
\emph{Step 3 (lower model).} By the lower bound of the lifted inequality in Appendix~\ref{proof:liftedproblem}(b) with $X=W$,
\[
  \psi_i(W)=\tF_i(Y)+\ip{g_i}{W-Y}+\tH(W)-\tG_i(X^k)\le\tG_i(W)-\tG_i(X^k)-\tfrac m2\|W-Y\|_\nu^2,
\]
and averaging with weights ${\boldsymbol{\lambda}}$,
\begin{equation}\label{eq:os3}
  \Psi(W)\le\sum_i\lambda_i\big[\tG_i(W)-\tG_i(X^k)\big]-\tfrac m2\|W-Y\|_\nu^2 .
\end{equation}
Chaining~\eqref{eq:os1}, \eqref{eq:os2} and~\eqref{eq:os3} gives the claim. (If $\tH(W)=+\infty$ the claim is trivial.)
\end{proof}

\begin{proof}[Proof of Lemma~\ref{lem:onestep}]
For the exact solution, ${\boldsymbol{\lambda}}={\boldsymbol{\lambda}}^k$ and $X^+=X^{k+1}=Z({\boldsymbol{\lambda}}^k)$. Condition~\eqref{eq:cs-gap} holds with $\delta=0$ by
complementary slackness (Lemma~\ref{lem:sub}(c)). For~\eqref{eq:three-point} with $\varepsilon=0$: the function
$f=\Phi_{\boldsymbol{\lambda}^k}+\iota_\mathcal{K}=h+\langle\mathcal G{\boldsymbol{\lambda}}^k,\cdot-Y\rangle_\nu+\mathrm{const}+\frac L2\|\cdot-Y\|_\nu^2$ is
$L$-strongly convex on $\Ltwo$ and minimized at $X^+$. For $W\in \mathcal{K}_\nu$ with $\tH(W)<\infty$ and $s\in(0,1)$, the point
$X^+ + s(W-X^+)$ lies in $\mathcal{K}_\nu$, and $L$-strong convexity gives
\[
  f(X^+)\le f\big(X^++s(W-X^+)\big)\le(1-s)f(X^+)+sf(W)-\tfrac L2s(1-s)\|W-X^+\|_\nu^2 .
\]
Subtracting $(1-s)f(X^+)$, dividing by $s$ and letting $s\downarrow0$ yields $f(X^+)+\frac L2\|W-X^+\|_\nu^2\le f(W)$, which
is~\eqref{eq:three-point} with $\varepsilon=0$. Lemma~\ref{lem:onestep-general} applies.
\end{proof}

%---------------------------------------------------------------------------------------------------------------------
\section{Proof of the level-set lemma (Lemma~\ref{lem:level})}
\label{app:level}

All iterates lie in $\mathcal{K}_\nu$ because they are outputs of the proximal step, and $\tH(X^k)<\infty$ because
$\Phi(X^{k+1})<\infty$. Apply Lemma~\ref{lem:onestep} with $W=X^k\in \mathcal{K}_\nu$ (and $m\ge0$ dropped):
\[
  \max_i\big[\tG_i(X^{k+1})-\tG_i(X^k)\big]\le\tfrac L2\|X^k-Y^k\|_\nu^2-\tfrac L2\|X^k-X^{k+1}\|_\nu^2=\tfrac L2\big(\gamma_k^2a_k-a_{k+1}\big),
\]
with $a_k=\|X^k-X^{k-1}\|_\nu^2$. All three methods have $\gamma_k\in[0,1)$ for $k\ge1$ (for MWAPG-C, $0\le t_k-1<t_{k+1}$; for
MWAPG-SC, $\beta<1$), and $a_0=0$ because $X^{-1}=X^0$. Hence, for each $i$ and $k\ge1$,
\[
  \tG_i(X^k)-\tG_i(X^0)=\sum_{j=0}^{k-1}\big[\tG_i(X^{j+1})-\tG_i(X^j)\big]\le\tfrac L2\sum_{j=0}^{k-1}\big(a_j-a_{j+1}\big)
  =\tfrac L2(a_0-a_k)=-\tfrac L2\|X^k-X^{k-1}\|_\nu^2 .
\]
Thus $G_i(\mu_k)\le G_i(\mu_0)$ for all $i$, i.e.\ $\mu_k\in\mathcal{S}_0$. \qed

\paragraph{From gaps to the merit function.}
The following lemma is used by all rate proofs. For a fixed $Z\in \mathcal{K}_\nu$ with $\tH(Z)<\infty$ let
\[
  \sigma_k(Z)=\min_i\big[\tG_i(X^k)-\tG_i(Z)\big].
\]

\begin{lemma}\label{lem:gap-to-merit}
Suppose $G_i(\mu_k)\le G_i(\mu_0)+S$ for all $i$, Assumption~\textbf{(A3$_S$)} holds, and
$\sigma_k(Z_{\rho^*})\le B(\|X^0-Z_{\rho^*}\|_\nu^2)$ for every $\rho^*\in\mathcal{P}^*\cap\mathcal{S}_S$, with $B$ nondecreasing. Then
$u_0(\mu_k)\le B(R_S)$. (For exact steps, $S=0$ and $R_S=R$.)
\end{lemma}

\begin{proof}
If no $\rho$ satisfies $G_i(\rho)<G_i(\mu_k)$ for all $i$, then $u_0(\mu_k)=0$. Otherwise, let $\rho$ be such a measure. Then
$G_i(\rho)<G_i(\mu_k)\le G_i(\mu_0)+S$, so $\rho\in\mathcal{S}_S$, and by~\textbf{(A3$_S$)} there is $\rho^*\in\mathcal{P}^*\cap\cP_2$ with
$G_i(\rho^*)\le G_i(\rho)$ for all $i$; in particular $\rho^*\in\mathcal{S}_S$. Hence
\[
  \min_i\big[G_i(\mu_k)-G_i(\rho)\big]\le\min_i\big[G_i(\mu_k)-G_i(\rho^*)\big]=\sigma_k(Z_{\rho^*})
  \le B(\|X^0-Z_{\rho^*}\|_\nu^2)=B(\mathcal{W}_2^2(\mu_0,\rho^*))\le B(R_S),
\]
using $\tG_i(Z_{\rho^*})=G_i(\rho^*)$ and Lemma~\ref{lemma:liftedproblem}(a). Taking the supremum over $\rho$ gives the claim.
\end{proof}

A key elementary fact is the recursion for $\sigma$. For every $k$ and every $Z$,
\begin{equation}\label{eq:sigma-rec}
  \sigma_{k+1}(Z)\le\sigma_k(Z)+\Delta_k,\qquad \Delta_k=\max_j\big[\tG_j(X^{k+1})-\tG_j(X^k)\big],
\end{equation}
because $\tG_i(X^{k+1})-\tG_i(Z)=[\tG_i(X^{k+1})-\tG_i(X^k)]+[\tG_i(X^k)-\tG_i(Z)]\le\Delta_k+[\tG_i(X^k)-\tG_i(Z)]$ for each $i$,
and the minimum over $i$ of the right-hand side is $\Delta_k+\sigma_k(Z)$.

%---------------------------------------------------------------------------------------------------------------------
\section{Proof of Theorem~\ref{thm:convex} (convex case)}
\label{app:convex}

\begin{lemma}[The sequence $t_k$]\label{lem:tk}
Let $t_0=0$ and $t_{k+1}=\frac12(1+\sqrt{1+4t_k^2})$. Then $t_{k+1}^2-t_{k+1}=t_k^2$, $t_1=1$, $t_{k+1}\ge1$ for $k\ge0$, and
$\frac{k+1}2\le t_k\le k$ for $k\ge1$. Moreover $\gamma_k=(t_k-1)/t_{k+1}\in[0,1)$ for $k\ge1$.
\end{lemma}

\begin{proof}
$t_{k+1}$ is the positive root of $t^2-t-t_k^2=0$. From $\sqrt{1+4t_k^2}\ge2t_k$ we get $t_{k+1}\ge t_k+\frac12$, and from
$\sqrt{1+4t_k^2}\le1+2t_k$ we get $t_{k+1}\le t_k+1$. With $t_1=1$, induction gives $1+\frac{k-1}2\le t_k\le k$. Finally
$t_k\ge1$ gives $\gamma_k\ge0$, and $t_k-1<t_k+\frac12\le t_{k+1}$ gives $\gamma_k<1$.
\end{proof}

\begin{proof}[Proof of Theorem~\ref{thm:convex}]
Fix $\rho^*\in\mathcal{P}^*\cap\mathcal{S}_0$, $Z=Z_{\rho^*}\in \mathcal{K}_\nu$, and write $\sigma_k=\sigma_k(Z)$, $t=t_{k+1}$, $Y=Y^k$. Define the
far points
\[
  U^k=X^{k-1}+t_k(X^k-X^{k-1}),\qquad k\ge0,
\]
so that $U^0=X^{-1}=X^0$ since $t_0=0$.

\emph{Step 1 (comparison point).} Let $W=(1-\frac1t)X^k+\frac1tZ$. Since $t\ge1$, $W$ is a convex combination of
$X^k,Z\in \mathcal{K}_\nu$, so $W\in \mathcal{K}_\nu$ by Lemma~\ref{lemma:liftedproblem}(a). By convexity of $\tF_i$ on $\Ltwo$ and of $\tH$ on $\mathcal{K}_\nu$,
$\tG_i(W)\le(1-\frac1t)\tG_i(X^k)+\frac1t\tG_i(Z)$, so
\[
  \sum_i\lambda^k_i\big[\tG_i(W)-\tG_i(X^k)\big]\le-\frac1t\sum_i\lambda^k_i\big[\tG_i(X^k)-\tG_i(Z)\big]\le-\frac1t\sigma_k,
\]
where the last step uses that each bracket is at least $\sigma_k$ and $\sum_i\lambda^k_i=1$. This holds whatever the sign of
$\sigma_k$.

\emph{Step 2 (one step).} Lemma~\ref{lem:onestep} (with $m=0$) and~\eqref{eq:sigma-rec} give
\begin{equation}\label{eq:conv-step}
  \sigma_{k+1}\le\Big(1-\frac1t\Big)\sigma_k+\frac L2\|W-Y\|_\nu^2-\frac L2\|W-X^{k+1}\|_\nu^2 .
\end{equation}

\emph{Step 3 (identities).} Using $Y=X^k+\gamma_k(X^k-X^{k-1})$ and $t\gamma_k=t_k-1$,
\[
\begin{aligned}
  t(W-Y)&=(t-1)X^k+Z-tX^k-(t_k-1)(X^k-X^{k-1})=Z-\big[X^{k-1}+t_k(X^k-X^{k-1})\big]=Z-U^k,\\
  t(W-X^{k+1})&=(t-1)X^k+Z-tX^{k+1}=Z-\big[X^k+t_{k+1}(X^{k+1}-X^k)\big]=Z-U^{k+1}.
\end{aligned}
\]
(For $k=0$ the first identity holds for any $\gamma_0$ because $X^{-1}=X^0$.)

\emph{Step 4 (energy).} Multiply~\eqref{eq:conv-step} by $t^2=t_{k+1}^2>0$ and use $t_{k+1}^2-t_{k+1}=t_k^2$ (with equality;
an inequality $\le$ would not suffice because $\sigma_k$ may be negative):
\[
  t_{k+1}^2\sigma_{k+1}\le t_k^2\sigma_k+\frac L2\|U^k-Z\|_\nu^2-\frac L2\|U^{k+1}-Z\|_\nu^2 .
\]
Hence the energy $E_k=t_k^2\sigma_k+\frac L2\|U^k-Z\|_\nu^2$ satisfies $E_{k+1}\le E_k$, and $E_0=\frac L2\|X^0-Z\|_\nu^2$.

\emph{Step 5 (rate).} For $k\ge1$, $t_k^2\sigma_k\le E_k\le E_0$, so by Lemma~\ref{lem:tk}
\[
  \sigma_k(Z_{\rho^*})\le\frac{L\|X^0-Z_{\rho^*}\|_\nu^2}{2t_k^2}\le\frac{2L\|X^0-Z_{\rho^*}\|_\nu^2}{(k+1)^2}.
\]
By Lemma~\ref{lem:level}, $\mu_k\in\mathcal{S}_0$, so Lemma~\ref{lem:gap-to-merit} with $S=0$ and $B(r)=Lr/(2t_k^2)$ gives
$u_0(\mu_k)\le LR/(2t_k^2)\le2LR/(k+1)^2$.
\end{proof}

\begin{remark}[Two sequences suffice]\label{rem:twoseq}
Single-objective accelerated methods over measures whose smooth part is convex only along generalized geodesics use more
elaborate schemes. Under~\textbf{(A1)} the potentials are convex on all of $\Ltwo$, so the two-sequence form of
Algorithm~\ref{alg:mwapg} suffices. The entropy inequality is used only between points of $\mathcal{K}_\nu$, namely $X^{k+1}$ and the
comparison point $W$.
\end{remark}

\begin{remark}[Bounded far points]
Step 4 also gives $\frac L2\|U^k-Z\|_\nu^2\le E_0-t_k^2\sigma_k$. If $\sigma_k(Z)\ge0$ (e.g.\ for $\rho^*$ dominating
$\mu_k$), then $\|U^k-Z\|_\nu\le\|X^0-Z\|_\nu$, the analogue of the bounded far point in FISTA.
\end{remark}

%---------------------------------------------------------------------------------------------------------------------
\section{Proof of Theorem~\ref{thm:sc} (strongly convex case)}
\label{app:sc}

Let $\theta=\sqrt{m/L}\in(0,1]$, so that $L\theta^2=m$, and $\beta=\frac{1-\theta}{1+\theta}$. Fix
$\rho^*\in\mathcal{P}^*\cap\mathcal{S}_0$, $Z=Z_{\rho^*}$, $\sigma_k=\sigma_k(Z)$.

\begin{proof}[Proof of Theorem~\ref{thm:sc}]
Define the far points
\[
  U^k=X^{k-1}+\tfrac1\theta(X^k-X^{k-1}),\qquad U^0=X^0 .
\]

\emph{Step 1 (comparison point).} Let $W=(1-\theta)X^k+\theta Z\in \mathcal{K}_\nu$. Now $\tF_i$ is $m$-strongly convex on $\Ltwo$ and
$\tH$ is convex on $\mathcal{K}_\nu$, so
\[
  \tG_i(W)\le(1-\theta)\tG_i(X^k)+\theta\tG_i(Z)-\tfrac m2\theta(1-\theta)\|X^k-Z\|_\nu^2,
\]
and, as in the convex case,
\[
  \sum_i\lambda^k_i\big[\tG_i(W)-\tG_i(X^k)\big]\le-\theta\sigma_k-\tfrac m2\theta(1-\theta)\|X^k-Z\|_\nu^2 .
\]

\emph{Step 2 (one step).} Lemma~\ref{lem:onestep} with strong convexity constant $m$ and~\eqref{eq:sigma-rec} give
\begin{equation}\label{eq:sc-step}
  \sigma_{k+1}\le(1-\theta)\sigma_k-\tfrac m2\theta(1-\theta)\|X^k-Z\|_\nu^2+\tfrac{L-m}2\|W-Y^k\|_\nu^2-\tfrac L2\|W-X^{k+1}\|_\nu^2 .
\end{equation}

\emph{Step 3 (identities).} First,
\[
  W-X^{k+1}=\theta Z-\big[X^{k+1}-(1-\theta)X^k\big]=\theta\Big(Z-\big[X^k+\tfrac1\theta(X^{k+1}-X^k)\big]\Big)=\theta(Z-U^{k+1}),
\]
so $\frac L2\|W-X^{k+1}\|_\nu^2=\frac{L\theta^2}2\|Z-U^{k+1}\|_\nu^2=\frac m2\|Z-U^{k+1}\|_\nu^2$. Second, let
$\hat Y=X^k+\frac\beta\theta(X^k-X^{k-1})$, so that $Y^k=X^k+\beta(X^k-X^{k-1})=(1-\theta)X^k+\theta\hat Y$ and
$W-Y^k=\theta(Z-\hat Y)$. Since $U^k=X^k+\frac{1-\theta}\theta(X^k-X^{k-1})$ and $\frac\beta\theta=\frac{1-\theta}{\theta(1+\theta)}$,
\[
  \hat Y=\frac{\theta X^k+U^k}{1+\theta},\qquad\text{so}\qquad Z-\hat Y=\frac{\theta}{1+\theta}(Z-X^k)+\frac1{1+\theta}(Z-U^k).
\]
By convexity of $\|\cdot\|_\nu^2$,
\[
  \|Z-\hat Y\|_\nu^2\le\frac\theta{1+\theta}\|Z-X^k\|_\nu^2+\frac1{1+\theta}\|Z-U^k\|_\nu^2 .
\]

\emph{Step 4 (cancellation).} Using $(L-m)\theta^2=m(1-\theta^2)=m(1-\theta)(1+\theta)$,
\[
  \tfrac{L-m}2\|W-Y^k\|_\nu^2=\tfrac{(L-m)\theta^2}2\|Z-\hat Y\|_\nu^2\le\tfrac m2(1-\theta)\big[\theta\|Z-X^k\|_\nu^2+\|Z-U^k\|_\nu^2\big].
\]
The term $\frac m2(1-\theta)\theta\|Z-X^k\|_\nu^2$ cancels exactly with the strong-convexity term in~\eqref{eq:sc-step}. Hence
\[
  \sigma_{k+1}+\tfrac m2\|U^{k+1}-Z\|_\nu^2\le(1-\theta)\Big[\sigma_k+\tfrac m2\|U^k-Z\|_\nu^2\Big],
\]
i.e.\ $E_{k+1}\le(1-\theta)E_k$ for $E_k=\sigma_k+\frac m2\|U^k-Z\|_\nu^2$, regardless of the sign of $E_k$.

\emph{Step 5 (rate).} By induction $\sigma_k\le E_k\le(1-\theta)^kE_0$ with
$E_0=\sigma_0(Z)+\frac m2\|X^0-Z\|_\nu^2\le u_0(\mu_0)+\frac m2\|X^0-Z\|_\nu^2$, because
$\sigma_0(Z_{\rho^*})=\min_i[G_i(\mu_0)-G_i(\rho^*)]\le u_0(\mu_0)$ by definition~\eqref{eqn:defofmeritfunction}. Lemma~\ref{lem:level}
gives $\mu_k\in\mathcal{S}_0$, and Lemma~\ref{lem:gap-to-merit} with $B(r)=(1-\theta)^k[u_0(\mu_0)+\frac m2r]$ gives the theorem.
\end{proof}

\begin{proof}[Proof of Corollary~\ref{cor:sc-explicit}]
By Lemma~\ref{lem:A3sc}, $\frac m2R\le\max_i\mathrm{KL}(\mu_0\|\pi_i)$, and by Proposition~\ref{proposition:meritforklobjective},
$u_0(\mu_0)\le\min_i\mathrm{KL}(\mu_0\|\pi_i)$ (take ${\boldsymbol{\lambda}}$ a vertex). Hence $u_0(\mu_k)\le2(1-\theta)^k\max_i\mathrm{KL}(\mu_0\|\pi_i)$, and
$(1-\theta)^k\le e^{-\theta k}$ gives the iteration count.
\end{proof}

%---------------------------------------------------------------------------------------------------------------------
\section{Proof of Proposition~\ref{prop:mwpg} (MWPG)}
\label{app:mwpg}

For MWPG, $Y^k=X^k$. Lemma~\ref{lem:onestep} with $W=X^k$ gives
$\max_i[\tG_i(X^{k+1})-\tG_i(X^k)]\le-\frac L2\|X^{k+1}-X^k\|_\nu^2$, which is the monotone decrease. In particular $\mu_k\in\mathcal{S}_0$ and,
for any fixed $Z$, $\sigma_k(Z)$ is nonincreasing in $k$.

Fix $\rho^*\in\mathcal{P}^*\cap\mathcal{S}_0$, $Z=Z_{\rho^*}$ and $D_k=\|X^k-Z\|_\nu^2$. Lemma~\ref{lem:onestep} with $W=Z$ (allowed since
$Z\in \mathcal{K}_\nu$) and the scalarization bound $\sum_i\lambda^k_i[\tG_i(Z)-\tG_i(X^k)]\le-\sigma_k$ give, with~\eqref{eq:sigma-rec},
\begin{equation}\label{eq:mwpg-step}
  \sigma_{k+1}\le\sigma_k+\Delta_k\le\tfrac{L-m}2D_k-\tfrac L2D_{k+1}=\tfrac L2\big(qD_k-D_{k+1}\big),\qquad q=1-\tfrac mL .
\end{equation}
\emph{Convex case ($m=0$, $q=1$).} Summing~\eqref{eq:mwpg-step} over $j=0,\dots,k-1$ gives
$\sum_{j=1}^k\sigma_j\le\frac L2D_0$, and monotonicity gives $k\sigma_k\le\sum_{j=1}^k\sigma_j$. Hence
$\sigma_k\le\frac L{2k}\|X^0-Z\|_\nu^2$.

\emph{Strongly convex case.} Multiply~\eqref{eq:mwpg-step} at step $j$ by $q^{-(j+1)}$:
$q^{-(j+1)}\sigma_{j+1}\le\frac L2(q^{-j}D_j-q^{-(j+1)}D_{j+1})$. Summing over $j=0,\dots,k-1$ and using $\sigma_j\ge\sigma_k$ for
$j\le k$,
\[
  \sigma_k\sum_{j=1}^kq^{-j}\le\sum_{j=1}^kq^{-j}\sigma_j\le\tfrac L2D_0,\qquad\text{and}\qquad\sum_{j=1}^kq^{-j}=\frac{q^{-k}-1}{1-q},
\]
so $\sigma_k\le\frac L2(1-q)D_0\frac{q^k}{1-q^k}=\frac m2D_0\frac{q^k}{1-q^k}$. Note that this tends to $\frac L{2k}D_0$ as
$m\to0$. In both cases Lemma~\ref{lem:gap-to-merit} concludes. Since $q^k/(1-q^k)\le\epsilon'$ once
$k\ge\kap\log(1+1/\epsilon')$, MWPG needs $O(\kap\log(1/\epsilon))$ iterations. \qed

%---------------------------------------------------------------------------------------------------------------------
\section{Proof of Theorem~\ref{thm:mono} (monotone variants)}
\label{app:mono}

Let $Z^0=X^{-1}=X^0$. In both variants $Z^{k+1}$ is the exact solution of~\eqref{eqn:proximalstep} with extrapolated point $Y^k$ and anchor
$X^k$, so Lemma~\ref{lem:onestep} applies with $X^+=Z^{k+1}$: for every $W\in \mathcal{K}_\nu$,
\begin{equation}\label{eq:mono-os}
  \Delta_k:=\max_j\big[\tG_j(Z^{k+1})-\tG_j(X^k)\big]\le\sum_i\lambda^k_i\big[\tG_i(W)-\tG_i(X^k)\big]+\tfrac{L-m}2\|W-Y^k\|_\nu^2-\tfrac L2\|W-Z^{k+1}\|_\nu^2 .
\end{equation}
\emph{Monotonicity.} By construction $X^{k+1}\in\{Z^{k+1},X^k\}$ and $X^{k+1}=Z^{k+1}$ only if $\tG_i(Z^{k+1})\le\tG_i(X^k)$ for all
$i$; hence $\tG_i(X^{k+1})\le\tG_i(X^k)$ for all $i$, and $\mu_k\in\mathcal{S}_0$ for all $k$.

\emph{Gap recursion.} Fix $Z=Z_{\rho^*}$ and $\sigma_k=\min_i[\tG_i(X^k)-\tG_i(Z)]$. If $Z^{k+1}$ is accepted, $X^{k+1}=Z^{k+1}$
and~\eqref{eq:sigma-rec} gives $\sigma_{k+1}\le\sigma_k+\Delta_k$. If it is rejected, then $\Delta_k>0$ and $X^{k+1}=X^k$, so
$\sigma_{k+1}=\sigma_k<\sigma_k+\Delta_k$. In both cases
\begin{equation}\label{eq:mono-rec}
  \sigma_{k+1}\le\sigma_k+\Delta_k .
\end{equation}

\emph{Convex case.} Let $t=t_{k+1}$, $W=(1-\frac1t)X^k+\frac1tZ\in \mathcal{K}_\nu$, and define the far points
$U^k=X^{k-1}+t_k(Z^k-X^{k-1})$, so $U^0=X^0$. As in Appendix~\ref{app:convex}, $\sum_i\lambda^k_i[\tG_i(W)-\tG_i(X^k)]\le-\sigma_k/t$.
The extrapolation of the monotone variant is
$Y^k=X^k+\frac{t_k}{t}(Z^k-X^k)+\frac{t_k-1}{t}(X^k-X^{k-1})=X^k+\frac1t(U^k-X^k)$ (expand $U^k-X^k=t_k(Z^k-X^k)+(t_k-1)(X^k-X^{k-1})$),
and therefore
\[
  t(W-Y^k)=(t-1)X^k+Z-tX^k-(U^k-X^k)=Z-U^k,\qquad t(W-Z^{k+1})=Z-\big[X^k+t(Z^{k+1}-X^k)\big]=Z-U^{k+1}.
\]
Combining~\eqref{eq:mono-os} (with $m=0$) and~\eqref{eq:mono-rec}, multiplying by $t^2$ and using $t_{k+1}^2-t_{k+1}=t_k^2$, we obtain
$E_{k+1}\le E_k$ for $E_k=t_k^2\sigma_k+\frac L2\|U^k-Z\|_\nu^2$, with $E_0=\frac L2\|X^0-Z\|_\nu^2$. The rest of the proof of
Theorem~\ref{thm:convex} applies unchanged. (For $k=0$, $t_0=0$ gives $Y^0=X^0$ whatever the momentum terms.)

\emph{Strongly convex case.} Let $W=(1-\theta)X^k+\theta Z$ and $U^k=X^{k-1}+\theta^{-1}(Z^k-X^{k-1})$, so $U^0=X^0$. Then
$W-Z^{k+1}=\theta\big(Z-[X^k+\theta^{-1}(Z^{k+1}-X^k)]\big)=\theta(Z-U^{k+1})$. The extrapolation
$Y^k=X^k+\frac\theta{1+\theta}(U^k-X^k)$ equals $(1-\theta)X^k+\theta\hat Y$ with $\hat Y=(\theta X^k+U^k)/(1+\theta)$, because
$(1-\theta)X^k+\frac{\theta(\theta X^k+U^k)}{1+\theta}=\frac{X^k+\theta U^k}{1+\theta}$; hence $W-Y^k=\theta(Z-\hat Y)$ with exactly the
same $\hat Y$ as in Appendix~\ref{app:sc}. Steps 1, 4 and 5 of that proof apply verbatim with~\eqref{eq:mono-rec} in place
of~\eqref{eq:sigma-rec}, and give $E_{k+1}\le(1-\theta)E_k$ for $E_k=\sigma_k+\frac m2\|U^k-Z\|_\nu^2$. \qed

\paragraph{Restart and tolerance.}
With the optional restart, a rejected step resets the momentum, so the next step is a plain proximal step from $X^{k+1}$; by
Proposition~\ref{prop:mwpg} it decreases every objective and is accepted (up to the subproblem accuracy). The monotonicity of
Theorem~\ref{thm:mono} is preserved; we do not claim the accelerated rates for the restarted variant, which is a heuristic in
the spirit of~\cite{odonoghue2015adaptive} and is faster in our experiments. In floating point and with inexact subproblems we
accept increases below a tolerance at the level of the subproblem accuracy ($10^{-10}$ in Experiment~1, $10^{-9}$ in
Experiment~3); by Theorem~\ref{thm:inexact} this only adds the accepted increases to the error terms.

%---------------------------------------------------------------------------------------------------------------------
\section{Proof of Theorem~\ref{thm:inexact} (inexact steps)}
\label{app:inexact}

Let $(\hat{\boldsymbol{\lambda}}^k,X^{k+1})$ be $(\varepsilon_k,\delta_k)$-solutions in the sense of~\eqref{eq:inexact}. Conditions~\eqref{eq:three-point}
and~\eqref{eq:cs-gap} hold with $(\varepsilon,\delta)=(\varepsilon_k,\delta_k)$, so Lemma~\ref{lem:onestep-general} gives the one-step
inequality with the additional term $\eta_k=\varepsilon_k+\delta_k$, with weights $\hat{\boldsymbol{\lambda}}^k$ in place of ${\boldsymbol{\lambda}}^k$. The weights
enter the proofs only through $\sum_i\hat\lambda^k_i[\tG_i(W)-\tG_i(X^k)]\le-\tau\sigma_k$, which holds for every $\hat{\boldsymbol{\lambda}}^k\in\Delt$.

\emph{Level set.} The argument of Appendix~\ref{app:level} gives
$\tG_i(X^k)-\tG_i(X^0)\le-\frac L2a_k+\sum_{j<k}\eta_j\le S$, so $\mu_k\in\mathcal{S}_S$.

\emph{Convex case.} Steps 1--4 of Appendix~\ref{app:convex} give $E_{k+1}\le E_k+t_{k+1}^2\eta_k$, hence
$t_k^2\sigma_k\le\frac L2\|X^0-Z\|_\nu^2+\sum_{j<k}t_{j+1}^2\eta_j$. With $t_k\ge\frac{k+1}2$ and Lemma~\ref{lem:gap-to-merit} (enlarged
level set $\mathcal{S}_S$),
\[
  u_0(\mu_k)\le\frac{2LR_S+4\sum_{j<k}t_{j+1}^2\eta_j}{(k+1)^2}.
\]
Since $t_{j+1}\le j+1$, the numerator is bounded if $\sum_j(j+1)^2\eta_j<\infty$, e.g.\ $\eta_j=O(j^{-3-\epsilon})$; with
$\eta_j=O(j^{-3})$ the rate is $O(\log k/k^2)$, and with $\eta_j=O(j^{-2})$ it is $O(1/k)$.

\emph{Strongly convex case.} Steps 1--4 of Appendix~\ref{app:sc} give $E_{k+1}\le(1-\theta)E_k+\eta_k$, hence
$\sigma_k\le(1-\theta)^kE_0+\sum_{j<k}(1-\theta)^{k-1-j}\eta_j$, and Lemma~\ref{lem:gap-to-merit} concludes. In particular, with
$\eta_j\le\bar\eta$ the error floor is $\bar\eta/\theta=\sqrt\kap\,\bar\eta$, and with $\eta_j=O(\varrho^j)$, $\varrho<1-\theta$, the
linear rate $(1-\theta)^k$ is preserved up to a constant. \qed

\begin{remark}[What the implementation controls]
With the exact proximal map at the current weights, $\varepsilon_k=0$ and $\delta_k=\max_i\psi_i(Z(\hat{\boldsymbol{\lambda}}))-\sum_i\hat\lambda_i\psi_i(Z(\hat{\boldsymbol{\lambda}}))$
is the primal--dual gap $\Phi(Z(\hat{\boldsymbol{\lambda}}))-q(\hat{\boldsymbol{\lambda}})$. Both Gaussian and particle implementations stop the dual solver when
$\delta_k$ is below a tolerance ($10^{-11}$, resp.\ $10^{-7}$). Note that $\delta_k$ can be computed without the entropy, since
$\tH(Z)$ is common to all $\psi_i$. In the particle implementation, $\varepsilon_k$ additionally contains the error of the network
fit and the restriction to the ICNN class, which we do not certify; Appendix~\ref{app:icnnval} validates it against the exact
Gaussian method.
\end{remark}

%---------------------------------------------------------------------------------------------------------------------
\section{Gaussian closed form: proof of Proposition~\ref{prop:gauss}}
\label{app:gauss}

Let $\nu=\mathcal N(0,I)$. For an affine map $X(\xi)=A\xi+b$ we have $\ip{X}{X'}=\E[\langle A\xi+b,A'\xi+b'\rangle]=\mathrm{tr}(A^\top A')+b^\top b'$,
and $\mu_X=\mathcal N(b,AA^\top)$, so $\tH(X)=-\frac d2\log(2\pi e)-\log|\det A|=H(\nu)-\log|\det A|$. For quadratic
$V_i(x)=\frac12(x-m_i)^\top P_i(x-m_i)$, $\tF_i(X)=\frac12\mathrm{tr}(A^\top P_iA)+\frac12(b-m_i)^\top P_i(b-m_i)$ and
$\nabla\tF_i(X)(\xi)=P_iA\xi+P_i(b-m_i)$, which is affine but in general not symmetric in its linear part.

\begin{proof}[Proof of Proposition~\ref{prop:gauss}]
Let $f(W)=\tH(W)+\frac L2\|W-T\|_\nu^2$ on $\mathcal{K}_\nu$, $S=\frac12(T_A+T_A^\top)=U\mathrm{diag}(s)U^\top$, $N=\frac12(T_A-T_A^\top)$, and
$A^\star=U\mathrm{diag}(a)U^\top$ with $a_j=\frac12(s_j+\sqrt{s_j^2+4/L})>0$, so that
\begin{equation}\label{eq:gauss-opt}
  A^\star-\tfrac1L(A^\star)^{-1}=S,
\end{equation}
since $a_j-\frac1{La_j}=s_j$ is equivalent to $a_j^2-s_ja_j-\frac1L=0$. Let $Z^\star(\xi)=A^\star\xi+T_b$, which lies in $\mathcal{K}_\nu$
(gradient of the convex quadratic $\frac12\xi^\top A^\star\xi+T_b^\top\xi$). Let $W\in \mathcal{K}_\nu$ with $\tH(W)<\infty$; write
$W=\nabla\varphi$ with $\varphi$ convex. Since $\nabla\varphi\in\Ltwo$ is defined $\nu$-a.e.\ and the interior of $\dom\varphi$ is a
convex set of full Gaussian measure, $\varphi$ is finite on $\R^d$.

\emph{Step 1 (entropy above its tangent).} Let $D^2_A\varphi$ be the Alexandrov Hessian of $\varphi$. By the change of variables
formula for convex gradients~\cite{mccann1995existence}, $\tH(W)=H(\nu)-\E_\nu\log\det D^2_A\varphi$. Concavity of
$\log\det$ on positive definite matrices gives $\log\det B\le\log\det A^\star+\mathrm{tr}((A^\star)^{-1}(B-A^\star))$, so
\[
  \tH(W)\ge\tH(Z^\star)-\E_\nu\,\mathrm{tr}\big((A^\star)^{-1}D^2_A\varphi\big)+d .
\]
The distributional Hessian $D^2\varphi$ is a matrix-valued measure equal to $D^2_A\varphi\,dx$ plus a positive semidefinite singular
part. Gaussian integration by parts (justified by truncation, using $\nabla\varphi\in\Ltwo$) gives, for any symmetric $M\succeq0$,
\[
  \E_\nu\langle M\xi,\nabla\varphi(\xi)\rangle=\int\mathrm{tr}(M\,dD^2\varphi)\,\gamma\ \ge\ \E_\nu\,\mathrm{tr}\big(MD^2_A\varphi\big),
\]
where $\gamma$ is the Gaussian density. With $M=(A^\star)^{-1}$ and $\E\langle (A^\star)^{-1}\xi,Z^\star(\xi)\rangle=d$, this yields
\begin{equation}\label{eq:gauss-subgrad}
  \tH(W)\ge\tH(Z^\star)+\ip{s^\star}{W-Z^\star},\qquad s^\star(\xi)=-(A^\star)^{-1}\xi .
\end{equation}
($s^\star=\nabla\log\mu_{Z^\star}\circ Z^\star$ is the Wasserstein gradient of $H$ at $\mu_{Z^\star}$, lifted.)

\emph{Step 2 (optimality).} Expanding the square,
$\frac L2\|W-T\|_\nu^2=\frac L2\|Z^\star-T\|_\nu^2+L\ip{Z^\star-T}{W-Z^\star}+\frac L2\|W-Z^\star\|_\nu^2$. Together with~\eqref{eq:gauss-subgrad},
\[
  f(W)\ge f(Z^\star)+L\ip{r}{W-Z^\star}+\tfrac L2\|W-Z^\star\|_\nu^2,\qquad r(\xi)=\tfrac1Ls^\star(\xi)+Z^\star(\xi)-T(\xi).
\]
By~\eqref{eq:gauss-opt}, $r(\xi)=(A^\star-\frac1L(A^\star)^{-1}-T_A)\xi=(S-T_A)\xi=-N\xi$ with $N$ skew-symmetric. By the same
integration by parts, $\E\langle N\xi,\nabla\varphi(\xi)\rangle=\int\mathrm{tr}(N\,dD^2\varphi)\gamma=0$, because $D^2\varphi$ is symmetric and
$N$ is skew; similarly $\E\langle N\xi,A^\star\xi+T_b\rangle=\mathrm{tr}(N^\top A^\star)=0$. Hence $\ip r{W-Z^\star}=0$ and
\[
  f(W)\ge f(Z^\star)+\tfrac L2\|W-Z^\star\|_\nu^2\qquad\forall W\in \mathcal{K}_\nu,
\]
so $Z^\star$ is the unique minimizer of $f$ over $\mathcal{K}_\nu$.
\end{proof}

\paragraph{Consequences.}
The Gaussian iterates are affine with symmetric positive definite linear part: $Y^k$ is a linear combination of such maps
(symmetric, possibly indefinite), $T_{\boldsymbol{\lambda}}=Y^k-\frac1L\sum_i\lambda_ig_i$ is affine with linear part
$Y^k_A-\frac1L\sum_i\lambda_iP_iY^k_A$, and the proximal map returns a symmetric positive definite matrix. Since the proposition
identifies the minimizer over all of $\mathcal{K}_\nu$, the Gaussian algorithm is Algorithm~\ref{alg:mwapg} exactly, and
Theorems~\ref{thm:convex}--\ref{thm:sc} apply with no approximation (up to the dual tolerance, covered by
Theorem~\ref{thm:inexact}). In the dual, $\psi_i(Z)-\tH(Z)=\mathrm{tr}((P_iY_A)^\top(Z_A-Y_A))+(P_i(Y_b-m_i))^\top(Z_b-Y_b)+c_i$ and
$Q_{ij}=\mathrm{tr}(Y_A^\top P_iP_jY_A)+(Y_b-m_i)^\top P_iP_j(Y_b-m_j)$.

\begin{algorithm}[h]
\caption{Gaussian proximal step (dual accelerated projected ascent)}\label{alg:gauss}
\begin{algorithmic}[1]
\REQUIRE $X^k=(A_k,b_k)$, $Y=(Y_A,Y_b)$, $L$, warm start ${\boldsymbol{\lambda}}^{\rm prev}$, tolerance $\tau=10^{-11}$
\STATE $g_i\gets(P_iY_A,\ P_i(Y_b-m_i))$, $c_i\gets\tF_i(Y)-\tG_i(X^k)$, $Q_{ij}\gets\ip{g_i}{g_j}$, $s\gets L/\lambda_{\max}(Q)$
\STATE $Z({\boldsymbol{\lambda}})$: symmetrize $T_A=Y_A-\frac1L\sum_i\lambda_iP_iY_A$, eigendecompose, apply $a=\frac12(s+\sqrt{s^2+4/L})$; $Z_b=T_b$
\STATE ${\boldsymbol{\lambda}}\gets{\boldsymbol{\lambda}}^{\rm prev}$, ${\boldsymbol{\lambda}}^-\gets{\boldsymbol{\lambda}}$, $\tau_1\gets1$
\REPEAT
\STATE $\tau_2\gets\frac12(1+\sqrt{1+4\tau_1^2})$; $v\gets{\boldsymbol{\lambda}}+\frac{\tau_1-1}{\tau_2}({\boldsymbol{\lambda}}-{\boldsymbol{\lambda}}^-)$
\STATE ${\boldsymbol{\lambda}}^-\gets{\boldsymbol{\lambda}}$; ${\boldsymbol{\lambda}}\gets P_{\Delt}\big(v+s\,\psi(Z(v))\big)$; $\tau_1\gets\tau_2$
\UNTIL{$\delta=\max_i\psi_i(Z({\boldsymbol{\lambda}}))-\langle \boldsymbol{\lambda},\psi(Z({\boldsymbol{\lambda}}))\rangle<\tau$}
\STATE \textbf{return} $X^{k+1}=Z({\boldsymbol{\lambda}})$, ${\boldsymbol{\lambda}}^k={\boldsymbol{\lambda}}$, $\delta_k=\delta$
\end{algorithmic}
\end{algorithm}

%---------------------------------------------------------------------------------------------------------------------
\section{Particle implementation with ICNN proximal steps}
\label{app:icnn}

\paragraph{Representation.}
We fix base samples $\xi_1,\dots,\xi_n\sim\nu=\mathcal N(0,I)$ once. The iterate $X^k$ is the particle array
$(X^k(\xi_j))_j$; it is the gradient of an ICNN~\eqref{eq:icnn}, so $X^k\in \mathcal{K}_\nu$ exactly. Extrapolation acts on arrays:
$Y^k(\xi_j)=X^k(\xi_j)+\gamma_k(X^k(\xi_j)-X^{k-1}(\xi_j))$. All $\mathcal{L}^2(\nu)$ inner products and the potential energies are
replaced by empirical means over the $\xi_j$, e.g.\ $\tF_i(Y)\approx\frac1n\sum_jV_i(Y(\xi_j))$.

\paragraph{Network.}
$\varphi_\vartheta(x)=\frac12x^\top Ax+b^\top x+\sum_{h=1}^{n_h}a_h\,\mathrm{softplus}(v_h^\top x+c_h)$ with $A=\Lambda\Lambda^\top+\epsilon_0I$,
$\epsilon=10^{-3}$, and $a_h=\mathrm{softplus}(\alpha_h)$. Then
\[
  \nabla\varphi_\vartheta(x)=Ax+b+\sum_ha_h\sigma(v_h^\top x+c_h)v_h,\qquad
  \nabla^2\varphi_\vartheta(x)=A+\sum_ha_h\sigma'(v_h^\top x+c_h)v_hv_h^\top\succeq\epsilon_0I,
\]
with $\sigma$ the logistic function. The network is initialized with $\Lambda$ such that $A=A_0$, $b=b_0$ (so that
$\nabla\varphi_\vartheta=X^0$ up to $\alpha_h=-6$, i.e.\ $a_h\approx2.5\cdot10^{-3}$), $v_h\sim\mathcal N(0,I)$ and $c_h\sim\mathcal N(0,1)$.
Because $\nabla^2\varphi_\vartheta\succ0$ everywhere, $\nabla\varphi_\vartheta$ is injective, $\mu=(\nabla\varphi_\vartheta)\push\nu$
has the density $\nu(x)/\det\nabla^2\varphi_\vartheta(x)$ at $\nabla\varphi_\vartheta(x)$, and
$\tH(\nabla\varphi_\vartheta)=H(\nu)-\E_\nu\log\det\nabla^2\varphi_\vartheta$ exactly.

\paragraph{Proximal fit.}
For a target array $T=(T(\xi_j))_j$ the fit minimizes
\[
  \ell(\vartheta)=-\frac1n\sum_{j=1}^n\log\det\nabla^2\varphi_\vartheta(\xi'_j)+\frac L{2n}\sum_{j=1}^n\big\|\nabla\varphi_\vartheta(\xi_j)-T(\xi_j)\big\|_2^2
  +\varrho\sum_ha_h,
\]
where $\xi'_j\sim\nu$ are fresh at every outer iteration (so the entropy is not overfitted to the particles), and the small
penalty $\varrho\sum_ha_h$ discourages kinks. Gradients of $\ell$ with respect to $(\Lambda,b,w,c,\alpha)$ are computed in closed
form (checked by finite differences, relative error $<10^{-6}$). We use L-BFGS with history 30 and at most 40 iterations per
fit, warm-started from the current network.

\paragraph{Dual Newton method.}
Each evaluation of $Z({\boldsymbol{\lambda}})$ is one warm-started fit at $T_{\boldsymbol{\lambda}}=Y-\frac1L\sum_i\lambda_ig_i$, which costs much more than in the
Gaussian case, so we use few dual iterations. With $\tilde\psi(Z)=(\ip{g_i}{Z-Y}+c_i)_i$ (the entropy is dropped, see
Lemma~\ref{lem:sub}(d)), we estimate the Jacobian $J_{:,j}\approx[\tilde\psi(Z({\boldsymbol{\lambda}}+he_j))-\tilde\psi(Z({\boldsymbol{\lambda}}))]/h$ with $h=0.05$,
symmetrize it and clip its eigenvalues to be negative (the true Hessian of $q$ is negative semidefinite modulo $\mathbf 1$), and
take the Newton step ${\boldsymbol{\lambda}}^+=\arg\max_{{\boldsymbol{\lambda}}'\in\Delt}\langle \tilde\psi,{\boldsymbol{\lambda}}'-{\boldsymbol{\lambda}}\rangle+\frac12({\boldsymbol{\lambda}}'-{\boldsymbol{\lambda}})^\top J({\boldsymbol{\lambda}}'-{\boldsymbol{\lambda}})$, a $K$-dimensional QP.

\begin{algorithm}[h]
\caption{ICNN proximal step (one outer iteration of Algorithm~\ref{alg:mwapg})}\label{alg:icnn}
\begin{algorithmic}[1]
\REQUIRE network $\vartheta$ with $\nabla\varphi_\vartheta(\xi_j)=X^k(\xi_j)$; arrays $Y$, $X^k$; weights ${\boldsymbol{\lambda}}$; $L$
\STATE $g_i\gets\nabla V_i(Y(\xi_j))_j$; $c_i\gets\frac1n\sum_j[V_i(Y(\xi_j))-V_i(X^k(\xi_j))]$ \COMMENT{$\tH(X^k)$ is common, dropped}
\STATE draw fresh $\xi'_1,\dots,\xi'_n\sim\nu$
\STATE $\vartheta\gets\textsc{Fit}(\vartheta,T_{\boldsymbol{\lambda}})$; $p\gets\tilde\psi(\nabla\varphi_\vartheta)$
\FOR[dual Newton]{$r=1,\dots,4$}
\STATE \textbf{if} $\max_ip_i-\langle \boldsymbol{\lambda},p\rangle<10^{-7}$ \textbf{then break}
\STATE $J_{:,j}\gets[\tilde\psi(\nabla\varphi_{\textsc{Fit}(\vartheta,T_{{\boldsymbol{\lambda}}+he_j})})-p]/h$, $j=1,\dots,K$; symmetrize, clip eigenvalues $<0$
\STATE ${\boldsymbol{\lambda}}\gets\arg\max_{{\boldsymbol{\lambda}}'\in\Delt}\langle p,{\boldsymbol{\lambda}}'-{\boldsymbol{\lambda}}\rangle+\frac12({\boldsymbol{\lambda}}'-{\boldsymbol{\lambda}})^\top J({\boldsymbol{\lambda}}'-{\boldsymbol{\lambda}})$
\STATE $\vartheta\gets\textsc{Fit}(\vartheta,T_{\boldsymbol{\lambda}})$; $p\gets\tilde\psi(\nabla\varphi_\vartheta)$
\ENDFOR
\STATE \textbf{return} $X^{k+1}=\nabla\varphi_\vartheta(\xi_j)_j$, ${\boldsymbol{\lambda}}^k={\boldsymbol{\lambda}}$, $\delta_k=\max_ip_i-\langle \boldsymbol{\lambda},p\rangle$
\end{algorithmic}
\end{algorithm}

\paragraph{Cost.}
One fit costs $O(\text{iters}\cdot n\,w\,d^2)$ operations plus $n$ determinants of $d\times d$ matrices. One outer iteration
performs at most $1+4(K+1)$ fits. There are no $O(n^2)$ particle interactions. In our numpy implementation on one CPU core,
an outer iteration takes 1.1--2.0\,s for $n=500$, $n_h=32$, $K=4$, $d=2$.

\paragraph{Hyperparameters.}
\begin{center}
\small
\begin{tabular}{lll}
\toprule
 & Experiment 2 & Validation (App.~\ref{app:icnnval})\\
\midrule
particles $n$ & 500 & 4000\\
ICNN width $n_h$ & 32 & 16\\
penalty $\varrho$ & $10^{-3}$ & $10^{-2}$\\
L-BFGS iterations per fit & 40 & 40\\
dual Newton steps, FD step $h$, tolerance & 4, $0.05$, $10^{-7}$ & 4, $0.05$, $10^{-7}$\\
step size & $1/L$ & $1/L$\\
$X^0$ & $\frac12\xi+(1,-2)$ & $\frac12\xi+(1,-2)$\\
\bottomrule
\end{tabular}
\end{center}

%---------------------------------------------------------------------------------------------------------------------

\section{On the discretization of A-MWGraD}
\label{app:amwgrad}

A-MWGraD~\cite{nguyen2026accelerated} discretizes a damped second-order flow with velocity $V$ and the combined field
$\bar\varphi(X)=\sum_i\lambda_i\varphi_i(X)$, $\varphi_i=-\nabla V_i-\nabla\log\mu$. Two orders of the updates are possible, with step $h$ and
$\eta=h^2$.

\paragraph{Explicit order (as printed).}
$X_{n+1}=X_n+hV_n$, $V_{n+1}=a_nV_n+h\bar\varphi(X_n)$. Eliminating the velocity,
$X_{n+1}-X_n=hV_n=a_{n-1}hV_{n-1}+\eta\bar\varphi(X_{n-1})$, i.e.
\[
  X_{n+1}=X_n+a_{n-1}(X_n-X_{n-1})+\eta\bar\varphi(X_{n-1}):
\]
momentum with a \emph{delayed} gradient. On a quadratic with curvature $\omega>0$ along some direction ($\bar\varphi(x)=-\omega x$)
and frozen $a$, the characteristic polynomial is $z^2-(1+a)z+(a+\eta\omega)$. By the Jury criterion for $z^2+c_1z+c_0$
($|c_0|<1$ and $|c_1|<1+c_0$), the iteration is stable if and only if $a+\eta\omega<1$, i.e.\ $\eta\omega<1-a$. With
$a_n=(n-1)/(n+2)\to1$ this fails for every fixed step. With the strongly convex momentum $a=(1-\sqrt{m\eta})/(1+\sqrt{m\eta})$ one
has $1-a\approx2\sqrt{m\eta}$, and stability for all curvatures $\omega\le\lambda_{\max}$ requires $\eta<4m/\lambda_{\max}^2$, i.e.\ a step
$\kap$ times smaller than $1/L$.

\paragraph{Semi-implicit order (heavy ball).}
$V_{n+1}=a_nV_n+h\bar\varphi(X_n)$, $X_{n+1}=X_n+hV_{n+1}$, i.e.\ $X_{n+1}=X_n+a_n(X_n-X_{n-1})+\eta\bar\varphi(X_n)$. The
characteristic polynomial $z^2-(1+a-\eta\omega)z+a$ is stable if and only if $0<\eta\omega<2(1+a)$, which holds for
$\eta=1/L$. For $a$ close to 1 the roots have modulus $\sqrt a$ and the iteration oscillates; in our runs the merit of plain
heavy-ball A-MWGraD varied over several orders of magnitude (Figure~\ref{fig:app-exp1}).

\paragraph{Safeguards used in the experiments.}
\emph{Gradient-norm restart:} let $\gamma_n=\lambda_n^\top Q_n\lambda_n=\|\bar\varphi(X_n)\|^2_{L^2(\mu_n)}$; if $\gamma_n>\gamma_{n-1}$, set
$X_{n-1}\gets X_n$ (zero velocity) and reset the momentum counter $j\gets0$; then $a=\max(0,(j-1)/(j+2))$. \emph{Strongly convex
variant:} $a=\min(\beta',(j-1)/(j+2))$ with $\beta'=(1-\sqrt{m\eta})/(1+\sqrt{m\eta})$; without the ramp, the constant momentum
immediately after a restart caused repeated restarts. Our methods evaluate the gradient at the extrapolated point and treat
the entropy implicitly, so they need neither safeguard.

\paragraph{Gaussian closed form of the baselines.}
MWGraD and A-MWGraD move particles. They do not update transport maps from a fixed base measure, and they have no
projection onto $\mathcal{K}_\nu$. MWGraD moves each particle by $x\gets x+\eta\bar\varphi(x)$, where $\bar\varphi=\sum_i\lambda_i\varphi_i$
and $\varphi_i=-\nabla V_i-\nabla\log\mu$. A-MWGraD gives each particle a velocity and uses the heavy-ball update of
Section~\ref{sec:exp}. If $\mu=\mathcal N(b,\Sigma)$, the exact score is affine, so every field is affine:
$\varphi_i(x)=B_i(x-b)+\beta_i$ with $B_i=\Sigma^{-1}-P_i$ and $\beta_i=-P_i(b-m_i)$. An affine field keeps a Gaussian
Gaussian. For MWGraD, the update is the pushforward of $\mu_k$ by $\mathrm{Id}+\eta\bar\varphi$:
\[
  b\gets b+\eta\bar\beta,\qquad \Sigma\gets(I+\eta\bar B)\,\Sigma\,(I+\eta\bar B)^\top .
\]
For A-MWGraD, we also need the velocity of each particle. We write the position of the particle with label
$\xi\sim\mathcal N(0,I)$ as $x=A\xi+b$, with $\Sigma=AA^\top$. Since the fields are affine, the positions and the velocities
stay affine in $\xi$. So the particle system with infinitely many particles is given exactly by $(A,b)$ and the velocity
$(V_A,V_b)$: MWGraD is $A\gets A+\eta\bar BA$, $b\gets b+\eta\bar\beta$, and A-MWGraD applies the heavy-ball update to $(A,b)$.
Here $A$ only labels the particles. It is not symmetric in general, so $\xi\mapsto A\xi+b$ is not an element of $\mathcal{K}_\nu$.
The min-norm weights solve $\min_{{\boldsymbol{\lambda}}\in\Delt}{\boldsymbol{\lambda}}^\top Q{\boldsymbol{\lambda}}$ with
$Q_{ij}=\langle\varphi_i,\varphi_j\rangle_{L^2(\mu)}=\mathrm{tr}(B_i\Sigma B_j)+\beta_i^\top\beta_j$. With the exact score, this
is the most favourable setting for the baselines. In Experiment~3 the targets are not Gaussian. There we project each field on
the affine fields, $\varphi_i(x)\approx C_i(x-m)+d_i$ with $C_i=-\E[\nabla V_i(x)(x-m)^\top]\Sigma^{-1}+\Sigma^{-1}$ and
$d_i=-\E[\nabla V_i(x)]$, where the expectations use the 1024 base samples $x_j=m+\Sigma^{1/2}\xi_j$, the same samples as for our
methods. The velocity of A-MWGraD is an affine field $C_v(x-m)+d_v$ of the current position; the update is given below.

\paragraph{Gaussian version of A-MWGraD.}
A-MWGraD~\cite{nguyen2026accelerated} is stated for particles, but its flow preserves Gaussians, and this gives an exact
Gaussian version. Recall the flow (Eq.~13 of~\cite{nguyen2026accelerated}):
\begin{equation}\label{eq:amw-flow}
  \partial_t\rho_t+\nabla\cdot(\rho_t\nabla\Phi_t)=0,\qquad
  \partial_t\Phi_t+\alpha_t\Phi_t+\tfrac12|\nabla\Phi_t|^2+\textstyle\sum_k \lambda_{t,k}\,\delta F_k[\rho_t]=0,
\end{equation}
with $F_k=\mathrm{KL}(\cdot\|\pi_k)$, $\delta F_k[\rho]=V_k+\log\rho$ up to a constant, and the min-norm weights
$\boldsymbol{\lambda}_t=\arg\min_{\boldsymbol{\lambda}\in\Delt}\int|\sum_k\lambda_k\nabla\delta F_k[\rho_t]|^2d\rho_t$. For Gaussian targets
$\pi_k=\mathcal N(\mu_k,P_k^{-1})$ and $\rho=\mathcal N(m,\Sigma)$, the descent fields are affine:
$-\nabla\delta F_k[\rho](x)=B_k(x-m)+\beta_k$ with $B_k=\Sigma^{-1}-P_k$ and $\beta_k=-P_k(m-\mu_k)$. We write
$\bar B=\sum_k\lambda_kB_k$, $\bar\beta=\sum_k\lambda_k\beta_k$, and $Q_{kl}=\mathrm{tr}(B_k\Sigma B_l)+\beta_k^\top\beta_l$, so that
$\boldsymbol{\lambda}=\arg\min_{\boldsymbol{\lambda}\in\Delt}\boldsymbol{\lambda}^\top Q\boldsymbol{\lambda}$.

\begin{proposition}[Gaussian A-MWGraD flow]\label{prop:amw-gauss}
Let $(m_t,\Sigma_t,d_t,S_t)$ solve
\begin{equation}\label{eq:amw-gauss-ode}
  \dot m_t=d_t,\qquad \dot\Sigma_t=S_t\Sigma_t+\Sigma_tS_t,\qquad
  \dot d_t=-\alpha_td_t+\bar\beta_t,\qquad \dot S_t=-\alpha_tS_t-S_t^2+\bar B_t,
\end{equation}
with $\Sigma_0\succ0$, $d_0=0$, $S_0=0$, where $\bar B_t,\bar\beta_t$ and the weights are evaluated at $(m_t,\Sigma_t)$.
Then, on every interval on which the solution exists, $S_t$ is symmetric and $\Sigma_t\succ0$, and the pair
$\rho_t=\mathcal N(m_t,\Sigma_t)$, $\Phi_t(x)=\frac12(x-m_t)^\top S_t(x-m_t)+d_t^\top x$ solves~\eqref{eq:amw-flow} with
$\Phi_0=0$. For centered targets ($\mu_k=0$, $m_0=0$) we get $m_t=0$, $d_t=0$, and with $\tilde S=S/2$ the equations become
$\dot\Sigma_t=2(\tilde S_t\Sigma_t+\Sigma_t\tilde S_t)$ and
$\dot{\tilde S}_t=-\alpha_t\tilde S_t-2\tilde S_t^2-\sum_k\lambda_{t,k}\nabla_\Sigma F_k^\dagger(\Sigma_t)$, the accelerated counterpart
of Proposition~3.1 of~\cite{nguyen2026accelerated}.
\end{proposition}

\begin{proof}
\emph{Symmetry and positivity.} $\bar B_t$ and $S_t^2$ are symmetric and $S_0=0$, so $S_t^\top$ solves the same equation as
$S_t$, and $S_t=S_t^\top$. Let $\dot U_t=S_tU_t$, $U_0=I$. Then $U_t\Sigma_0U_t^\top$ solves the $\Sigma$-equation, so
$\Sigma_t=U_t\Sigma_0U_t^\top$, and $\det U_t=\exp(\int_0^t\mathrm{tr}\,S_s\,ds)>0$ gives $\Sigma_t\succ0$.
\emph{Continuity equation.} The velocity $\nabla\Phi_t(x)=S_t(x-m_t)+d_t$ is affine, so it maps Gaussians to Gaussians, and
$\rho_t=\mathcal N(m_t,\Sigma_t)$ solves the continuity equation if and only if $\dot m_t=d_t$ and
$\dot\Sigma_t=S_t\Sigma_t+\Sigma_tS_t$.
\emph{Hamilton--Jacobi equation.} Since the potential is defined up to a function of $t$, it suffices to check the gradient
of the second equation in~\eqref{eq:amw-flow}:
$\partial_t\nabla\Phi_t+\alpha_t\nabla\Phi_t+\nabla^2\Phi_t\nabla\Phi_t-\sum_k\lambda_{t,k}(B_k(x-m_t)+\beta_k)=0$. At a fixed $x$,
$\partial_t\nabla\Phi_t=\dot S_t(x-m_t)-S_t\dot m_t+\dot d_t$ and $\nabla^2\Phi_t\nabla\Phi_t=S_t^2(x-m_t)+S_td_t$. With $\dot m_t=d_t$,
the left-hand side equals $(\dot S_t+\alpha_tS_t+S_t^2-\bar B_t)(x-m_t)+(\dot d_t+\alpha_td_t-\bar\beta_t)$, which is zero
by~\eqref{eq:amw-gauss-ode}. The weights are the min-norm weights, because
$\int|\sum_k\lambda_k(B_k(x-m)+\beta_k)|^2d\rho=\boldsymbol{\lambda}^\top Q\boldsymbol{\lambda}$.
\end{proof}

\paragraph{Discrete Gaussian A-MWGraD (as used in our experiments).}
We discretize the flow with the stabilized particle update of Section~\ref{sec:exp}, with $h=\sqrt\eta$: the velocity is updated
first, $v^{n+1}=a_nv^n+h\,\bar\varphi_n(x^n)$, and the particles move with the new velocity, $x^{n+1}=x^n+hv^{n+1}$. We store the
velocity as an affine field of the current position, $v^n(x)=S_n(x-m_n)+d_n$. Since the fields are affine, this is exact for
infinitely many particles:
\begin{equation}\label{eq:amw-gauss-disc}
\begin{aligned}
  S'&=a_nS_n+h\bar B_n, & d_{n+1}&=a_nd_n+h\bar\beta_n, & T_n&=I+hS',\\
  m_{n+1}&=m_n+h\,d_{n+1}, & \Sigma_{n+1}&=T_n\Sigma_nT_n^\top, & S_{n+1}&=S'\,T_n^{-1},
\end{aligned}
\end{equation}
with $S_0=0$, $d_0=0$. The last update writes the velocity of the moved particles as a function of their new position. The
momentum coefficient is $a_n=(j-1)/(j+2)$, or $\min(\beta',(j-1)/(j+2))$ in the strongly convex variant, where $j$ counts the steps
since the last restart; the velocity is set to zero ($S=0$, $d=0$, $j=0$) whenever
$\|\bar\varphi_n\|_{L^2(\rho_n)}^2=\mathbf w^\top Q\mathbf w$ increases. The heavy-ball order and the restart are our changes to
Algorithm~2 of~\cite{nguyen2026accelerated}; they make the method stable at step $1/L$ and favour it.

%---------------------------------------------------------------------------------------------------------------------
\section{Additional experimental details and results}
\label{app:exp}

\subsection{Experiment 1: all condition numbers and final distributions}
\label{app:exp1}

\begin{figure}[h]
\centering
\includegraphics[width=0.6\linewidth]{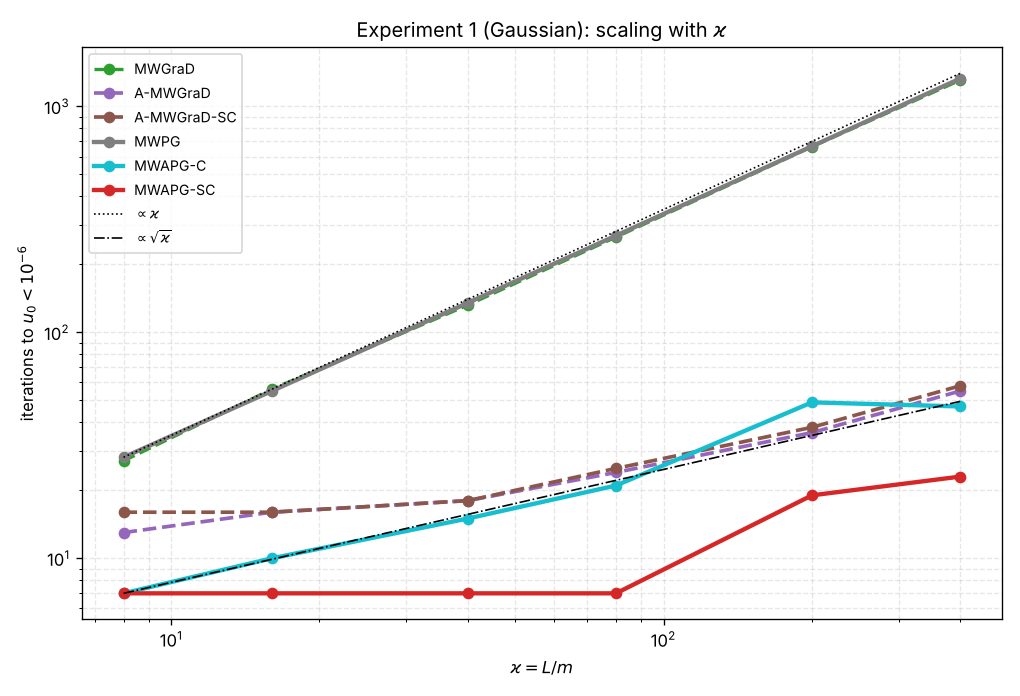}
\caption{Experiment 1: iterations to $u_0<10^{-6}$ against $\kap$, with reference slopes $\kap$ and $\sqrt\kap$.}
\label{fig:exp1-scaling}
\end{figure}

\begin{figure}[h]
\centering
\includegraphics[width=0.62\linewidth]{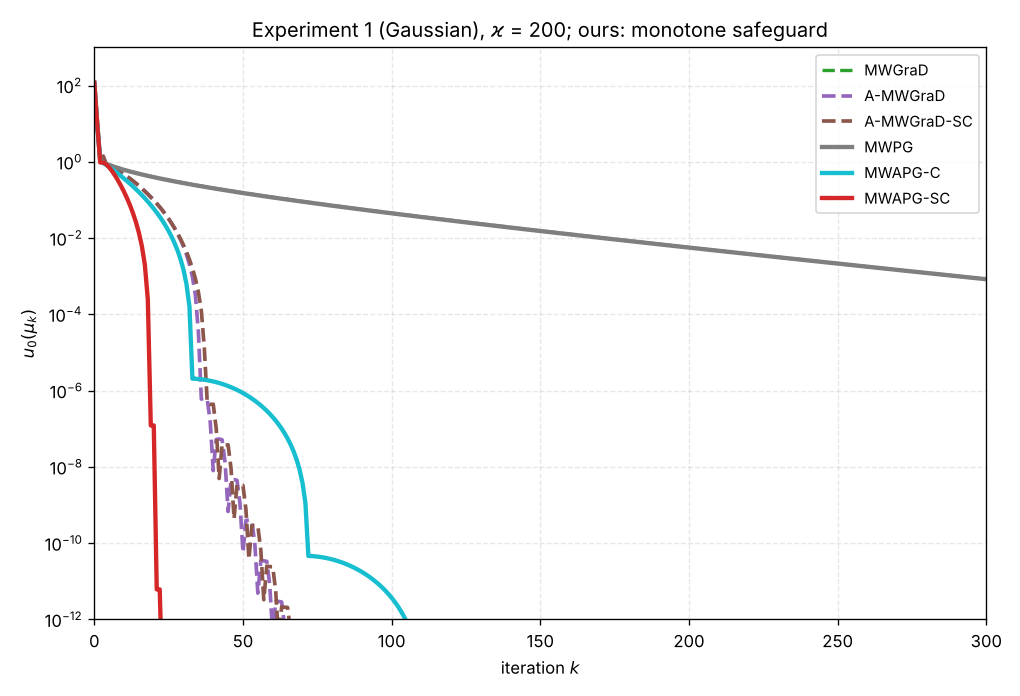}
\caption{Experiment 1, $\kap=200$: merit $u_0(\mu_k)$ along the iterations. MWGraD (dashed green) lies under MWPG (grey).}
\label{fig:exp1-curves}
\end{figure}

\begin{figure}[h]
\centering
\includegraphics[width=0.49\linewidth]{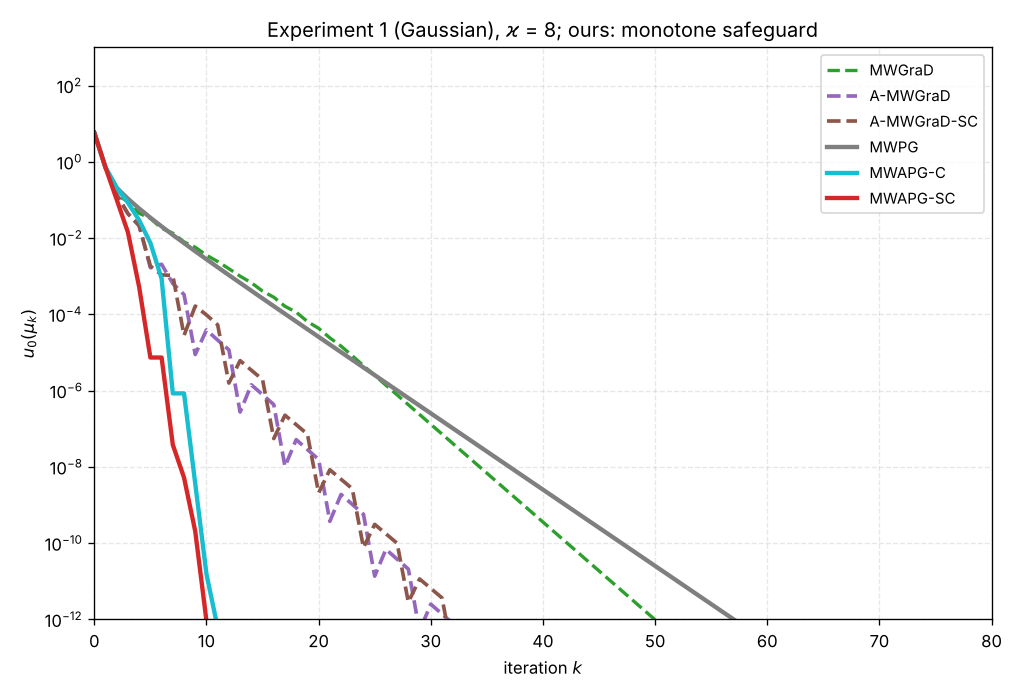}\hfill
\includegraphics[width=0.49\linewidth]{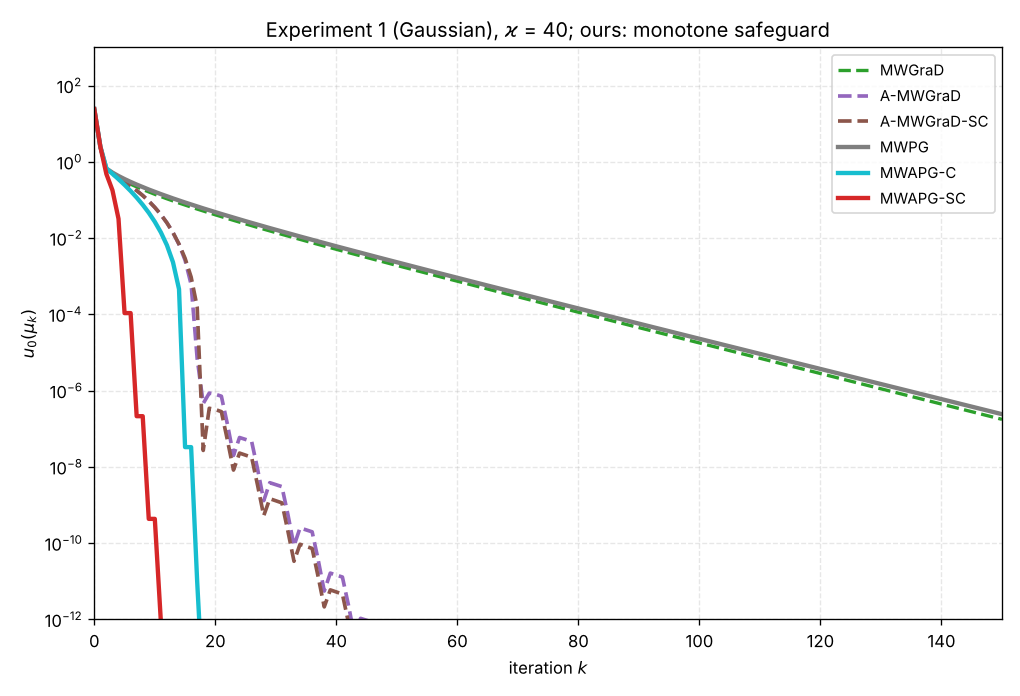}
\caption{Experiment 1: $u_0(\mu_k)$ for $\kap=8$ (left) and $\kap=40$ (right).}
\label{fig:app-exp1}
\end{figure}

\begin{figure}[h]
\centering
\includegraphics[width=0.78\linewidth]{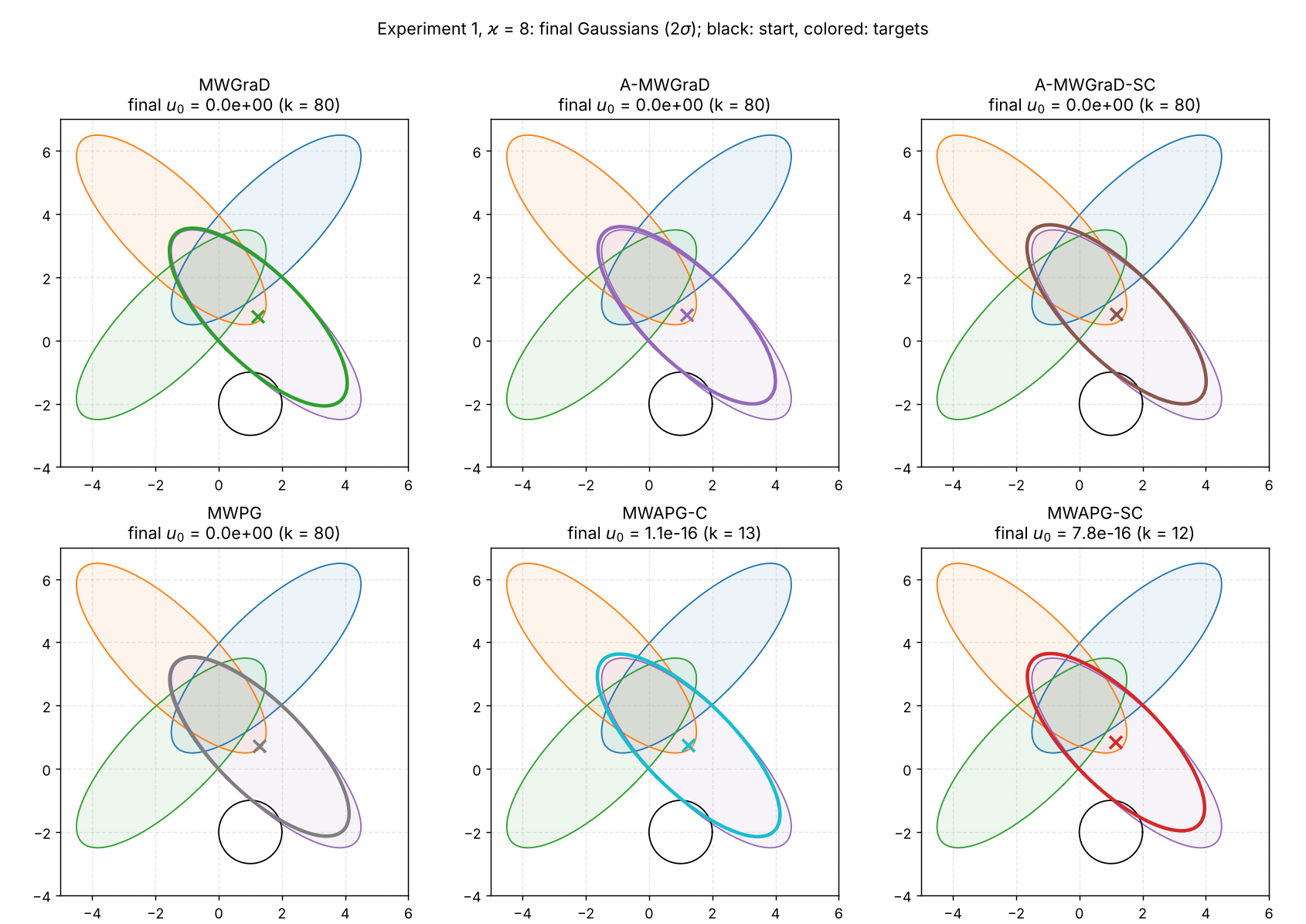}
\caption{Experiment 1, $\kap=8$: final Gaussians ($2\sigma$ ellipses) of all methods; black: $\mu_0$, coloured: targets.}
\label{fig:app-shapes}
\end{figure}

\begin{figure}[h]
\centering
\includegraphics[width=0.78\linewidth]{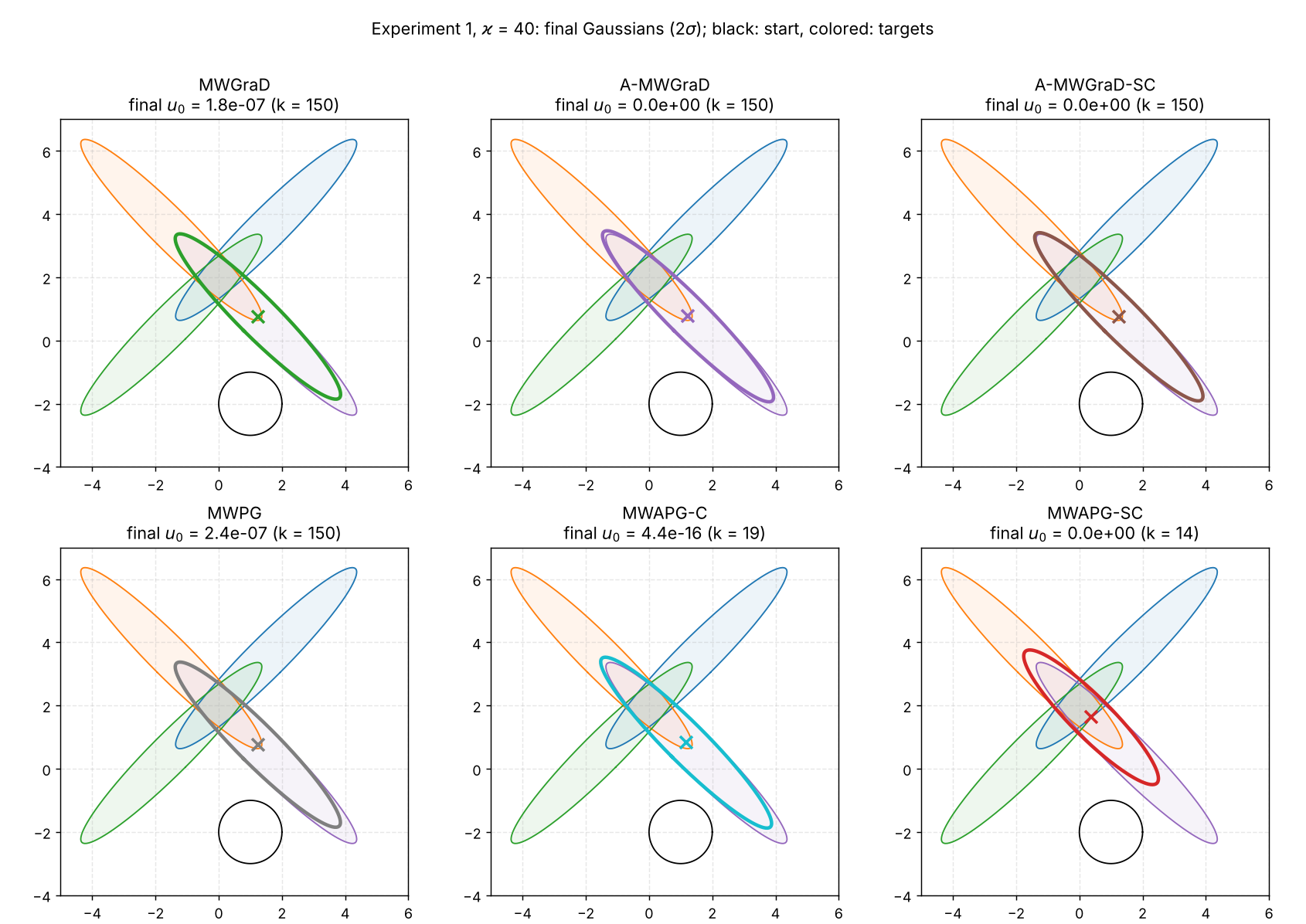}
\caption{Experiment 1, $\kap=40$: final Gaussians ($2\sigma$ ellipses).}
\label{fig:app-shapes40}
\end{figure}

\begin{figure}[h]
\centering
\includegraphics[width=0.78\linewidth]{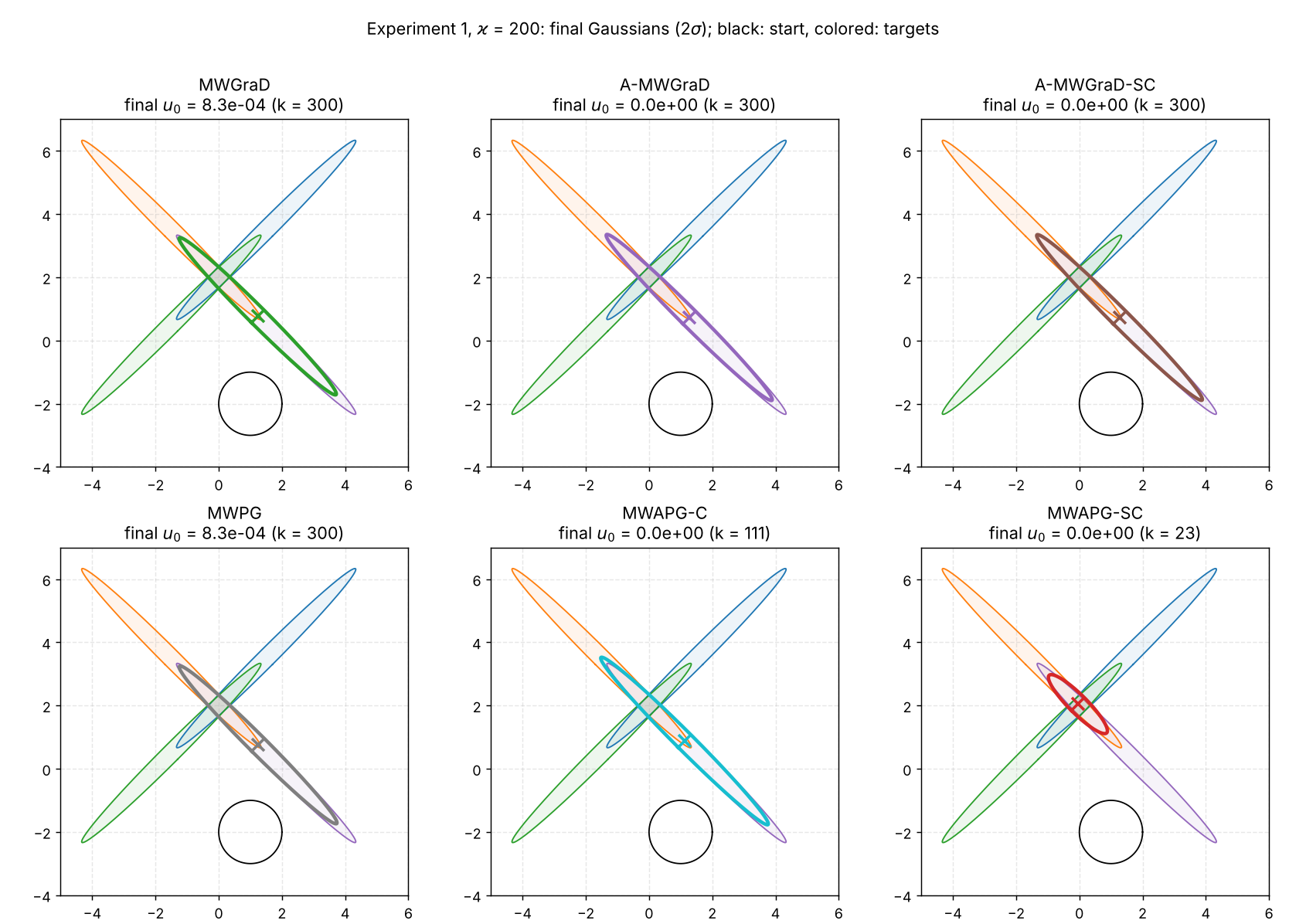}
\caption{Experiment 1, $\kap=200$: final Gaussians ($2\sigma$ ellipses).}
\label{fig:app-shapes200}
\end{figure}

Figure~\ref{fig:app-exp1} complements Figure~\ref{fig:exp1-curves} with $\kap=8$ and $\kap=40$. The ordering of the methods
is the same at every $\kap$; the gaps grow with $\kap$. Figures~\ref{fig:app-shapes}--\ref{fig:app-shapes200} show the final Gaussians. All methods converge
to Pareto-optimal Gaussians $\pi_{\boldsymbol{\lambda}}$, which become elongated along the diagonal of targets 2 and 4 as $\kap$ grows. The
accelerated methods end at different Pareto points than the plain ones, closer to the common region around $c$: the Pareto
point reached depends on the trajectory, as for all methods of this type, and none of the methods is designed to reach a
particular one.

\subsection{Monotone safeguard}
Table~\ref{tab:exp1-variants} compares the three versions of our accelerated methods. The plain versions overshoot; the
monotone versions never increase any objective. Since $u_0$ is order-preserving (if $G_i(\mu')\le G_i(\mu)$ for all $i$, then
$u_0(\mu')\le u_0(\mu)$ by~\eqref{eqn:defofmeritfunction}), $u_0$ is then monotone as well. Without restart, the monotone
versions can be slower (MWAPG-C at $\kap=200$: 89 vs.\ 34 iterations), because a rejected step keeps the momentum that caused
it; the restart removes most of this cost, and for MWAPG-SC it is even faster than the plain version for $\kap\le80$.
The figures of the paper use the monotone versions with restart.

\begin{table}[h]
\centering
\caption{Experiment 1: iterations to $u_0<10^{-6}$ of the plain, monotone, and monotone-with-restart versions of our accelerated
methods (Algorithm~\ref{alg:mwapg}), and the largest rise of $u_0$ along the trajectory (orders of magnitude, over all $\kap$, while $u_0>10^{-9}$).}
\label{tab:exp1-variants}
\small
\begin{tabular}{rrrrrrr}
\toprule
 & \multicolumn{3}{c}{MWAPG-C} & \multicolumn{3}{c}{MWAPG-SC}\\
\cmidrule(lr){2-4}\cmidrule(lr){5-7}
$\kap$ & plain & monotone & mono.+restart & plain & monotone & mono.+restart\\
\midrule
8 & 7 & 7 & 7 & 8 & 9 & 7 \\
16 & 10 & 10 & 10 & 9 & 12 & 7 \\
40 & 15 & 15 & 15 & 11 & 19 & 7 \\
80 & 21 & 21 & 21 & 9 & 21 & 7 \\
200 & 34 & 89 & 49 & 21 & 30 & 19 \\
400 & 47 & 47 & 47 & 20 & 37 & 23 \\
\midrule
max.\ rise of $u_0$ & 1.31 & 0 & 0 & 2.63 & 0 & 0\\
\bottomrule
\end{tabular}
\end{table}

\subsection{Sensitivity to the step size}
\label{app:exp1_stepsize}
\begin{figure}[h]
\centering
\includegraphics[width=0.85\linewidth]{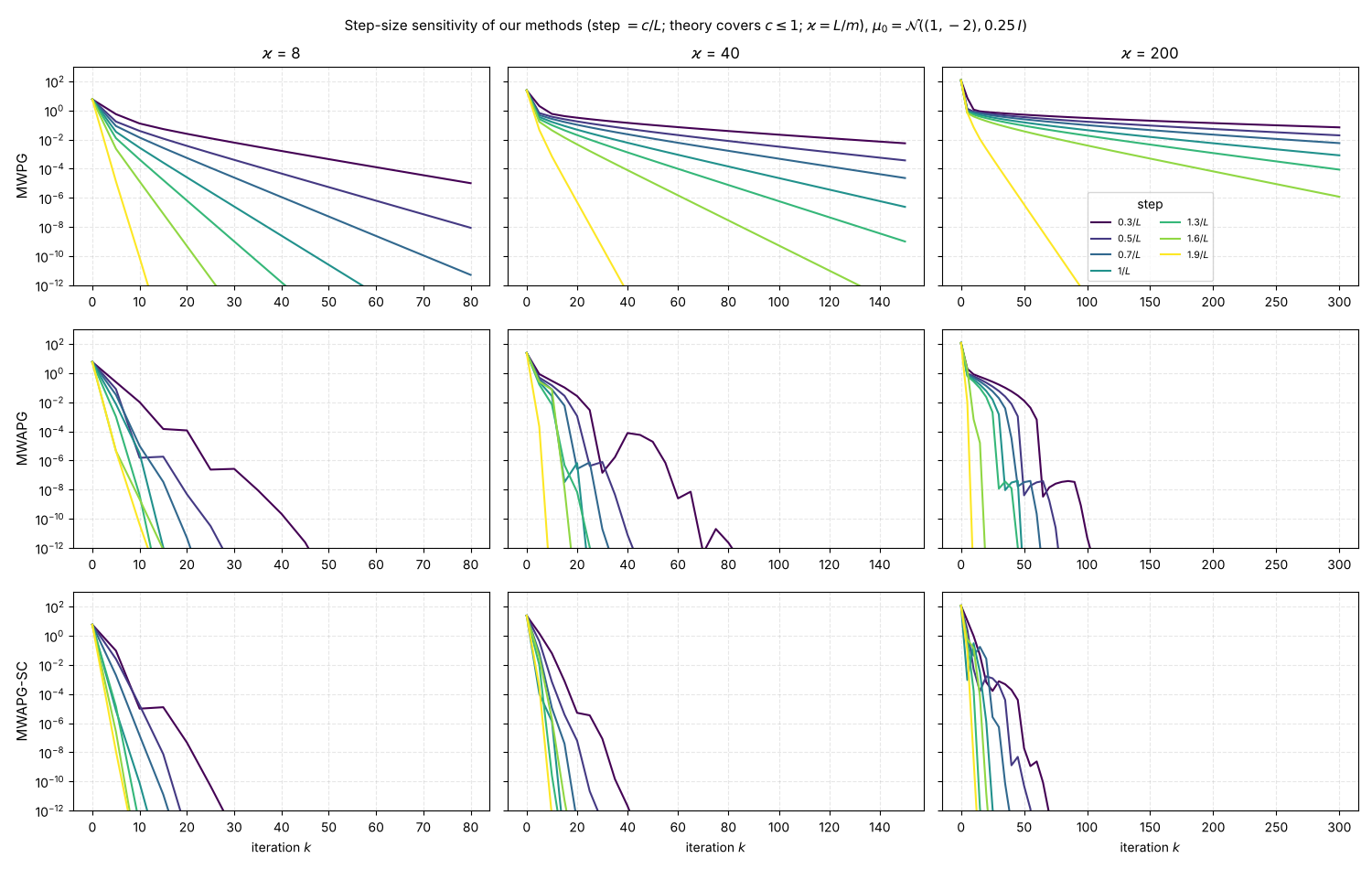}
\caption{Step-size sensitivity of MWPG, MWAPG-C and MWAPG-SC on the Experiment 1 targets ($\kap=8,40,200$).}
\label{fig:app-steps}
\end{figure}

The theory covers steps $c/L$ with $c\le1$, by replacing $L$ with $L/c$. Figure~\ref{fig:app-steps} runs our three methods with
$c=0.3$, $0.5$, $0.7$, $1$, $1.3$, $1.6$ and $1.9$ on the Experiment 1 targets. All runs converge, also beyond the theoretical range, and
larger steps are faster. The momentum methods show small non-monotone bumps, as expected for accelerated methods; MWAPG-SC is
the least sensitive. In the main experiments we use $c=1$ for all methods, which is the largest step covered by the theory and
is not tuned.

\subsection{Higher dimension}
\label{app:exp1_high_dim}
\begin{figure}[h]
\centering
\includegraphics[width=0.82\linewidth]{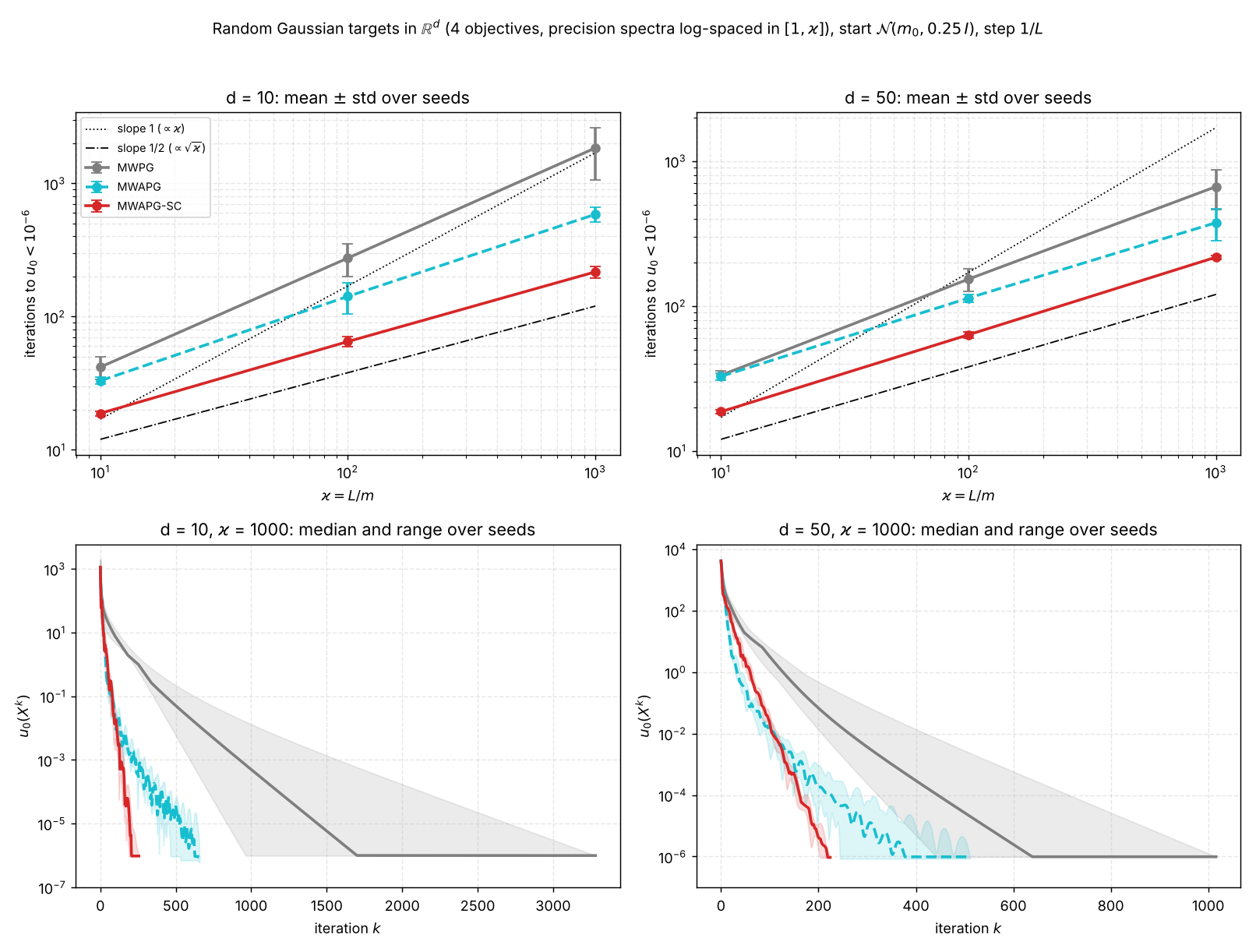}
\caption{Random Gaussian targets in $\R^{10}$ and $\R^{50}$: scaling with $\kap$ (top) and convergence curves at $\kap=1000$ (bottom). Plain (non-monotone) versions of our methods.}
\label{fig:hd}
\end{figure}

We drew four random Gaussian targets in $\R^d$, $d\in\{10,50\}$, with precision spectra log-spaced in $[1,\kap]$, random
rotations and random means, and five seeds per setting, started from $\mathcal N(m_0,0.25I)$. Table~\ref{tab:hd} and
Figure~\ref{fig:hd} report the iterations of our methods to $u_0<10^{-6}$. MWAPG-SC scales with slope $0.53$ in $\kap$ in both
dimensions, with small spread across seeds, in agreement with Theorem~\ref{thm:sc}. MWAPG-C is between MWPG and MWAPG-SC. The
iteration counts do not grow with the dimension.

\begin{table}[h]
\centering
\caption{Random Gaussian targets in $\R^d$: iterations to $u_0<10^{-6}$, mean $\pm$ std over 5 seeds.}
\label{tab:hd}
\small
\begin{tabular}{rrrrr}
\toprule
$d$ & $\kap$ & MWPG & MWAPG-C & MWAPG-SC\\
\midrule
10 & 10 & $42\pm8$ & $33\pm2$ & $19\pm1$\\
10 & 100 & $275\pm75$ & $142\pm38$ & $65\pm6$\\
10 & 1000 & $1846\pm779$ & $586\pm76$ & $217\pm22$\\
50 & 10 & $33\pm3$ & $33\pm2$ & $19\pm1$\\
50 & 100 & $153\pm27$ & $113\pm7$ & $63\pm3$\\
50 & 1000 & $665\pm203$ & $375\pm95$ & $217\pm7$\\
\bottomrule
\end{tabular}
\end{table}

\subsection{Validation of the ICNN implementation against the exact method}
\label{app:icnnval}

\begin{figure}[h]
\centering
\includegraphics[width=\linewidth]{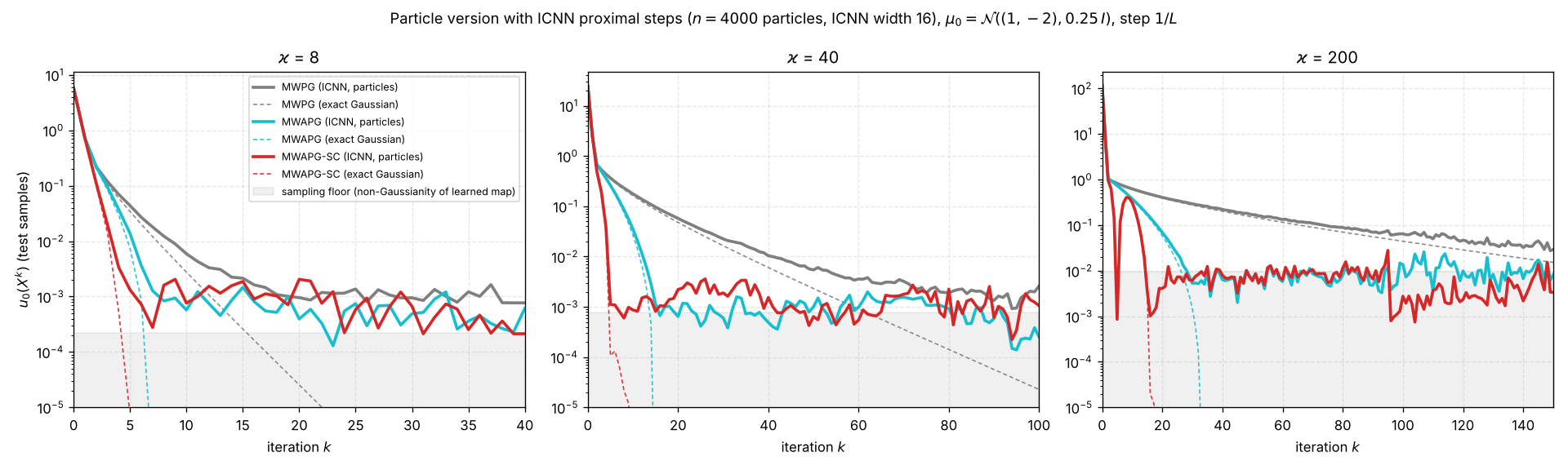}
\caption{ICNN particle implementation (solid) versus exact Gaussian iterates (dashed), Experiment 1 targets. Grey band: floor
set by the non-Gaussianity of the learned map.}
\label{fig:icnnval}
\end{figure}

\begin{figure}[h]
\centering
\includegraphics[width=\linewidth]{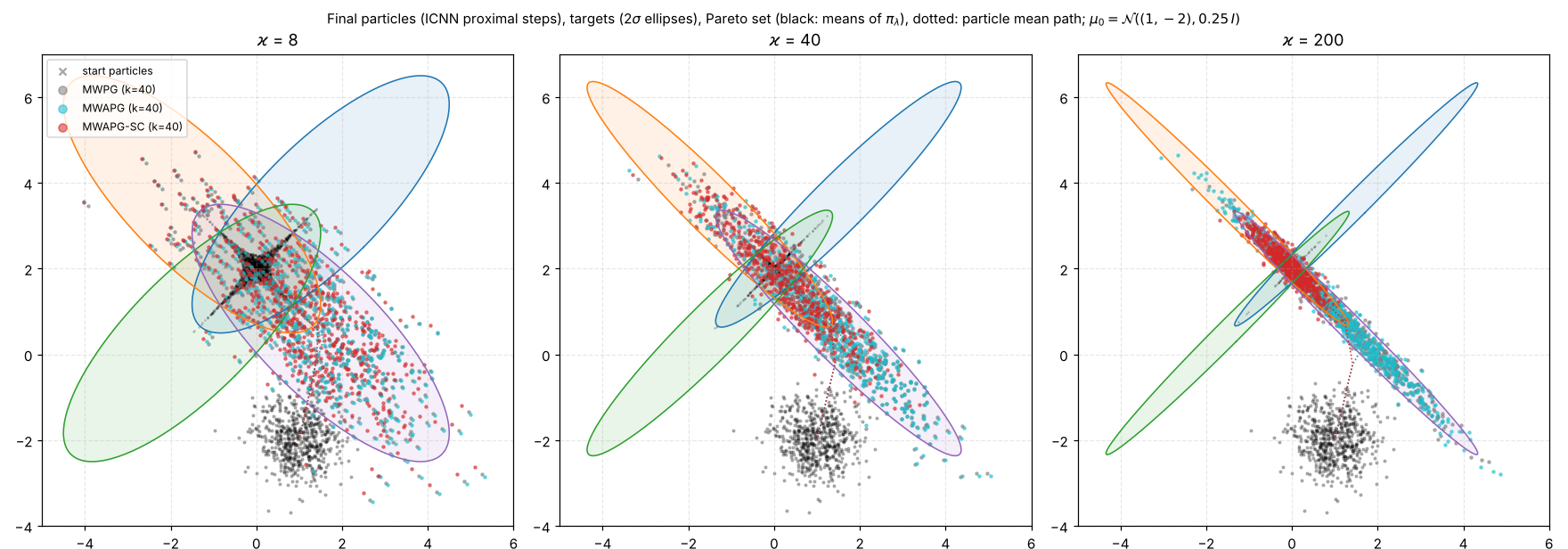}
\caption{Final particles of the ICNN runs ($k=40$), targets ($2\sigma$ ellipses), means of the Pareto points $\pi_{\boldsymbol{\lambda}}$ (black)
and the path of the particle mean (dotted).}
\label{fig:icnnval-part}
\end{figure}

To check that the particle implementation follows Algorithm~\ref{alg:mwapg}, we ran it on the Gaussian targets of Experiment 1,
where the exact iterates are known in closed form, with $n=4000$ particles, width 16 and the same start
$\mu_0=\mathcal N((1,-2),0.25I)$. Since the targets are Gaussian, $\mathrm{KL}(\mu\|\pi_{\boldsymbol{\lambda}})=\mathrm{KL}(\mathcal N(\bar m,C)\|\pi_{\boldsymbol{\lambda}})+[h(\mathcal N(\bar m,C))-h(\mu)]$,
where $\bar m,C$ are the mean and covariance of $\mu$ and $h$ is the differential entropy. The second bracket, the
non-Gaussianity of $\mu$, is computed exactly from the network, $h(\mu)=h(\nu)+\E_\nu\log\det\nabla^2\varphi_\vartheta$, on $2\cdot10^4$
test samples. Figure~\ref{fig:icnnval} shows that the particle iterates follow the exact Gaussian iterates of all three methods
until they reach the floor given by the non-Gaussianity of the learned map ($\approx10^{-3}$ for $\kap\le40$, $\approx10^{-2}$ for
$\kap=200$), and that the acceleration is preserved: at $\kap=200$, MWAPG-C and MWAPG-SC reach the floor in about 20 iterations,
while MWPG has not reached it after 150. Figure~\ref{fig:icnnval-part} shows the final particles, which cover the Pareto-optimal
Gaussians reached by the exact methods.

\subsection{Experiment 2: details, wall-clock time}
\label{app:exp2}

\begin{figure*}[h]
\centering
\includegraphics[width=0.9\linewidth]{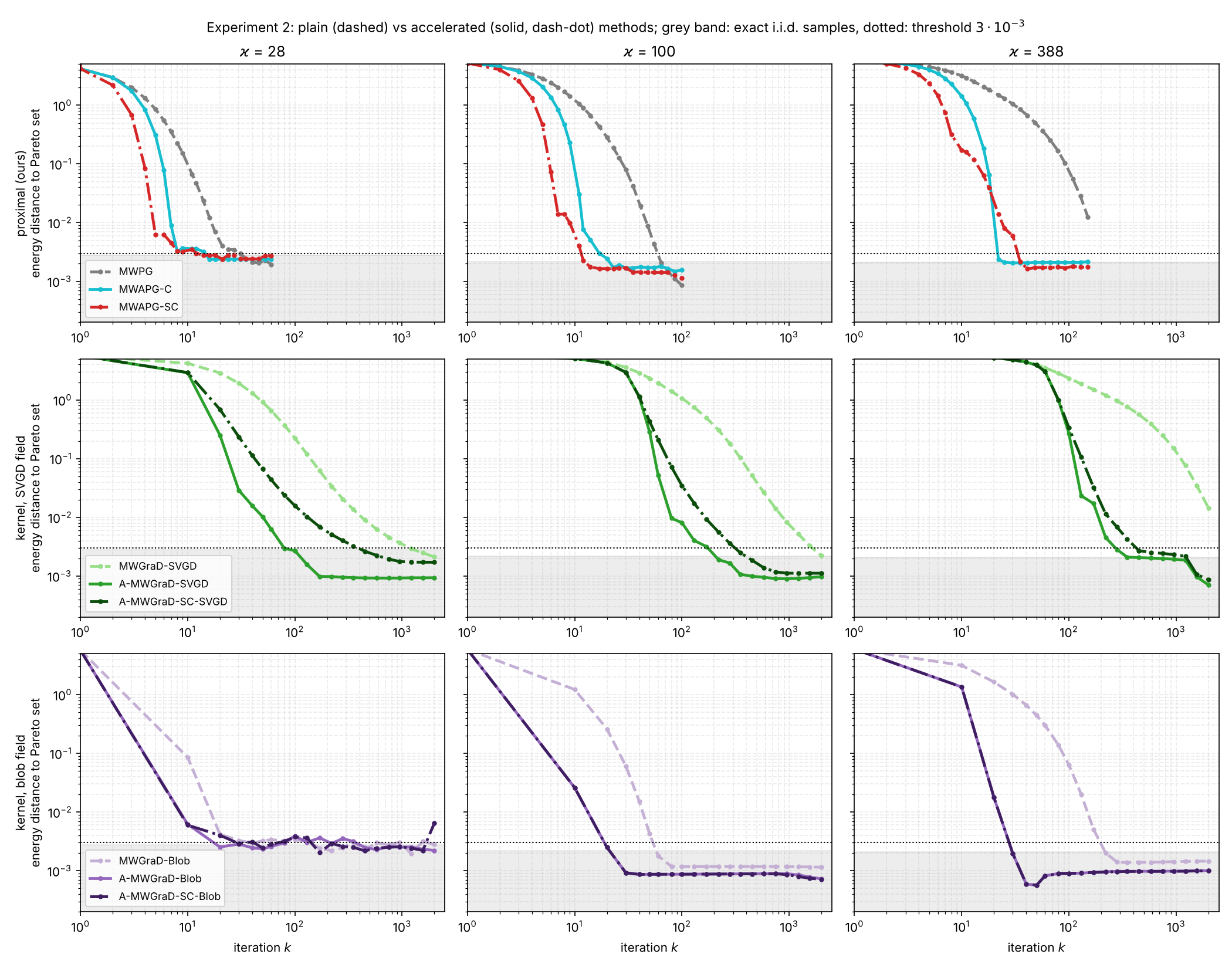}
\caption{Experiment 2: energy distance to the Pareto set against iteration. Rows: method families; dashed: plain method, solid
and dash-dot: accelerated methods. Grey band: exact i.i.d.\ samples; dotted line: threshold.}
\label{fig:exp2-curves}
\end{figure*}

\paragraph{Pareto densities and metric.}
The Pareto densities $\pi_{\boldsymbol{\lambda}}\propto e^{-\sum_i\lambda_iV_i}$ are tabulated on a $401\times401$ grid over $[-5,5]\times[-3,7]$, which
contains all their mass, and sampled exactly (grid cell chosen with probability proportional to the density, then uniform
jitter within the cell). The best Gaussian floor $\min_{G,{\boldsymbol{\lambda}}}\mathrm{KL}(G\|\pi_{\boldsymbol{\lambda}})$ is computed by optimizing over Gaussian
parameters and ${\boldsymbol{\lambda}}$ with quadrature on the grid. For the energy distance we use $2000$ reference samples for each of the 165
grid weights ${\boldsymbol{\lambda}}\in\{0,\frac18,\dots,1\}^4\cap\Delt$, and report $\min_{\boldsymbol{\lambda}}[2\E\|X-Y_{\boldsymbol{\lambda}}\|_2-\E\|X-X'\|_2-\E\|Y_{\boldsymbol{\lambda}}-Y'_{\boldsymbol{\lambda}}\|_2]$ over the
500 particles. The i.i.d.\ floor is the mean of this quantity for 500 exact samples of four different $\pi_{\boldsymbol{\lambda}}$.

\paragraph{Kernel baselines.}
SVGD fields: $\varphi_i(x_a)=\frac1n\sum_b[k(x_a,x_b)(-\nabla V_i(x_b))+\nabla_{x_b}k(x_a,x_b)]$ with the Gaussian kernel and bandwidth
$\sigma^2=\mathrm{med}(\|x_a-x_b\|_2^2)/(2\log(n+1))$. Blob fields~\cite{carrillo2019blob}: $\varphi_i=-\nabla V_i-\nabla\log(k*\mu)-\nabla(k*(\mu/(k*\mu)))$
(both terms of the blob formula), with the Mahalanobis kernel $k(x,y)=\exp(-\|C^{-1/2}(x-y)\|_2^2/(2\sigma^2))$, $C$ the particle covariance,
and $\sigma^2$ equal to $0.3$ times the median rule in whitened coordinates. The weights are the exact min-norm solution in
$\mathcal{L}^2(\mu_k)$ (300 projected-gradient iterations, warm-started), and the step is $\eta=1/L$.

\begin{figure}[h]
\centering
\includegraphics[width=\linewidth]{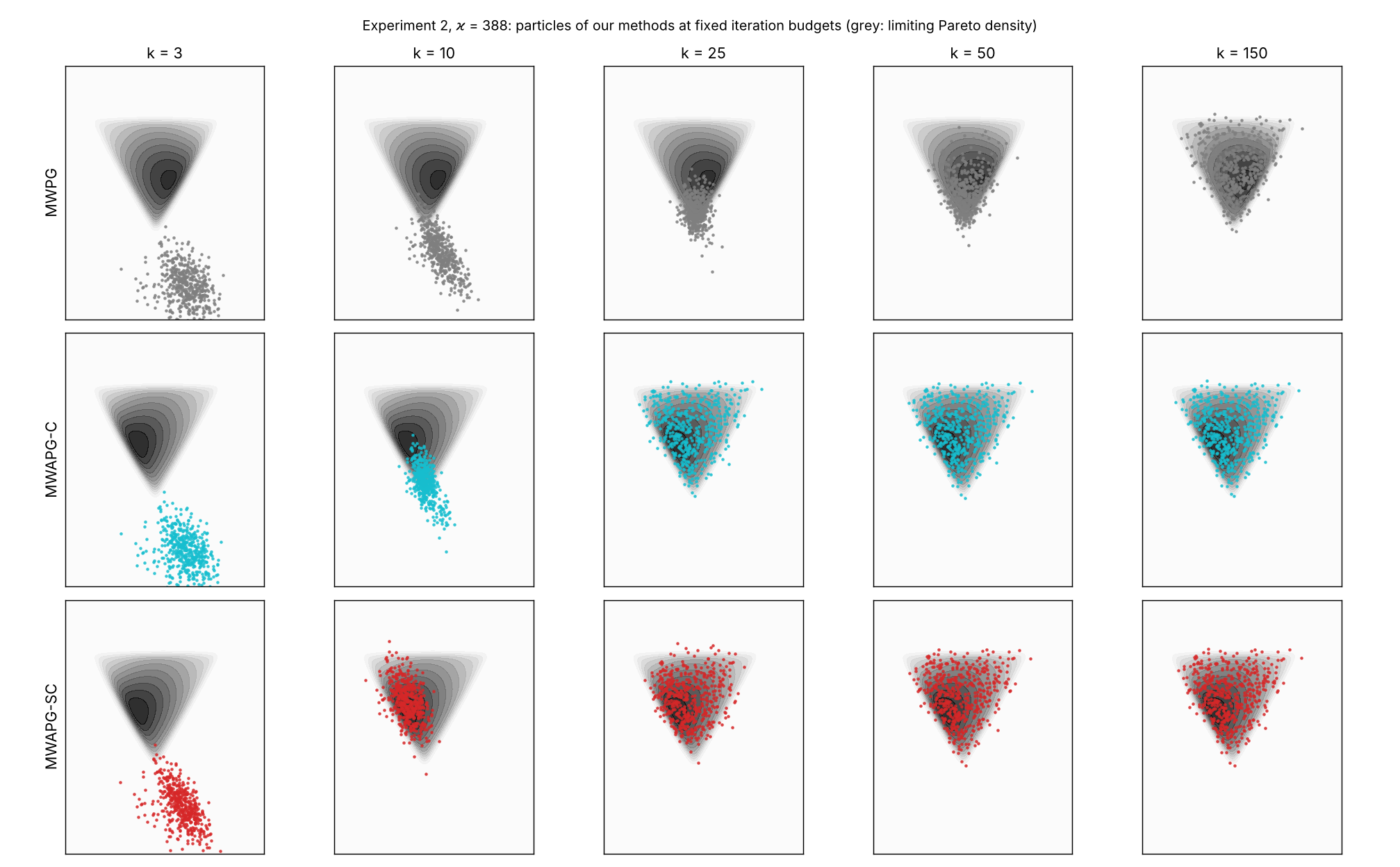}
\caption{Experiment 2, $\kap=388$: particles of MWPG, MWAPG-C and MWAPG-SC after $k=3,10,25,50,150$ iterations, over the
limiting Pareto density (grey).}
\label{fig:exp2-snap}
\end{figure}

\begin{figure*}[h]
\centering
\includegraphics[width=\linewidth]{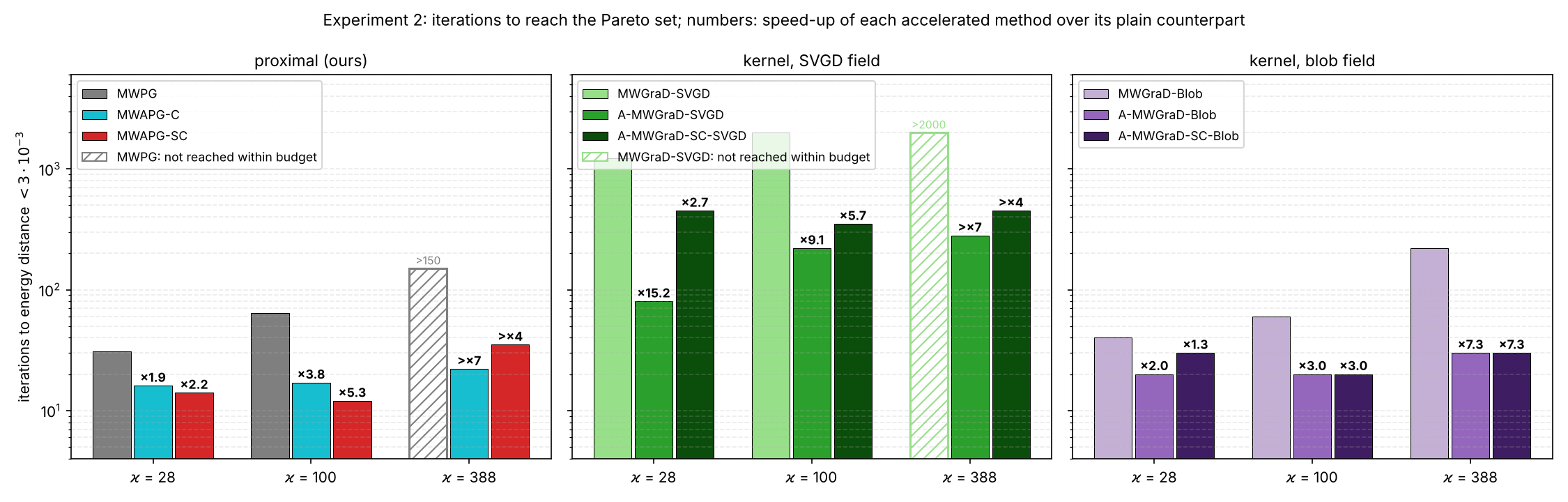}
\caption{Experiment 2: iterations to reach energy distance $3\cdot10^{-3}$, and speed-up of each accelerated method over its
plain counterpart. Hatched bars: threshold not reached within the budget.}
\label{fig:exp2-speedup}
\end{figure*}

\begin{figure}[h]
\centering
\includegraphics[width=\linewidth]{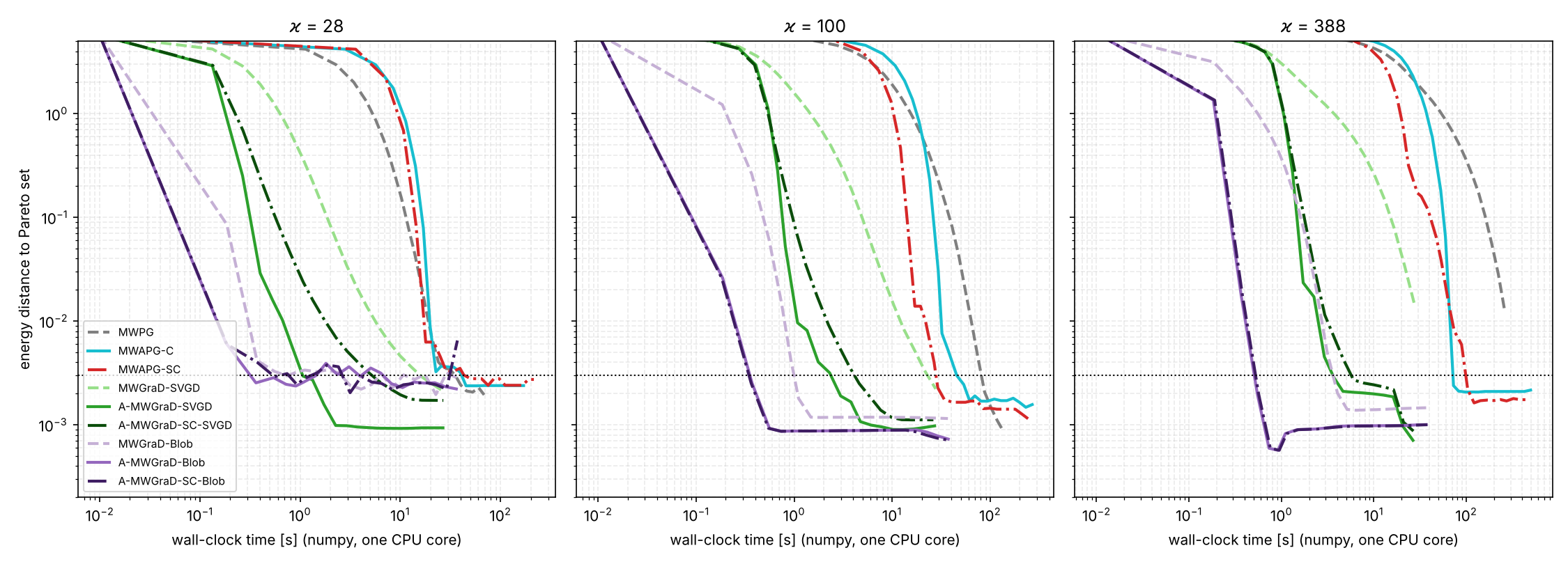}
\caption{Experiment 2: energy distance against wall-clock time (numpy, one CPU core).}
\label{fig:exp2-time}
\end{figure}

\paragraph{Results.}
Table~\ref{tab:exp2} gives iterations and wall-clock times. Figure~\ref{fig:exp2-time} plots the energy distance against time.
Per iteration our methods are about 100 times more expensive than the kernel methods at $n=500$, $d=2$, so in wall-clock time
the accelerated blob methods are fastest on this problem. The comparison in iterations isolates the effect of acceleration,
which is the subject of this paper; the cost per iteration of our implementation is dominated by the network fits, which are
not optimized (numpy, one CPU core, no amortization).

\begin{table}[h]
\centering
\caption{Experiment 2: iterations (seconds) to energy distance $<3\cdot10^{-3}$; in brackets: speed-up over the non-accelerated method of
the same family; ``--'': not reached within the budget. The monotone runs were timed while other jobs shared the two CPU cores;
their extra cost per iteration (one evaluation of the $K$ objectives) is small.}
\label{tab:exp2}
\small
\begin{tabular}{lrrr}
\toprule
 & $\kap=28$ & $\kap=100$ & $\kap=388$ \\
\midrule
MWPG            & 31 (36\,s) & 64 (88\,s) & -- ($>150$) \\
MWAPG-C (monotone)  & 16 (45\,s) [$\times1.9$] & 17 (46\,s) [$\times3.8$] & \textbf{22} (73\,s) [$>\times6.8$] \\
MWAPG-SC (monotone) & 14 (50\,s) [$\times2.2$] & \textbf{12} (29\,s) [$\times5.3$] & 35 (103\,s) [$>\times4.3$] \\
MWAPG-C (plain)     & 14 (26\,s) [$\times2.2$] & 14 (27\,s) [$\times4.6$] & 22 (44\,s) [$>\times6.8$] \\
MWAPG-SC (plain)    & \textbf{8} (15\,s) [$\times3.9$] & 35 (68\,s) [$\times1.8$] & 41 (66\,s) [$>\times3.7$] \\
\midrule
MWGraD-SVGD     & 1220 (16\,s) & 2000 (28\,s) & -- ($>2000$) \\
A-MWGraD-SVGD   & 80 (1.1\,s) [$\times15$] & 220 (3.0\,s) [$\times9.1$] & 280 (3.7\,s) [$>\times7.1$] \\
A-MWGraD-SC-SVGD& 450 (6.1\,s) [$\times2.7$] & 350 (4.6\,s) [$\times5.7$] & 450 (6.0\,s) [$>\times4.4$] \\
\midrule
MWGraD-Blob     & 40 (0.7\,s) & 60 (1.1\,s) & 220 (4.0\,s) \\
A-MWGraD-Blob   & 20 (0.4\,s) [$\times2.0$] & 20 (0.4\,s) [$\times3.0$] & 30 (0.6\,s) [$\times7.3$] \\
A-MWGraD-SC-Blob& 30 (0.6\,s) [$\times1.3$] & 20 (0.4\,s) [$\times3.0$] & 30 (0.6\,s) [$\times7.3$] \\
\bottomrule
\end{tabular}
\end{table}

\subsection{Experiment 3: setting and main results}
\label{app:exp3main}

\paragraph{Setting.}
Each of $K$ clients holds a dataset $D_i$ and has the posterior
$\pi_i\propto\mathcal N(0,I)\prod_{(x,y)\in D_i}\sigma(y\,x^\top\theta)$. Thus $V_i$ is $1$-strongly convex and $L_i$-smooth with
$L_i=1+\frac14\lambda_{\max}(X_i^\top X_i)$, and Assumption~1 holds exactly. As in federated learning, the common
distribution should suit every client. We use two real datasets.
\emph{Breast cancer} (Wisconsin, $569\times30$): there are $K=4$ clients, given by the quartiles of the feature ``mean texture''.
This creates a covariate shift between clients (benign rate 28--91\%); $d=31$ and $\kap=587$.
\emph{Digits} ($8\times8$ images, $1797\times64$): the task is ``digit $\ge5$''. There are $K=5$ clients with label skew, and client
$i$ holds the digits $\{i,i+5\}$; $d=62$ and $\kap=1169$. Each client keeps 30\% of its data for testing. We use 5 random splits.

\paragraph{Methods.}
Logistic posteriors are close to Gaussian, so we run our methods in the Gaussian family. That is, we run
Algorithm~\ref{alg:mwapg} on the closed convex cone of affine Brenier maps, with the proximal map of Proposition~\ref{prop:gauss}.
We estimate $\E_\nu V_i(A\xi+b)$ on 1024 whitened base samples. All our results apply to this problem without change
(Appendix~\ref{app:exp3}). The baselines are MWGraD, A-MWGraD and A-MWGraD-SC in the same family (with the exact Gaussian score; see
Appendix~\ref{app:amwgrad}), and MOO-SVGD, MWGraD-SVGD/Blob and A-MWGraD-SVGD/Blob with 100 particles. Our methods use the step $1/L$ without tuning. We tune
the step and the bandwidth of each baseline on the test NLL of split 0, which favours the baselines. We report three quantities.
The first is the Pareto stationarity $w(\mu)$. It is the optimal value of the subproblem at $Y=X$ with the sign flipped, and it is
zero exactly at Pareto critical points. The second is the test accuracy and NLL, averaged over clients. The third is the fidelity
to the exact Pareto posterior $\pi_{\hat{\boldsymbol{\lambda}}}$ at the final weights of each method, which we sample by MALA.

\begin{table*}[t]
\centering
\caption{Experiment 3 (mean over 5 splits). $k_w$: iterations until Pareto stationarity $w<10^{-4}$ (median over splits; in
brackets: final $w$, after 300 iterations for MWAPG-C/SC and 3000 for all others); Acc/NLL: average over clients of the test
accuracy and test NLL; $\log\frac{\mathrm{var}}{\mathrm{var}^\star}$: mean log ratio of the marginal variances to those of the exact
Pareto posterior $\pi_{\hat{\boldsymbol{\lambda}}}$ (0 is exact, $<0$ under-dispersed). n/a: $w$ is not defined for particle methods,
because the entropy of an empirical measure is not finite; --: MOO-SVGD has no global weights
$\hat{\boldsymbol{\lambda}}$. Our methods: monotone versions with
restart, step $1/L$ untuned; baselines are tuned on the validation set.}
\label{tab:exp3}
\small\setlength{\tabcolsep}{3pt}
\resizebox{\linewidth}{!}{\begin{tabular}{l rccr rccr}
\toprule
 & \multicolumn{4}{c}{Breast cancer ($d=31$, $K=4$, $\kap=587$)} & \multicolumn{4}{c}{Digits ($d=62$, $K=5$, $\kap=1169$)}\\
\cmidrule(lr){2-5}\cmidrule(lr){6-9}
Method & $k_w$ & Acc & NLL & $\log\frac{\mathrm{var}}{\mathrm{var}^\star}$ & $k_w$ & Acc & NLL & $\log\frac{\mathrm{var}}{\mathrm{var}^\star}$\\
\midrule
MWPG & 699 ($7\cdot10^{-10}$) & 0.972 & 0.099 & $-0.01$ & 1207 ($2\cdot10^{-7}$) & 0.850 & 0.374 & $-0.04$ \\
\textbf{MWAPG-C} & 82 ($1\cdot10^{-10}$) & 0.972 & 0.098 & $-0.01$ & 178 ($6\cdot10^{-6}$) & 0.851 & 0.375 & $-0.04$ \\
\textbf{MWAPG-SC} & 72 ($<10^{-12}$) & 0.973 & 0.095 & $-0.01$ & 137 ($9\cdot10^{-10}$) & 0.851 & 0.375 & $-0.04$ \\
\midrule
MWGraD-G & 337 ($6\cdot10^{-6}$) & 0.972 & 0.098 & $-0.01$ & 699 ($3\cdot10^{-5}$) & 0.852 & 0.375 & $-0.04$ \\
A-MWGraD-G & 113 ($5\cdot10^{-6}$) & 0.971 & 0.096 & $-0.01$ & 163 ($3\cdot10^{-5}$) & 0.850 & 0.376 & $-0.04$ \\
A-MWGraD-SC-G & 113 ($5\cdot10^{-6}$) & 0.972 & 0.096 & $-0.01$ & 234 ($4\cdot10^{-5}$) & 0.831 & 0.416 & $-0.03$ \\
\midrule
MWGraD-SVGD & n/a & 0.971 & 0.096 & $+0.01$ & n/a & 0.873 & 0.322 & $+0.95$ \\
A-MWGraD-SVGD & n/a & 0.972 & 0.089 & $-0.28$ & n/a & 0.864 & 0.337 & $-0.17$ \\
MWGraD-Blob & n/a & 0.974 & 0.093 & $-9.43$ & n/a & 0.803 & 0.487 & $-2.81$ \\
A-MWGraD-Blob & n/a & 0.974 & 0.093 & $-3.84$ & n/a & 0.732 & 0.681 & $-2.85$ \\
MOO-SVGD & n/a & 0.972 & 0.107 & -- & n/a & 0.849 & 0.380 & -- \\
\bottomrule
\end{tabular}}
\end{table*}

\paragraph{Results.}
\begin{itemize}
\item \textbf{Convergence to the Pareto set.} MWAPG-C and MWAPG-SC reach $w<10^{-4}$ in 82 and 72 iterations on breast cancer,
and in 178 and 137 iterations on digits. This is $7$--$10\times$ faster than MWPG. MWAPG-SC reaches $w<10^{-8}$ in 178 and 263
iterations, in 9\,s and 39\,s. No objective ever increases; the plain versions overshoot by up to $0.9$ nats. The Gaussian
baselines also reach $w<10^{-4}$ (A-MWGraD with restart: 113 and 163 iterations). However, they then stall at
$w\approx5\cdot10^{-6}$ on breast cancer and $3\cdot10^{-5}$ on digits, for every step size. This floor comes from the
discretization. The baselines move the sample points $x_j=m+\Sigma^{1/2}\xi_j$ along their fields, so their fixed point is a zero
of the fields at the 1024 base samples. This point differs slightly from the stationary point of the sampled objective, which
$w$ measures and which our methods solve exactly. The floor decreases when the number of base samples grows: for MWGraD on breast
cancer, it is $1.7\cdot10^{-5}$, $6.3\cdot10^{-6}$ and $1.2\cdot10^{-6}$ with 256, 1024 and 4096 samples.
\item \textbf{Fidelity.} Our Gaussian solutions agree with the exact Pareto posterior. The means differ by 0.02--0.03 posterior
standard deviations, and the variances differ by 1--4\%. Their predictive distribution equals that of the exact posterior
(dotted line in Figure~\ref{fig:exp3}, Appendix~\ref{app:exp3}). The blob particles collapse: their variances are $e^{-2.8}$ to
$e^{-9.4}$ times too small, and their test NLL on digits becomes worse over the iterations. SVGD is over-dispersed on digits
($e^{0.95}$).
\item \textbf{Predictive performance} is similar for all well-behaved methods. One iteration of our methods costs about $5\times$
one iteration of the Gaussian baselines (Appendix~\ref{app:exp3}).
\end{itemize}

Appendix~\ref{app:scalar} compares our method with fixed-weight scalarization started from the pooled posterior.

\subsection{Experiment 3: details and fidelity}
\label{app:exp3}

\paragraph{Predictive performance.} is similar for all well-behaved methods. The test NLL is not the objective and need not
decrease monotonically: MWAPG-SC passes through distributions that predict slightly better than the Pareto point it converges
to, while every objective decreases. MWGraD-SVGD attains a lower NLL on digits (0.322
vs.\ 0.375); the exact Pareto posterior at \emph{its} weights has the same NLL (0.322), so the gain comes from reaching a
different Pareto point, not from a better approximation: all methods converge to start- and dynamics-dependent Pareto points.

\begin{figure*}[h]
\centering
\includegraphics[width=\linewidth]{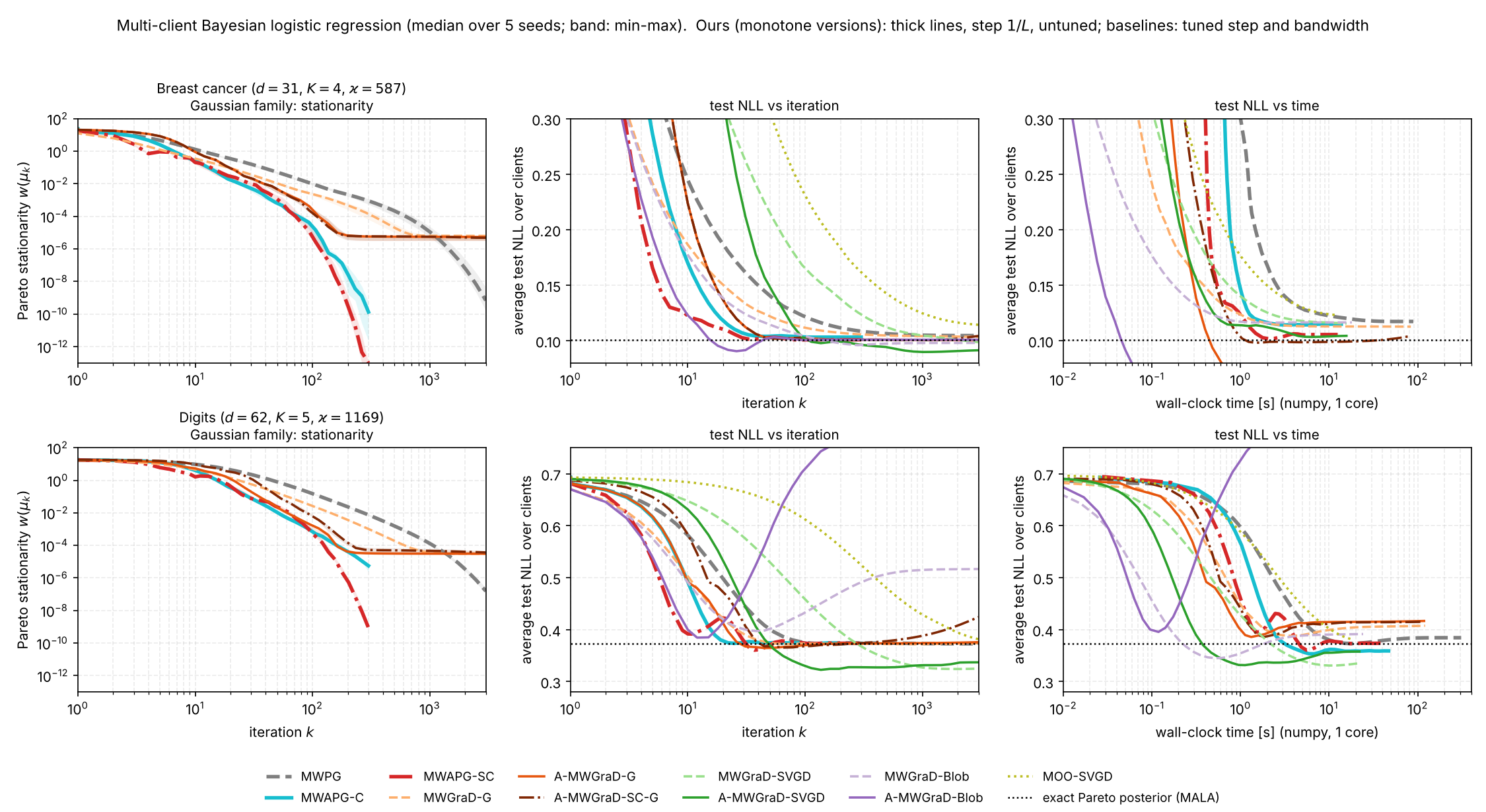}
\caption{Experiment 3: Pareto stationarity of the Gaussian-family methods (left), and average test NLL against iterations (middle) and
wall-clock time (right) for all methods. Median over 5 splits; dotted: exact Pareto posterior at the weights of MWAPG-SC (MALA).}
\label{fig:exp3}
\end{figure*}

\paragraph{Wall-clock time.} An iteration of our method costs about $5\times$ one of the Gaussian baselines, because of the
dual solve. In time to a stable test NLL the Gaussian baselines are fastest (0.3--1\,s), ours need 1--5\,s, and the SVGD methods
2--20\,s (Figure~\ref{fig:exp3}, right); only our methods certify convergence to the Pareto set.

\paragraph{The Gaussian family satisfies our assumptions exactly.}
Let $\xi_1,\dots,\xi_n$ ($n=1024$) be base samples, centred and whitened so that $\frac1n\sum_j\xi_j=0$ and
$\frac1n\sum_j\xi_j\xi_j^\top=I$, and let $\hat F_i(A,b)=\frac1n\sum_jV_i(A\xi_j+b)$. For a direction $\Delta=(\Delta_A,\Delta_b)$,
$\frac{d^2}{dt^2}\hat F_i=\frac1n\sum_j(\Delta_A\xi_j+\Delta_b)^\top\nabla^2V_i(\cdot)(\Delta_A\xi_j+\Delta_b)$, which lies between
$m\|\Delta\|^2$ and $L\|\Delta\|^2$ because $\frac1n\sum_j|\Delta_A\xi_j+\Delta_b|^2=\mathrm{tr}(\Delta_A^\top\Delta_A)+|\Delta_b|^2$ by
whitening. The set $\{A\xi+b:A\text{ symmetric }\succeq0\}$ is a closed convex cone, $-\log\det A$ is convex on it, and the proximal map
of $-\log\det$ restricted to it is the closed form of Proposition~\ref{prop:gauss} (the skew part of $T_A$ is orthogonal to
symmetric matrices). Hence Lemmas~\ref{lem:sub}--\ref{lem:level} and Theorems~\ref{thm:convex}--\ref{thm:inexact} hold verbatim for
this problem, whose Pareto points are the Gaussian Pareto-optimal approximations. The stationarity measure $w$ is computed with
one additional subproblem solve (dual tolerance $10^{-12}$); for the Gaussian baselines, at the symmetric square root of their
covariance.

\paragraph{Baselines.}
Gaussian MWGraD uses the exact fields $-\nabla V_i(A\xi+b)+A^{-\top}\xi$, projected on affine maps, and the min-norm weights
(300 accelerated projected-gradient steps; 5000 steps give identical results). Particle baselines use 100 particles drawn from the
prior; SVGD and blob fields as in Experiment~2, with the Euclidean RBF kernel. MOO-SVGD solves a min-norm problem per particle
(100 steps). The selected (step$\times L$, bandwidth scale) are, breast cancer: MWGraD-G 0.5, A-MWGraD-G 0.5, A-MWGraD-SC-G 0.5,
MOO-SVGD (1, 3), MWGraD-SVGD (1.9, 3), A-MWGraD-SVGD (0.5, 3), MWGraD-Blob (1.9, 0.3), A-MWGraD-Blob (1.9, 10); digits:
MWGraD-G 1.9, A-MWGraD-G 1.9, A-MWGraD-SC-G 1, MOO-SVGD (0.5, 3), MWGraD-SVGD (1.9, 3), A-MWGraD-SVGD (0.5, 10),
MWGraD-Blob (1.9, 1), A-MWGraD-Blob (1.9, 1). Budgets: 300 iterations for MWAPG-C/SC, 3000 for all others.

\begin{figure}[h]
\centering
\includegraphics[width=\linewidth]{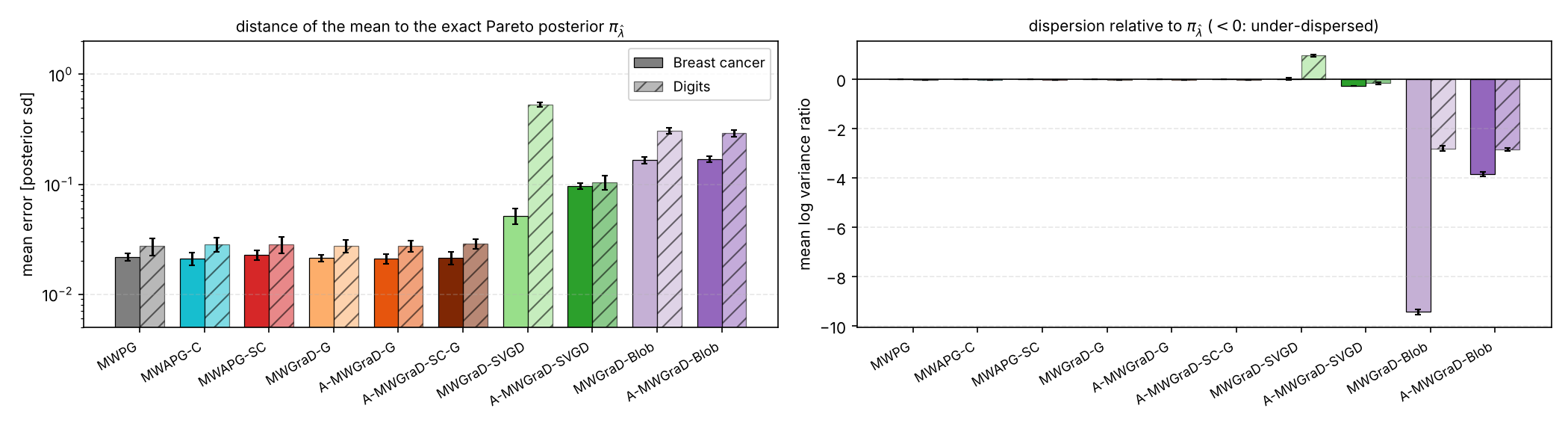}
\caption{Experiment 3: fidelity of each method's final distribution to the exact Pareto posterior at its own weights (mean $\pm$
std over 5 splits; solid: breast cancer, hatched: digits).}
\label{fig:exp3-fid}
\end{figure}

\paragraph{Exact Pareto posteriors.}
For the final weights $\hat{\boldsymbol{\lambda}}$ of each method, $\pi_{\hat{\boldsymbol{\lambda}}}\propto e^{-\sum_i\hat\lambda_iV_i}$ is sampled by MALA with
8 chains, preconditioned by the inverse Hessian at the mode (Newton), 5000 steps with 1500 adaptation steps (acceptance
0.57--0.62), keeping every 5th state (5600 samples). The residual mean error of 0.02--0.03 standard deviations for the Gaussian
methods is at the level of the Monte Carlo error of this reference. MOO-SVGD has no global weights and is excluded from the
fidelity comparison. Figure~\ref{fig:exp3-fid} shows the fidelity metrics.

\subsection{Experiment 3: multiobjective method vs.\ scalarization}
\label{app:scalar}

\paragraph{Setting.}
The pooled posterior has potential $\frac\alpha2\|\theta\|_2^2+\sum_i\ell_i(\theta)$ ($\ell_i$ the negative log-likelihood of client $i$), i.e.
$\sum_iV_i-(K-1)\frac\alpha2\|\theta\|_2^2$; it is computed by the single-objective version of our method (Gaussian VI, 400
iterations). A fixed scalarization with weights ${\boldsymbol{\lambda}}$ is the single objective $V_{\boldsymbol{\lambda}}=\sum_i\lambda_iV_i$, whose minimizer
is the Pareto point $\pi_{\boldsymbol{\lambda}}$; it is solved by the same accelerated method (monotone MWAPG-SC, $K=1$, 250--400 iterations, with
its own smoothness constant $L_{\boldsymbol{\lambda}}\le L$). We report the improvement $G_i(\mu_0)-G_i(\mu)$ of each client's objective over the
start $\mu_0$ (computed exactly up to the common Monte Carlo approximation of the potentials, no normalizing constants are
needed), and test metrics. Random weights are 40 draws from the Dirichlet$(1,\dots,1)$ distribution on split 0.

\begin{table*}[h]
\centering
\caption{Multiobjective MWAPG-SC vs.\ uniform scalarization (mean over 5 splits). Worst/mean: smallest and average KL improvement
over the start across clients [nats]; spread: largest minus smallest improvement; Acc/NLL: average over clients of the test
accuracy/NLL; worst NLL: largest client test NLL. No method makes any client worse in these runs.}
\label{tab:scalar}
\small\setlength{\tabcolsep}{4pt}
\begin{tabular}{lllrrrrrr}
\toprule
Dataset & Start & Method & Worst & Mean & Spread & Acc & NLL & Worst NLL\\
\midrule
Breast cancer & pooled posterior & MWAPG-SC (ours) & $3.70\pm0.14$ & 3.76 & 0.23 & 0.972 & 0.092 & 0.144 \\
 &  & uniform scalarization & $1.90\pm0.31$ & 3.97 & 3.36 & 0.972 & 0.092 & 0.147 \\
\midrule
Breast cancer & prior & MWAPG-SC (ours) & $179.05\pm4.58$ & 194.77 & 45.23 & 0.973 & 0.095 & 0.150 \\
 &  & uniform scalarization & $176.60\pm5.18$ & 196.09 & 52.59 & 0.972 & 0.092 & 0.147 \\
\midrule
Digits & pooled posterior & MWAPG-SC (ours) & $19.38\pm0.39$ & 19.40 & 0.03 & 0.892 & 0.276 & 0.403 \\
 &  & uniform scalarization & $12.31\pm0.41$ & 19.65 & 15.29 & 0.894 & 0.275 & 0.419 \\
\midrule
Digits & prior & MWAPG-SC (ours) & $589.59\pm4.88$ & 591.06 & 2.43 & 0.851 & 0.375 & 0.593 \\
 &  & uniform scalarization & $502.07\pm7.55$ & 631.59 & 227.11 & 0.894 & 0.275 & 0.419 \\
\bottomrule
\end{tabular}
\end{table*}

\begin{figure*}[h]
\centering
\includegraphics[width=\linewidth]{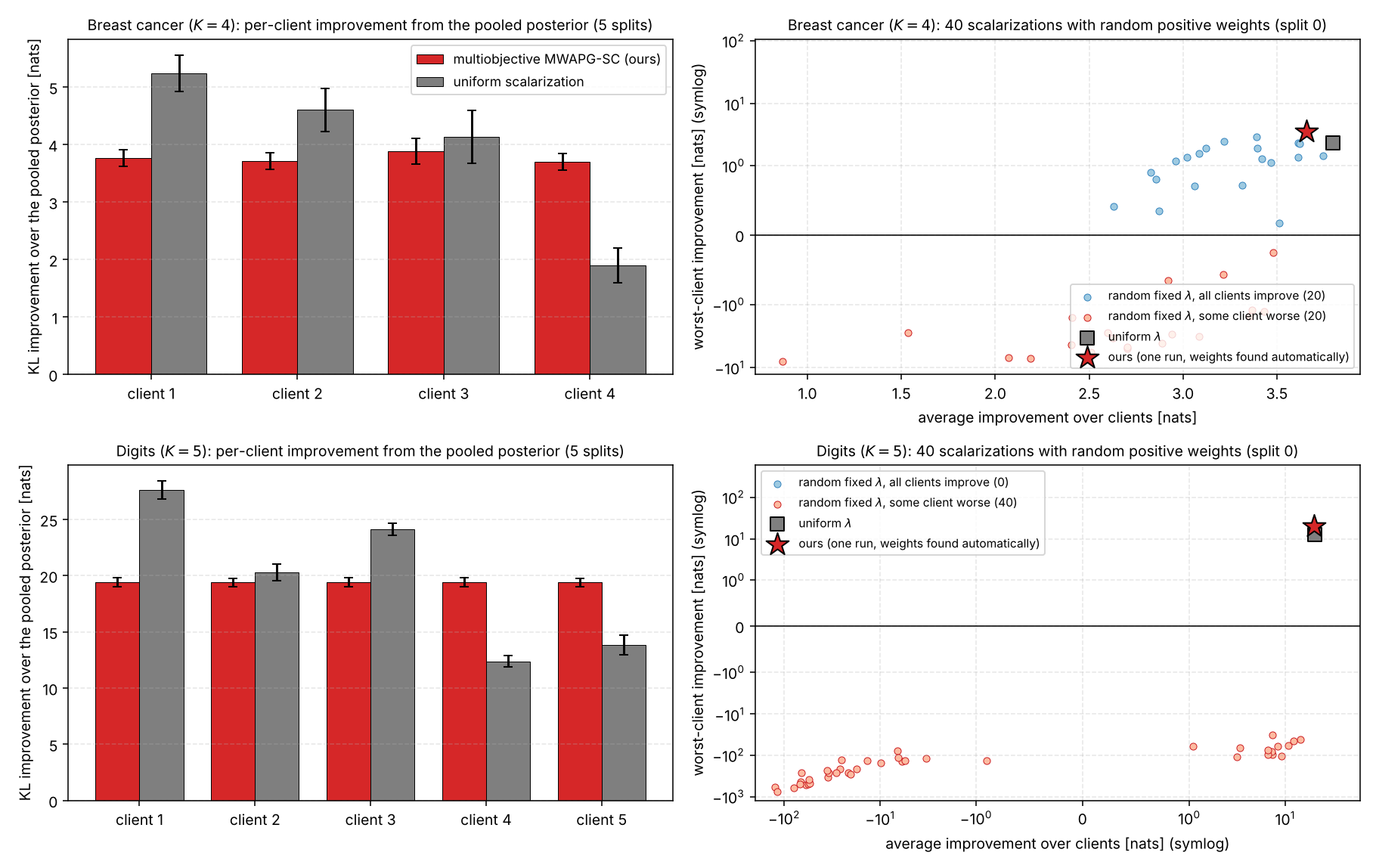}
\caption{Left: KL improvement of each client over the pooled posterior (mean $\pm$ std over 5 splits). Right: average vs.\
worst-client improvement of 40 scalarizations with random positive weights (split 0, symmetric-log axes); red: some client is
worse than the pooled posterior. The multiobjective method (star) finds its weights in a single run.}
\label{fig:scalar}
\end{figure*}

\paragraph{Discussion.}
From the pooled posterior, all clients gain equally under our method, and its weights (e.g.\ $(0.19,0.17,0.22,0.41)$ on breast
cancer) are far from uniform. Fixed weights trade clients against each other: on digits even mildly non-uniform weights move
the posterior so far from some client that its objective increases by tens to hundreds of nats. Started from the prior, both
methods improve all clients by hundreds of nats; our method again balances the gains (spread 2.4 vs.\ 227 nats on digits) by
putting most weight on the hardest client ($\lambda_4\approx0.65$), which lowers the average test performance on digits (NLL 0.375
vs.\ 0.275). Balancing KL gains relative to an uninformative start is therefore not a substitute for choosing the weights; the
method is most useful when an existing model is to be improved for every data source, which is what its guarantee
($G_i(\mu_k)\le G_i(\mu_0)$ for all $i$) expresses.

\subsection{Experiment 4: details and acceleration}
\label{app:exp4}
\paragraph{Model.}
We follow the Bayesian multi-task setting of~\citet{phan2022stochastic}, also used
for A-MWGraD~\citep{nguyen2026accelerated}, with $K=2$ tasks and training data $\mathcal{D}$.
A network has a shared encoder $\mathbf{s}$ and task heads $\mathbf{h}_1,\dots,\mathbf{h}_K$.
The posterior is represented by $n=5$ networks (particles)
$(\mathbf{s}^j,\mathbf{h}^j_1,\dots,\mathbf{h}^j_K)$, $j=1,\dots,n$. Each iteration has two steps.
First, given the heads, the multiobjective method moves the encoders $\{\mathbf{s}^j\}_{j=1}^n$
toward the $K$ targets $\pi_i(\mathbf{s})=p(\mathbf{s}\mid\mathbf{h}_i,\mathcal{D})$, $i=1,\dots,K$.
Second, for each task $i$, the heads $\{\mathbf{h}^j_i\}_{j=1}^n$ are updated by SVGD toward
$p(\mathbf{h}_i\mid\mathbf{s},\mathcal{D})$. The targets therefore change with the heads, and the
entropy enters only through a kernel approximation of its gradient, so our theory does not apply.

\paragraph{Data.}
Multi-MNIST and Multi-Fashion~\citep{phan2022stochastic} overlay two MNIST digits or two
Fashion-MNIST items, and Multi-Fashion+MNIST~\citep{sabour2017dynamic} overlays one of each. Each
image has one label per task. Each dataset has 120{,}000 training and 20{,}000 test images. We
report the test accuracy averaged over the two tasks, using the ensemble prediction of the
$n$ networks.

\paragraph{Baselines.}
We compare with MWGraD~\citep{nguyen2025multiple} and A-MWGraD~\citep{nguyen2026accelerated}.
With the SVGD field, MWGraD coincides with MT-SGD~\citep{phan2022stochastic}.
MGDA~\citep{desideri2012multiple} was outperformed by the multi-target samplers on this
benchmark~\citep{nguyen2025multiple}, so we do not report MT-SGD and MGDA separately.
\paragraph{Protocol.}
The script is the A-MWGraD training script~\cite{nguyen2026accelerated} with a common (``fair'') protocol for all methods: the same
entropy weight $\tau=10^{-4}$ in the kernel term of the shared update, raw minibatch gradients (no score normalisation) and plain
SGD steps for the shared encoder, the momentum coefficient $(k-1)/(k+2)$ with a fixed restart every 1000 iterations for A-MWGraD,
and $(t_k-1)/t_{k+1}$ with the same restart period for MWAPG. MWAPG is run \emph{without} the monotone safeguard of
Algorithm~\ref{alg:mwapg}, unlike in Experiments~1--3: the safeguard accepts a step only if no objective
$F_i=\int V_i\,d\mu+H(\mu)$ increases, but here the entropy enters only through a kernel approximation of its gradient, so
$F_i$ cannot be evaluated, and a comparison of minibatch losses would be dominated by sampling noise. Neither momentum method
uses a safeguard or the adaptive restart; both use only the fixed restart. The heads of all methods are updated by SVGD in output space with
Adam (step $10^{-4}$). Every tenth training image is held out, leaving 108{,}000 training images (422 iterations per epoch);
we evaluate on the 20{,}000 test images every 50 iterations, with the ensemble prediction (average of the softmax outputs of the
5 networks). The reported number is the mean over the last 10 evaluations. The learning rate was fixed as $10^{-3}$ empirically. %Table~\ref{tab:exp4-full} gives both values so that every comparison at a
%common learning rate can be read off. On Multi-Fashion with Blob at $\eta=10^{-3}$, two seeds are available for every method.

\paragraph{Discussion.}
Table~\ref{tab:exp4-early} reports the average test error over the first 2000 iterations, which measures the speed of each
method. Momentum is the main effect: A-MWGraD and MWAPG have about one half to one third of the early error of MWGraD and MWPG.
MWAPG has the lowest early error in all six settings. Its gain over A-MWGraD is small with SVGD (0.5--0.9 points) but clear with
Blob (2.1--4.5 points, i.e.\ a 12--20\% lower error). MWPG and MWGraD are almost identical (within 0.24 points), so the gain of MWAPG
comes from the combination of momentum with the anchored proximal subproblem, not from the proximal step alone.

Table~\ref{tab:exp4-full} reports the average test accuracy over the last 10 evaluations
of training. Acceleration is the dominant factor: on every dataset and kernel, the momentum
methods (A-MWGraD, MWAPG) outperform the non-accelerated ones by $3$--$10$ points, with the
largest gains under the SVGD kernel (about $8$ points on Multi-MNIST and $10$ points on
Multi-Fashion and Multi-F+M). MWPG matches MWGraD in every setting, with differences of at
most $0.10$ points, below the seed-to-seed standard deviation, so the proximal formulation
preserves the accuracy of the baseline.
Among the accelerated methods, MWAPG achieves the best accuracy in five of the six settings,
and the size of its advantage depends on the kernel. With SVGD, MWAPG and A-MWGraD are on par
(differences of $-0.06$, $+0.12$ and $+0.11$ points, comparable to the standard deviations).
With the Blob kernel, MWAPG is clearly better, improving accuracy by $+0.45$ (Multi-MNIST),
$+1.27$ (Multi-Fashion) and $+2.43$ points (Multi-F+M); these gaps exceed the standard
deviations by a wide margin and grow with task difficulty, being largest on the mixed
Multi-F+M dataset.
MWAPG is also considerably more robust to the choice of kernel. Moving from SVGD to Blob
reduces the accuracy of A-MWGraD by $0.81$, $1.33$ and $3.73$ points on the three datasets,
but that of MWAPG by only $0.30$, $0.18$ and $1.41$ points. In contrast, the non-accelerated
methods benefit from the Blob kernel, gaining $4.3$--$5.2$ points over SVGD.
Finally, MWAPG is highly consistent across seeds, with standard deviations between $0.03$ and
$0.10$ in five of the six settings.

\begin{figure}[h]
\centering
\includegraphics[width=\linewidth]{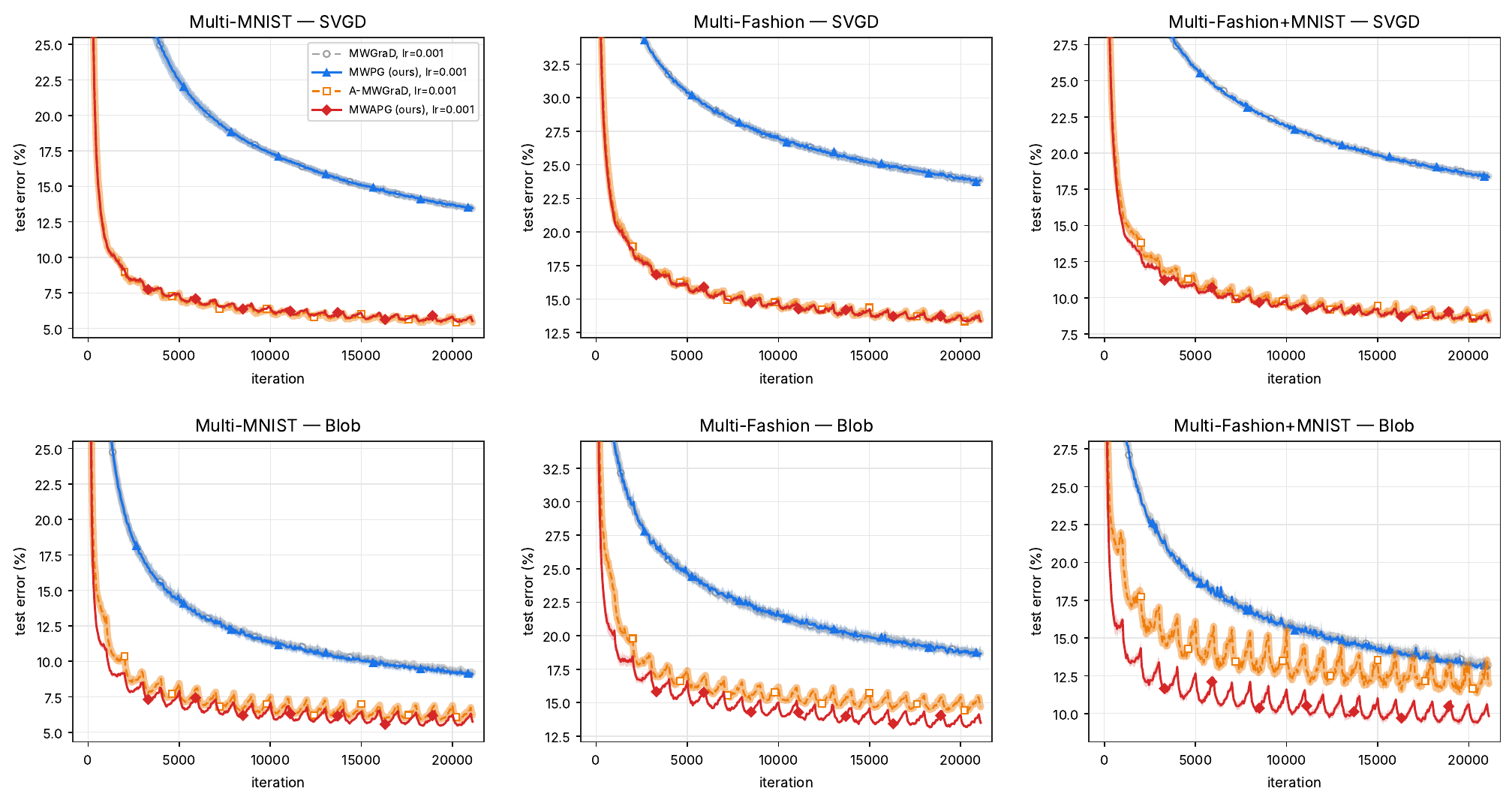}
\caption{Experiment 4: average test error against iterations, mean $\pm$ s.d.\ over seeds. Rows: kernel (SVGD, Blob); columns: dataset. The saw-tooth of the momentum methods comes from the restart every 1000
iterations.}
\label{fig:exp4-curves}
\end{figure}

Figure~\ref{fig:exp4-curves} shows the test error over training iterations. The momentum methods converge
substantially faster: they reach the final error of the non-accelerated methods after only
$650$--$3050$ iterations, compared with the ${\sim}21{,}000$ iterations the latter require,
and their average error over the first $2000$ iterations is $13$--$26\%$, compared with
$35$--$50\%$. The curves of MWPG and MWGraD coincide over the entire run (mean absolute
difference $0.01$--$0.11$ points), confirming that the proximal formulation reproduces the
behaviour of the baseline throughout training and not only at convergence. Among the
accelerated methods, MWAPG reaches this error level earlier than A-MWGraD in every setting,
most clearly with the Blob kernel ($2050$ vs.\ $3050$ iterations on Multi-MNIST and $1200$
vs.\ $2100$ on Multi-Fashion). On Multi-F+M with the Blob kernel, MWAPG drops below this
level after $3050$ iterations and stays there, whereas A-MWGraD keeps rising above it after
each momentum restart until the end of training; its restart oscillations are also larger
($1.9$ vs.\ $1.3$ points peak-to-peak per restart period). Both momentum methods exhibit a
saw-tooth pattern caused by the restart every $1000$ iterations; with the SVGD kernel the two
curves are nearly indistinguishable, consistent with the final accuracies in
Table~\ref{tab:exp4-full}.

\begin{table}[t]
\caption{Experiment 4: average test error (\%) over the first 2000 iterations (40 evaluations), mean over seeds.}
\label{tab:exp4-early}
\centering
\begin{tabular}{llcccc}
\toprule
Dataset & Kernel & MWGraD & MWPG & A-MWGraD & MWAPG \\
\midrule
Multi-MNIST & SVGD & 50.23 & 50.23 & 17.02 & \textbf{16.57} \\
 & Blob & 34.84 & 35.05 & 15.52 & \textbf{13.47} \\
Multi-Fashion & SVGD & 47.01 & 47.00 & 25.73 & \textbf{25.09} \\
 & Blob & 38.83 & 38.85 & 25.10 & \textbf{21.96} \\
Multi-F+M & SVGD & 47.24 & 47.25 & 20.90 & \textbf{19.96} \\
 & Blob & 35.08 & 35.32 & 21.91 & \textbf{17.44} \\
\bottomrule
\end{tabular}
\end{table}

\end{document}